\documentclass{elsarticle}

\usepackage{graphicx} %
\usepackage{xcolor}
\usepackage{stix,fdsymbol}
\usepackage{makecell}
\usepackage{xfrac}
\usepackage{multirow,arydshln,diagbox}
\usepackage{pifont}
\usepackage{subcaption,wrapfig}
\usepackage{placeins}
\usepackage{url}

\graphicspath{{figures/}}

\newcommand{\eg}{\textit{e.g.}}
\newcommand{\ie}{\textit{i.e.}}

\newcommand{\dpad}{\hphantom{0}}

\definecolor{newwhite}{RGB}{240,255,220}

\title{FlyAwareV2: A Multimodal Cross-Domain UAV Dataset for Urban Scene Understanding}
\author{Francesco Barbato\footnotemark[1], Matteo Caligiuri\footnotemark[1], Pietro Zanuttigh\footnotemark[2]}

\begin{document}
    \renewcommand*{\thefootnote}{\fnsymbol{footnote}}
    
    \begin{abstract}
The development of computer vision algorithms for Unmanned Aerial Vehicle (UAV) applications in urban environments heavily relies on the availability of large-scale da\-ta\-sets with accurate annotations. However, collecting and annotating real-world UAV data is extremely challenging and costly. 
To address this limitation, we present FlyAwareV2, a novel multimodal dataset encompassing both real and synthetic UAV imagery tailored for urban scene understanding tasks. 
Building upon the recently introduced SynDrone and FlyAware datasets, FlyAwareV2 introduces several new key contributions: 1) Multimodal data (RGB, depth, semantic labels) across diverse environmental conditions including varying weather and daytime; 2) Depth maps for real samples computed via state-of-the-art monocular depth estimation; 3) Benchmarks for RGB and multimodal semantic segmentation on standard architectures; 4) Studies on synthetic-to-real domain adaptation to assess the generalization capabilities of models trained on the synthetic data. 
With its rich set of annotations and environmental diversity, FlyAwareV2 provides a valuable resource for research on UAV-based 3D urban scene understanding.

\noindent\textbf{Dataset link:} \url{https://medialab.dei.unipd.it/paper_data/FlyAwareV2}

\footnotetext[1]{Authors contributed equally.}
\footnotetext[2]{Corresponding Author.}
\end{abstract}

    \maketitle
    
    \renewcommand*{\thefootnote}{\arabic{footnote}}
    \setcounter{footnote}{0}
    
    \section{Introduction} \label{sec:intro}
The rapid diffusion of Unmanned Aerial Vehicles (UAVs) has revolutionized a wide range of applications, from surveillance and monitoring to precision agriculture and urban planning \cite{motlagh2017uav,kim2018designing,bhatnagar2022mapping}. 
UAV technology has seen a rapid rise in popularity in recent years, driven by a growing range of applications that span from recreational uses to deep integration in critical industrial and agricultural operations \cite{hassanalian2017classifications}.
Drones are now deployed across a variety of domains. For example, police forces use them for security purposes, allowing rapid and efficient monitoring of areas without deploying personnel, or providing a bird’s-eye perspective to support ground operations \cite{bisio2024rf}. 
In agriculture, UAVs are widely used for field evaluation and monitoring, as well as for the precise application of fertilizers \cite{liu2021boost,aslan2022comprehensive}. They have also become widespread in cinematography and photography, offering unmatched flexibility for aerial shots. Beyond these applications, UAVs have demonstrated their effectiveness in numerous other scenarios \cite{xu2022power}.

\begin{wrapfigure}{r}{.6\textwidth}
    \centering
    \setlength{\belowcaptionskip}{-1em}
    \includegraphics[width=.95\linewidth]{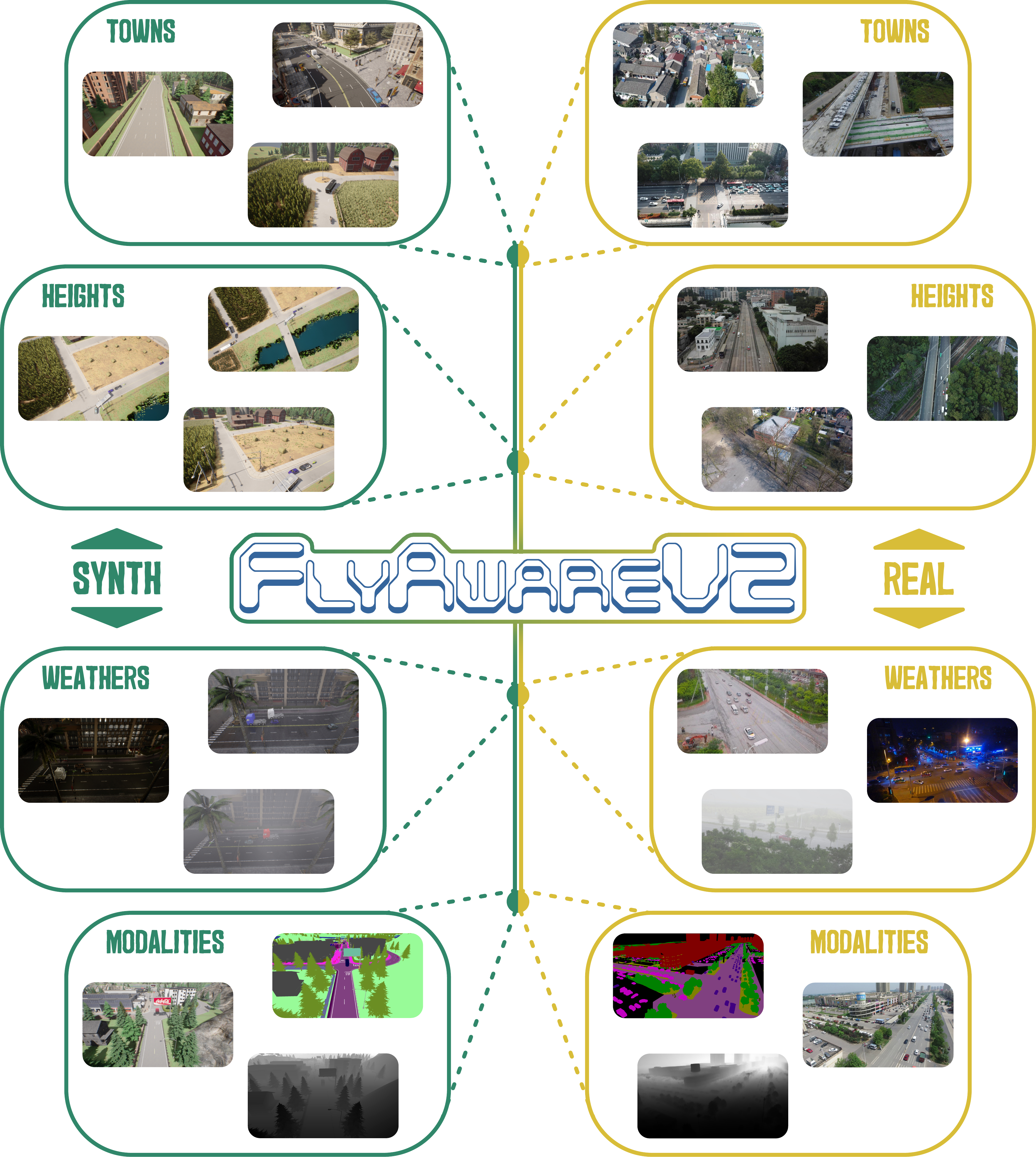}
    \caption{We introduce FlyAwareV2, a mixed-reality multimodal dataset for UAV imagery. We provide synthetic and real samples in varying weather conditions, with ground-truth depth information as well as semantic segmentation labels.}
    \label{fig:graphabstract}
\end{wrapfigure}

As a result, UAVs are increasingly expected to meet high\-er standards in terms of performance, reliability, and operational capabilities. This growing expectation requires the integration of multimodal sensors capable of capturing a comprehensive representation of the surrounding environment, as well as the development of advanced intelligent systems that allow UAVs to accurately interpret sensory data, make informed decisions in real time, reliably avoid obstacles, and perform fully autonomous operations when necessary \cite{liu2025survey}.
All of this necessitates the use of advanced computer vision capabilities, typically achieved using powerful deep learning models. However, the development of robust vision algorithms based on machine learning for UAV imagery is limited by the scarcity of large-scale, accurately annotated datasets that capture the complexity and diversity of the real world.

Existing UAV datasets for tasks like object detection and tracking \cite{benchmark2016benchmark,zhu2021detection,jiang2021anti} provide valuable resources, but often lack the dense pixel-level annotations required for semantic scene understanding. On the other hand, datasets designed for the semantic segmentation of aerial views \cite{nigam2018ensemble,lyu2020uavid,chen2018large} tend to be limited in size, scene variability, sensor modalities, and the range of annotated classes. Moreover, most available datasets focus solely on clear daytime conditions, failing to represent the challenges posed by adverse weather scenarios that UAVs routinely encounter during operations.

To bridge this gap, we introduce FlyAwareV2, a novel multimodal dataset that includes synthetic and real data tailored to understanding urban scenes from aerial imagery under various environmental conditions (a visual example of the content is reported in Figure~\ref{fig:graphabstract}). Building upon our previous efforts on synthetic data generation \cite{rizzoli2023syndrone} and real-world adverse weather translation \cite{rizzoli2025cars}, FlyAwareV2 offers several unique contributions:
\begin{enumerate}
    \item Multimodal data, consisting of color images, depth maps, and semantic labels, cover a wide range of time and weather conditions, such as daytime, nighttime, rain, and fog. This facilitates the development and evaluation of multimodal learning techniques for robust scene understanding.
    \item Depth maps for real samples computed using the state-of-the-art Marigold monocular depth estimation model \cite{marigold}. Leveraging this approach can alleviate the lack of accurate depth annotations in existing UAV datasets.
    \item A unified dataset combining synthetic data from simulators derived from the autonomous driving field and real-world UAV images. This enables systematic studies on cross-domain adaptation and generalization from synthetic to real environments.
    \item Extensive benchmarking for the multimodal semantic segmentation task using popular deep learning architectures, providing solid baselines for future research.
    \item Comprehensive analysis of synthetic-to-real domain adaptation performance, assessing the generalization capabilities of models trained solely on synthetic data when tested on real UAV imagery under diverse conditions.
\end{enumerate}

As already pointed out, the development of FlyAwareV2 is motivated by the need for large, diverse, and accurately annotated datasets to fuel advances in UAV perception, particularly in challenging real-world operating scenarios involving variable weather and illumination. Although simulated environments offer the ability to generate virtually unlimited amounts of annotated data \cite{dosovitskiy2017carla,testolina2023selma}, they often do not capture the nuances and complexities present in real-world observations. In contrast, manually annotating large real-world UAV datasets is extremely labor-intensive and costly. Our contribution aims to combine the complementary strengths of both synthetic and real data sources.

A key contribution of FlyAwareV2 is facilitating multimodal scene understanding by providing co-registered RGB, depth, and semantic data streams. The combination of information across multiple sensor modalities has been shown to improve the robustness and accuracy of perception systems \cite{lahat2015multimodal,rizzoli2022multimodal}, a key requirement for autonomous UAV operations. However,  existing UAV datasets lack coherent multi-sensor data, limiting their applicability for multimodal algorithms development and evaluation.

Another crucial aspect is representing the diverse environmental conditions faced by UAVs during deployments. Adverse weather phenomena such as rain, fog, and night operations can severely degrade the performance of vision algorithms tuned for clear daytime scenarios \cite{xia2023cmda,bruggemann2023refign}. Although recent driving datasets have made strides in this direction \cite{sakaridis2021acdc,yu2018bdd100k}, comparable resources for UAVs have been lacking. FlyAwareV2 aims to close this gap by providing adverse weather data for both synthetic and real aerial imagery, enabling a systematic study of domain adaptation and generalization across environmental conditions.

The importance of large and accurately annotated datasets cannot be overstated for developing data-driven computer vision solutions, especially in the context of safety-critical applications such as autonomous UAV navigation. Through this work, we strive to provide the research community with a valuable resource that can accelerate progress in this rapidly evolving field. In this paper, we present the details of the FlyAwareV2 dataset, extensive experimental evaluations, and insights gained from our analysis, pa\-ving the way for future advances in robust UAV perception in real-world conditions.

    \section{Related Work} \label{sec:related}
With the increasing use of unmanned aerial vehicles (UAVs) in various applications such as surveillance, monitoring, and mapping \cite{motlagh2017uav,kim2018designing}, there has been a growing need for robust computer vision algorithms tailored to aerial imagery. However, the development of such algorithms is hindered by the lack of large-scale annotated data sets that capture the diversity of real-world scenarios encountered by UAVs. This section reviews existing datasets and methods for UAV-based computer vision tasks, with a focus on semantic segmentation.

\subsection*{Datasets for UAV Computer Vision} \label{sub:related:dataset}
Early datasets such as Aeroscapes \cite{nigam2018ensemble} and ICG Drone \cite{icgdrone} pioneered the collection of aerial annotated images, but were limited in scale and diversity. Aeroscapes contains 141 video sequences with 11 semantic classes, while ICG Drone provides high-resolution residential scenes with $22$ classes, but lacks common road objects. The UAVid dataset \cite{lyu2020uavid} offered video sequences from low-altitude UAVs with $300$ labeled frames suitable for semantic segmentation. However, its small size and limited frame rate restrict its utility for training modern deep networks.  

Recent efforts have aimed to create larger and more comprehensive UAV datasets. The Urban Drone Dataset (UDD) \cite{chen2018large} focuses on 3D reconstruction from aerial data across four cities, but is constrained to just four semantic classes. WoodScape \cite{yogamani2019woodscape} provides a multi-task dataset with fisheye cameras and LiDAR from UAVs, enabling applications beyond semantic segmentation. However, it lacks adverse weather conditions that are critical for robust UAV operations.

A notable limitation of most existing real-world UAV datasets is their small size, lack of environmental diversity, and restricted set of annotated classes. This has motivated the use of synthetic data generation. Datasets like SynWoodScape \cite{sekkat2022synwoodscape} and OmniScape \cite{sekkat2020omniscape} leverage game engines to render synthetic aerial views, but are limited to clear daytime conditions. The IDDA \cite{alberti2020idda} and SELMA \cite{testolina2023selma} datasets provide various synthetic driving scenarios with adverse weather and annotations for autonomous driving tasks, although from a ground vehicle perspective.

The SynDrone dataset \cite{rizzoli2023syndrone} represents one of the first attempts to create a large-scale multimodal synthetic dataset specifically for UAV applications. It offers more than $72$K images from drone viewpoints at multiple altitudes, with annotations for semantic segmentation and object detection. However, SynDrone only considers clear daytime conditions, limiting its applicability to real-world UAV deployment in variable weather.

\subsection*{Methods for UAV Semantic Segmentation} \label{sub:related:segmentation}
Given the scarcity of large real-world datasets, several works have explored the usage of unsupervised domain adaptation (UDA) to leverage synthetic data for training models that can be deployed in real imagery. Some approaches use adversarial learning \cite{michieli2020adversarial,luo2019taking} or self-training \cite{zou2018unsupervised} to align features between synthetic and real domains. Others employ data translation to render synthetic data in real-world styles \cite{araslanov2021self,choi2019self}. However, these methods primarily consider ground-level viewpoints and clear daytime conditions.

Only a few studies have specifically targeted UAV semantic segmentation. Nigam et al. \cite{nigam2018ensemble} used ensemble knowledge transfer to adapt a model from the synthetic GTA-V data set to Aeroscapes UAV data. Marcu et al. \cite{marcu2020semantics} proposed semi-supervised label propagation on aerial video sequences. However, these methods do not account for the challenges of adverse weather conditions faced by UAVs in real-world operations.

\begin{wraptable}{r}{.59\textwidth}
    \centering
    \setlength{\belowcaptionskip}{-4em}
    \setlength{\tabcolsep}{.1em}
    \resizebox{.58\textwidth}{!}{%
    \begin{tabular}{cc|cc}
        Fine ID & Fine Name & Coarse Name & Coarse ID \\
        \hline
         0 & Building & \multirow{3}{*}{Building} & \multirow{3}{*}{0} \\
         1 & Fence \\
         8 & Wall \\
         \hdashline
         4 & Road Line & \multirow{5}{*}{Road} & \multirow{5}{*}{1} \\
         5 & Road \\
         6 & Sidewalk \\
         12 & Bridge \\
         13 & Rail Track \\
         \hdashline
         22 & Car & \multirow{6}{*}{Vehicle} & \multirow{6}{*}{2} \\
         23 & Truck \\
         24 & Bus \\
         25 & Train \\
         26 & Motorcycle \\
         27 & Bicycle \\
         \hdashline
         7 & Vegetation & \multirow{3}{*}{Vegetation} & \multirow{3}{*}{3} \\
         11 & Ground \\
         19 & Terrain \\
         \hdashline
         20 & Person & \multirow{2}{*}{Human} & \multirow{2}{*}{4} \\
         21 & Rider \\
         \hdashline
         -1 & Unlabeled & \multirow{10}{*}{Unlabeled} & \multirow{10}{*}{-1} \\
         2 & Fence \\
         3 & Pole \\
         9 & Traffic Sign \\
         10 & Sky \\
         14 & Guard Rail \\
         15 & Traffic Light \\
         16 & Static \\
         17 & Dynamic  \\
         18 & Water 
    \end{tabular}}
    \caption{Coarse-to-Fine Class Mapping}
    \label{tab:cmap}
\end{wraptable}
    
In summary, while significant progress has been made in UAV datasets and vision algorithms, there is a pressing need for data and methods that can handle the multimodal nature of UAV sensors and the diverse operating conditions encountered in practical deployments, including varying weather, lighting, and viewpoints. The FlyAwareV2 dataset presented in this work aims to fill this gap by providing a comprehensive multimodal benchmark with synthetic and real data under adverse environmental conditions.

    \section{The FlyAwareV2 Dataset} \label{sec:dataset}
The dataset proposed in this work extends and improves the original FlyAware dataset in two key ways: the former consists of adding high-resolution depth maps to all samples, both real and synthetic, while the latter introduces novel weather conditions to the synthetic data and refines the weather augmentation strategy for the real samples.

In this section, we present a detailed description of the proposed dataset, highlighting its organization, data sources, and the additional annotations that accompany it. The dataset contains approximately $290$k frames, of which $288$k are synthetically generated and $2$k are real-world samples. All frames depict drone perspectives over diverse urban and rural environments, captured across multiple spatial and environmental conditions to maximize variability.

\subsection{Synthetic UAV Data} \label{sub:dataset:synthetic}
The synthetic portion is constructed from $24$k unique scenes, each rendered under four different weather conditions and at three flight altitudes, resulting in a broad spectrum of environmental and illumination conditions. 
    
Synthetic sequences were generated from $8$ FullHD ($1920 \allowbreak \times \allowbreak 1080$px) video streams rendered at $25$Hz. Each sequence corresponds to a different simulated environment, resulting in a total of roughly $3$k frames for each of the eight unique environments. To emulate realistic drone behavior, we adjusted flight altitude to three representative levels: $20$m, $50$m, and $80$m. In addition to changing height, we varied the camera tilt to simulate realistic observation angles: $30^\circ$ at $20$m, $60^\circ$ at $50$m, and $90^\circ$ at $80$m. This procedure introduces substantial heterogeneity by combining both geometric (height) and viewpoint (orientation) changes.

Each sample in the dataset is paired with depth information. The ground-truth depth maps have been obtained directly from the underlying 3D scene geometry through the rendering engine. 
Beyond depth, we include semantic segmentation labels. Both training and test splits are annotated with fine-grained labels considering $28$ semantic classes, a detailed overview is reported in Table \ref{tab:cmap}. 

The synthetic data was generated using a customized CARLA~0.9.12 simulator~\cite{dosovitskiy2017carla,testolina2023selma,rizzoli2023syndrone}. CARLA, originally developed using Unreal Engine 4 (UE4), provides photorealistic rendering, realistic physics via NVIDIA PhysX, and basic Non-Player Character (NPC) logic to simulate vehicular and pedestrian behaviors. Our modified version extends CARLA with a larger and more diverse set of UE4 assets, encompassing static objects (\textit{e.g.}, buildings, vegetation, traffic signs) and dynamic ones (\textit{e.g.}, vehicles, cyclists, pedestrians), all modeled with consistent scale and realistic proportions.
Note that, in our modified version, the semantic class taxonomy has been extended for better compatibility with established autonomous driving benchmarks~\cite{cordts2016cityscapes,sakaridis2021acdc}. 
    
To support this, additional vehicle categories such as trains, trams, busses, and trucks were introduced~\cite{testolina2023selma}, further enriching the diversity of dynamic entities present in the dataset.
More in detail, the base CARLA library includes $24$ car models, $6$ truck models, $4$ motorcycle models, and $3$ bicycle models, all customizable by color. It also provides $41$ pedestrian models that vary in ethnicity, body build, and clothing, allowing a diverse population simulation. The simulator also offers $8$ detailed towns (Town01–07 and Town10HD), each with unique buildings, layouts, and landmarks, effectively creating $8$ different simulation environments.
Data collection in CARLA is managed through virtual sensors that can be precisely positioned, oriented, and attached to parent actors with rigid or spring-arm dynamics. Sensor outputs can be recorded at every simulation step, with synchronous simulation ensuring consistent timing across multiple high-resolution sensors.

\subsection{Real World UAV Data} \label{sub:dataset:real}
The real-world portion of the dataset is built from $2$k original frames that are further enhanced to increase environmental diversity.

The real samples, split into training and test sets, are derived from the VisDrone~\cite{zhu2021detection} and UAVid~\cite{lyu2020uavid} datasets, respectively.
These datasets include mainly scenes in daylight and clear weather conditions; therefore, we applied synthetic augmentation techniques to introduce variable weather conditions, as detailed in Sec.~\ref{sub:augment:real_weather}.
This results in a set of frames that better matches the environmental variability of the synthetic data. The real samples are split into two resolutions: training images are provided in HD ($1320 \times 720$px), whereas test images have a higher 4K resolution ($3840 \times 2160$px).

Each sample in the dataset is also paired with depth information. Since depth was not provided in the source datasets, we added estimated depth annotations using a state-of-the-art monocular depth predictor, as detailed in Sec.~\ref{sub:augment:depth_est}.

The semantic annotations are provided for the $200$ test frames, thus allowing us to evaluate Unsupervised Domain Adapation (UDA) strategies.
Compared to synthetic data, the label set is coarser, consisting of $8$ semantic classes, which are further consolidated into $5$ super-classes for evaluation purposes. Table~\ref{tab:cmap} reports the mapping between the label sets of the synthetic and real data, thus allowing for synthetic-to-real adaptation, cross-domain training, and coarse-to-fine understanding strategies.

    \section{Data Augmentation Strategies} \label{sec:augment}
In order to build a complete and coherent dataset with multimodal data and all the weather conditions for all the settings we had to resort to some augmentation strategies. In particular, in this section, we detail how we obtained the depth data (Sec. \ref{sub:augment:depth_est}) and how we simulated the various weather conditions for both synthetic and real data (Secs. \ref{sub:augment:synth_weather} and \ref{sub:augment:real_weather}).

\subsection{Depth Estimation}\label{sub:augment:depth_est}
\begin{figure*}
    \centering
    \begin{subfigure}{\textwidth}
        \centering
        \begin{subfigure}{.32\textwidth}
            \centering
            \includegraphics[width=\textwidth]{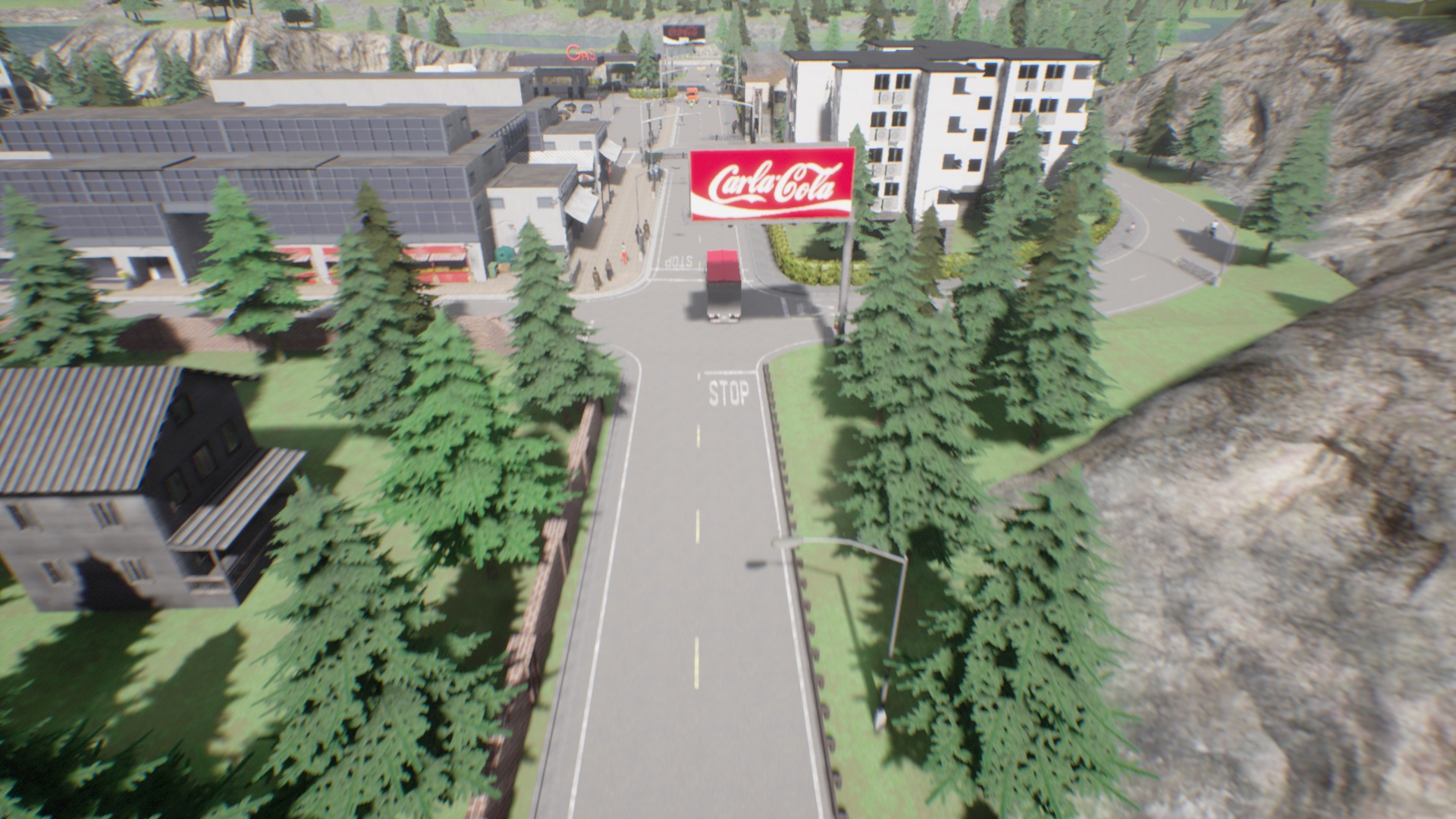}
        \end{subfigure}
        \begin{subfigure}{.32\textwidth}
            \centering
            \includegraphics[width=\textwidth]{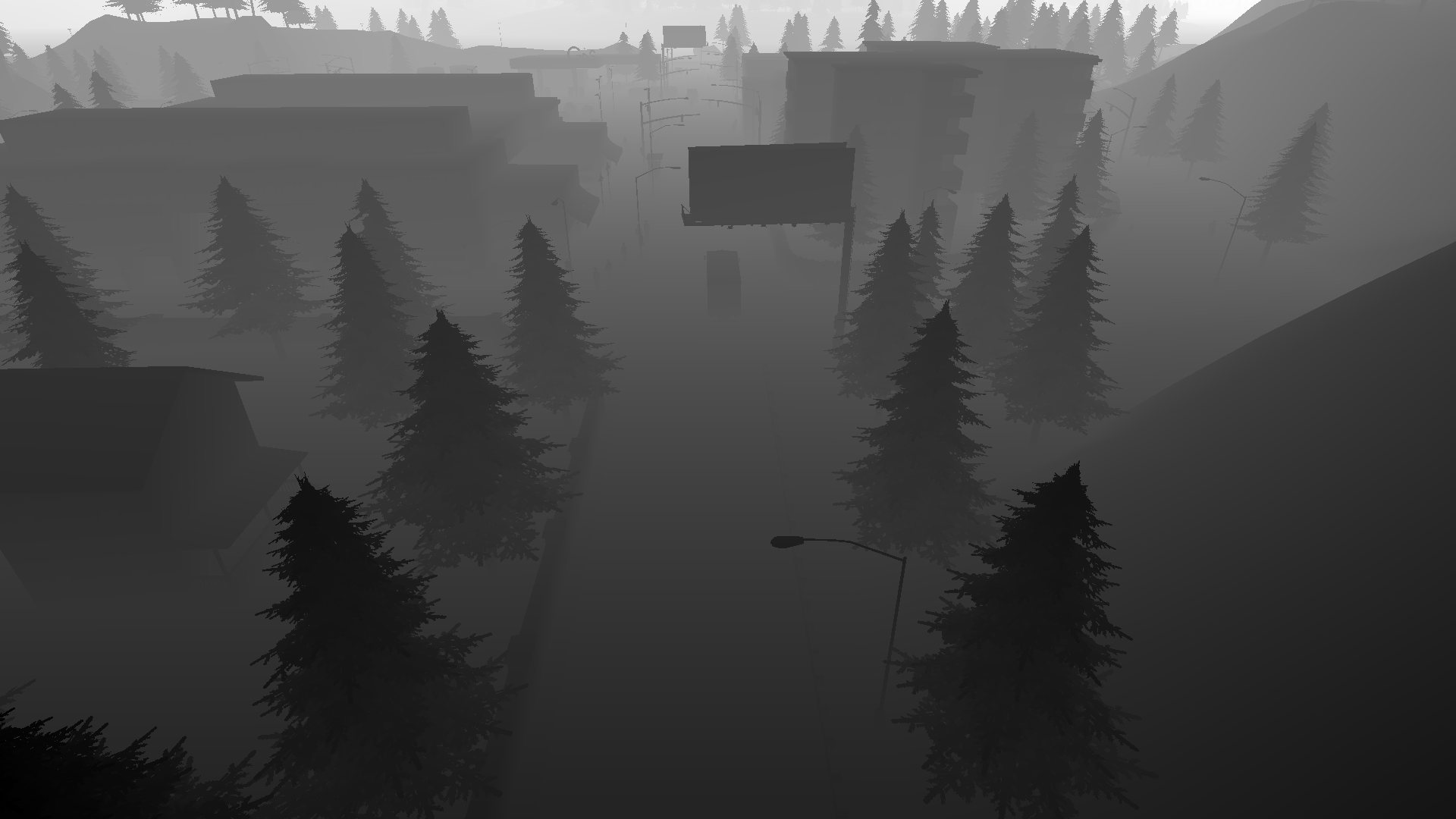}
        \end{subfigure}
        \begin{subfigure}{.32\textwidth}
            \centering
            \includegraphics[width=\textwidth]{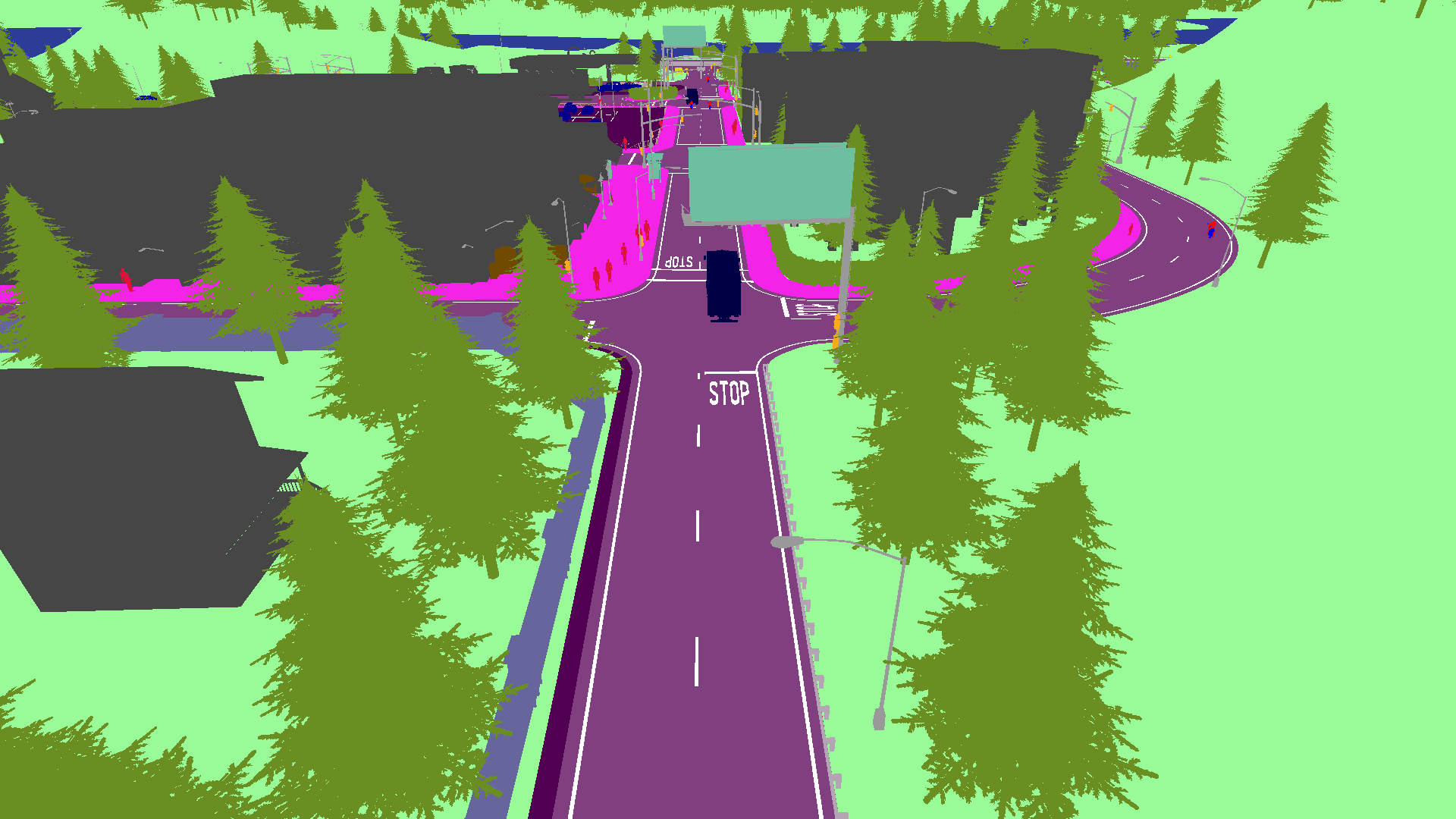}
        \end{subfigure}
    \end{subfigure}
    \begin{subfigure}{\textwidth}
        \centering
        \begin{subfigure}{.32\textwidth}
            \centering
            \includegraphics[width=\textwidth]{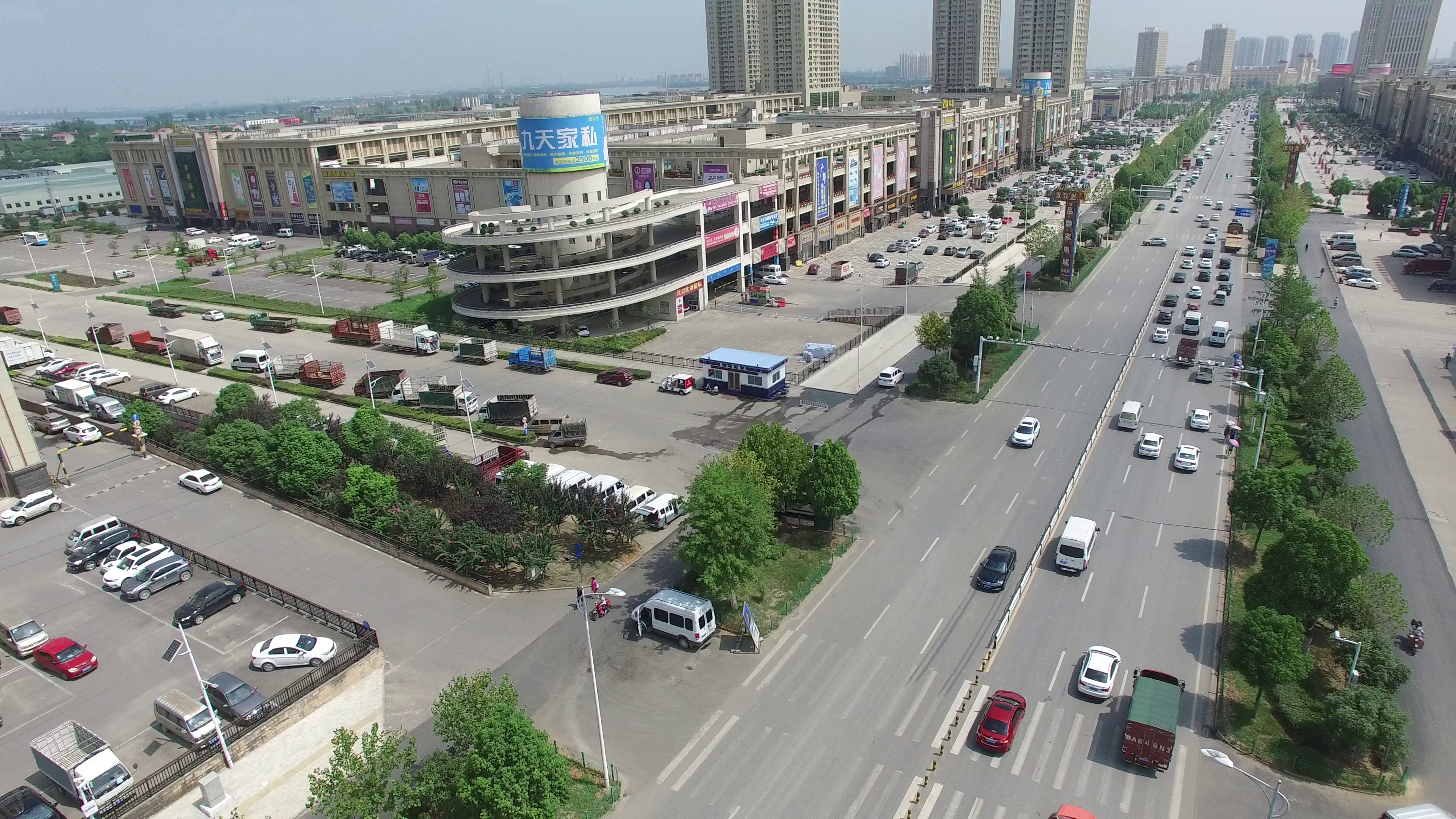}
        \end{subfigure}
        \begin{subfigure}{.32\textwidth}
            \centering
            \includegraphics[width=\textwidth]{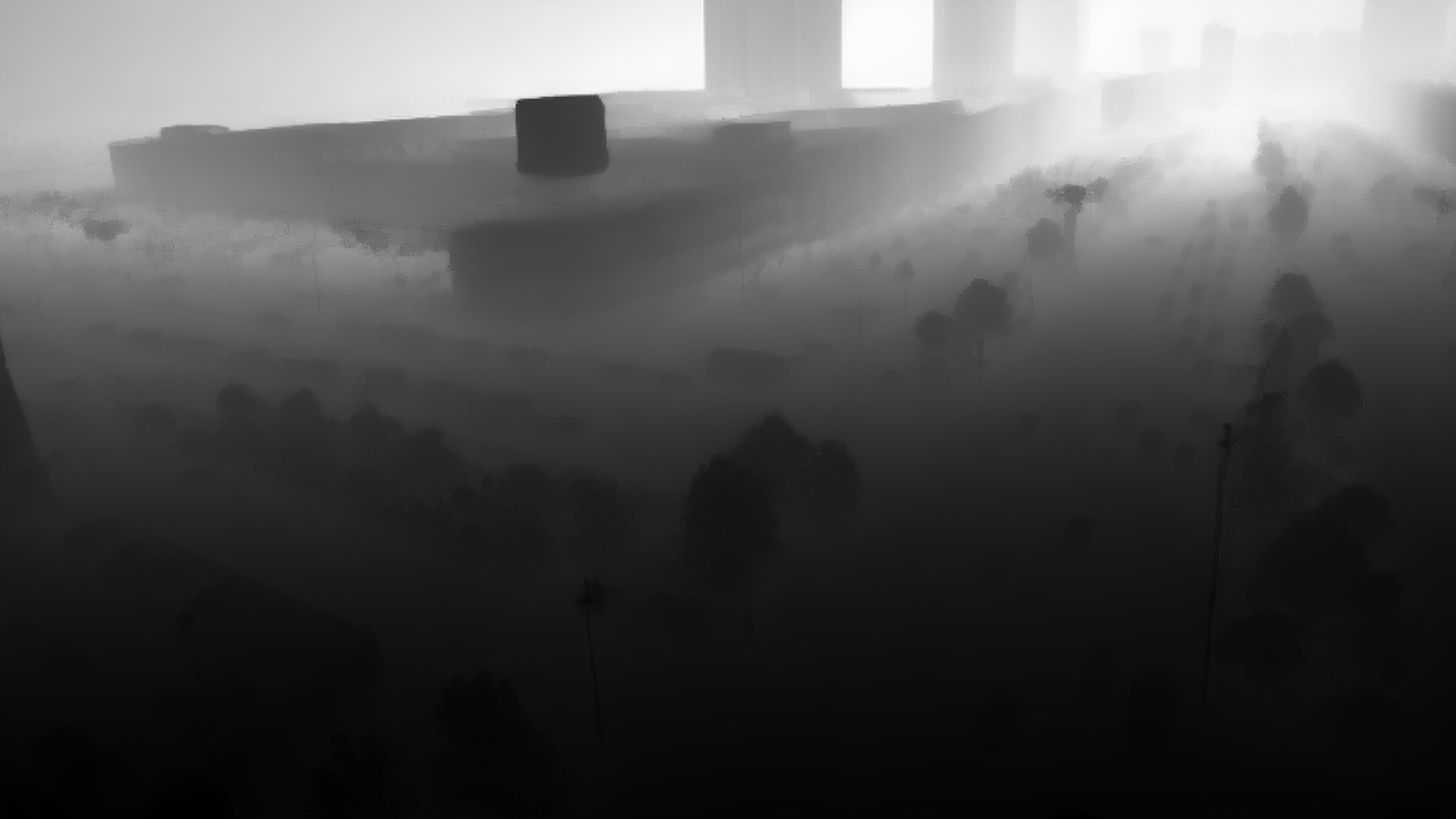}
        \end{subfigure}
        \begin{subfigure}{.32\textwidth}
            \centering
            \includegraphics[width=\textwidth]{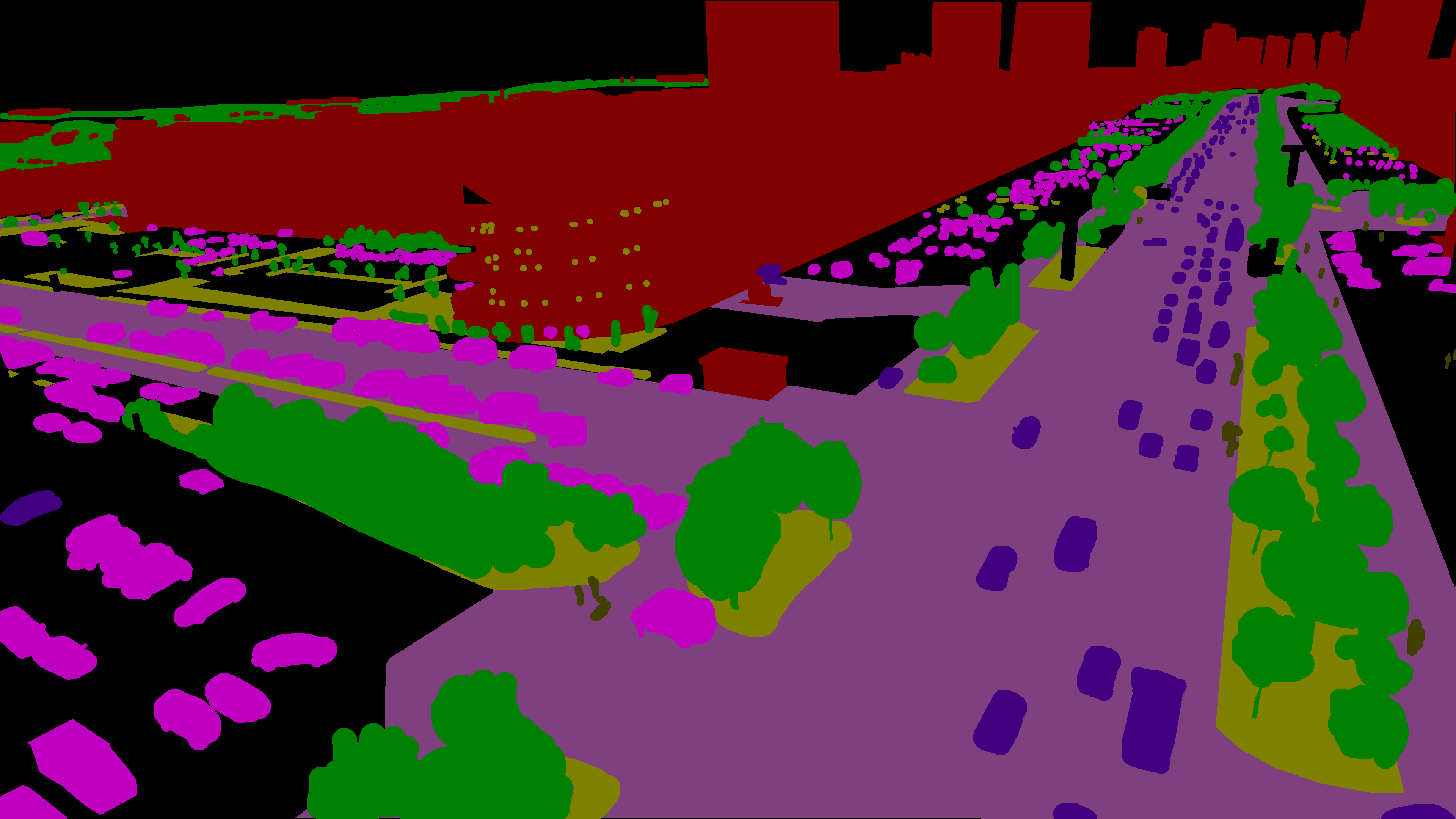}
        \end{subfigure}
    \end{subfigure}
    \caption{The FlyAwareV2 dataset provides color information, depth data and semantic labels for each frame.}
    \label{fig:modalities}
\end{figure*}

As shown in Figure \ref{fig:modalities}, the proposed dataset provides the 3D information for each scene represented through a depth map aligned with the color view.
For the synthetic samples, we simply extracted the ground truth depth information from the underlying 3D geometry of the scenes. 
    
This was not possible for the real samples.
An alternative would be to reconstruct the 3D geometry of real scenes using structure-from-motion techniques; unfortunately, the frame rate in most real-world datasets is not high enough to obtain reliable results with this strategy.
Given these limitations, we had to resort to monocular depth estimation techniques to generate the depth information for the real samples.
    
Nowadays, state-of-the-art strategies for this task are based on deep learning \cite{rajapaksha2024deep}.
For this dataset, we chose to employ the highly performing Marigold \cite{marigold} monocular depth estimation model, which is based on the idea of training a diffusion model for color-to-depth domain translation. 
To align the monocular depths with those of the synthetic samples, we employed the 16-bit depth generation pipeline and re-normalized all depth samples (both synthetic and real) during processing. 

For the training of our benchmark architectures, we normalized independently each depth sample. 
That is,  we rescaled all depthmaps in the range $[0,1]$, regardless of the original maximum or minimum produced by Marigold. In this way, the absolute depth information available in the synthetic samples is destroyed, but the coherence between real and synthetic depth is increased, allowing for better domain transfer of models trained on synthetic data.

\subsection{Weather conditions for synthetic data} \label{sub:augment:synth_weather}
\begin{figure*}[t]
    \centering
    \begin{subfigure}{\textwidth}
        \begin{subfigure}{.245\textwidth}
            \centering
            \includegraphics[width=\textwidth]{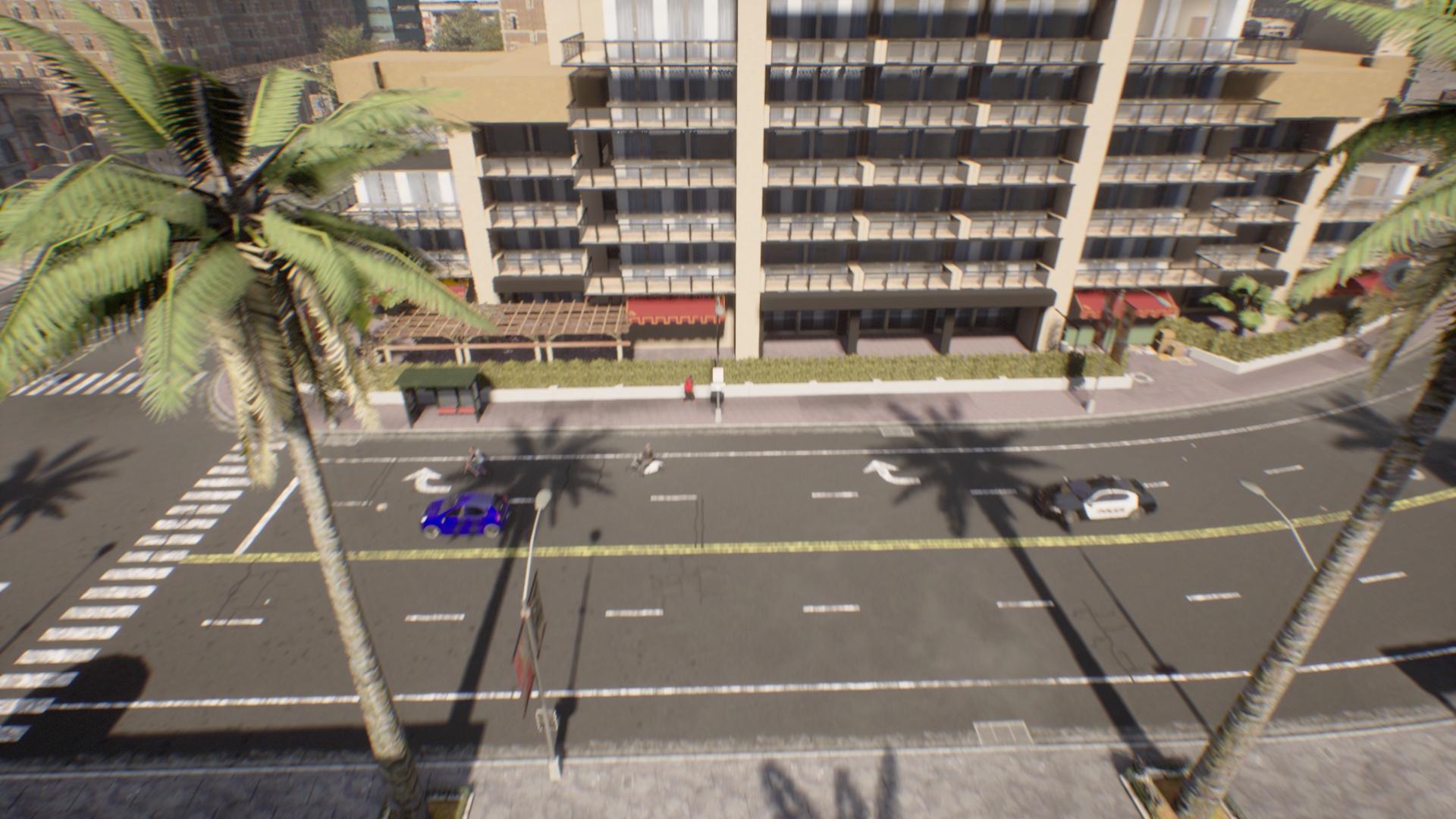}
        \end{subfigure}
        \begin{subfigure}{.245\textwidth}
            \centering
            \includegraphics[width=\textwidth]{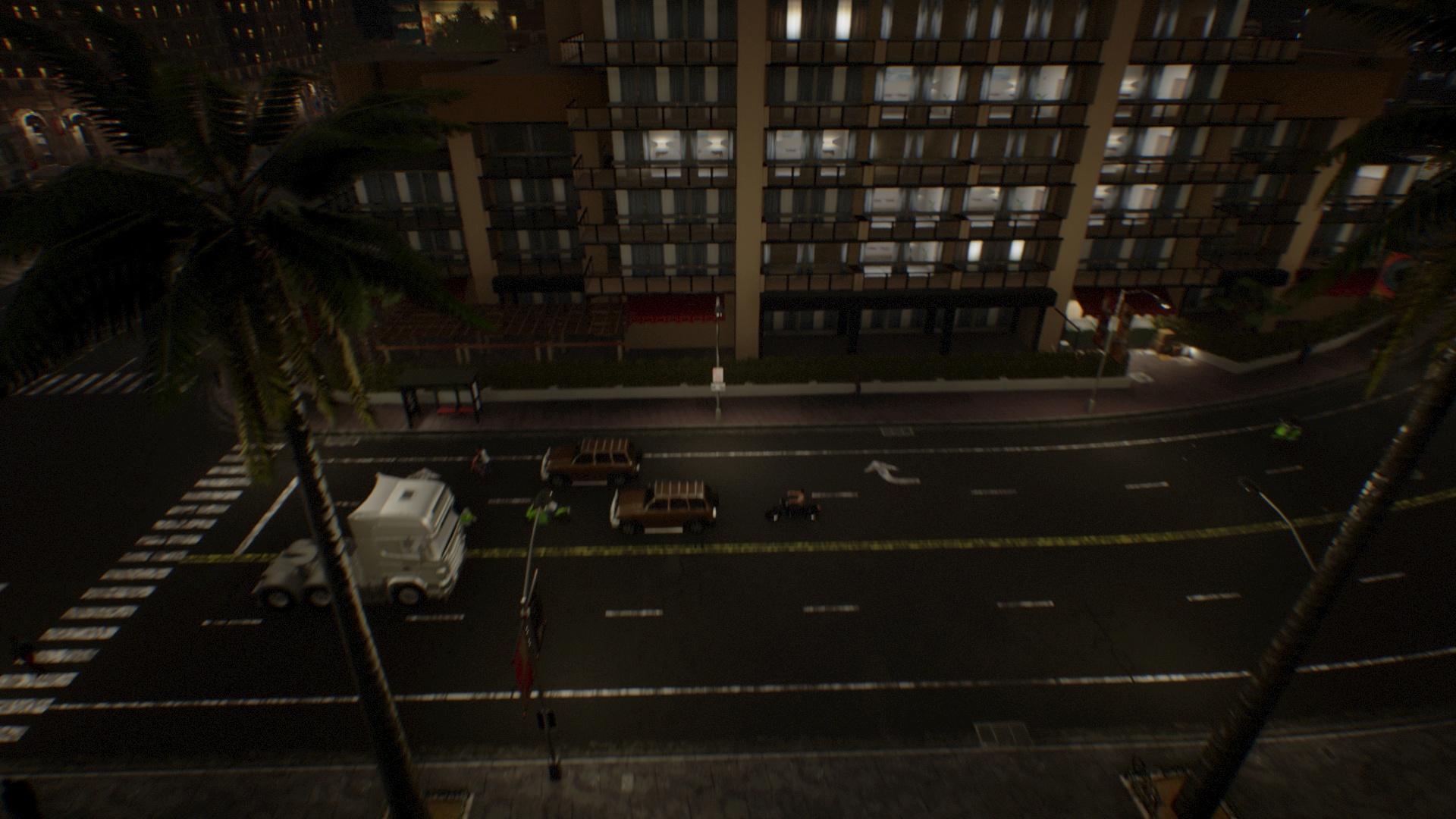}
        \end{subfigure}
        \begin{subfigure}{.245\textwidth}
            \centering
            \includegraphics[width=\textwidth]{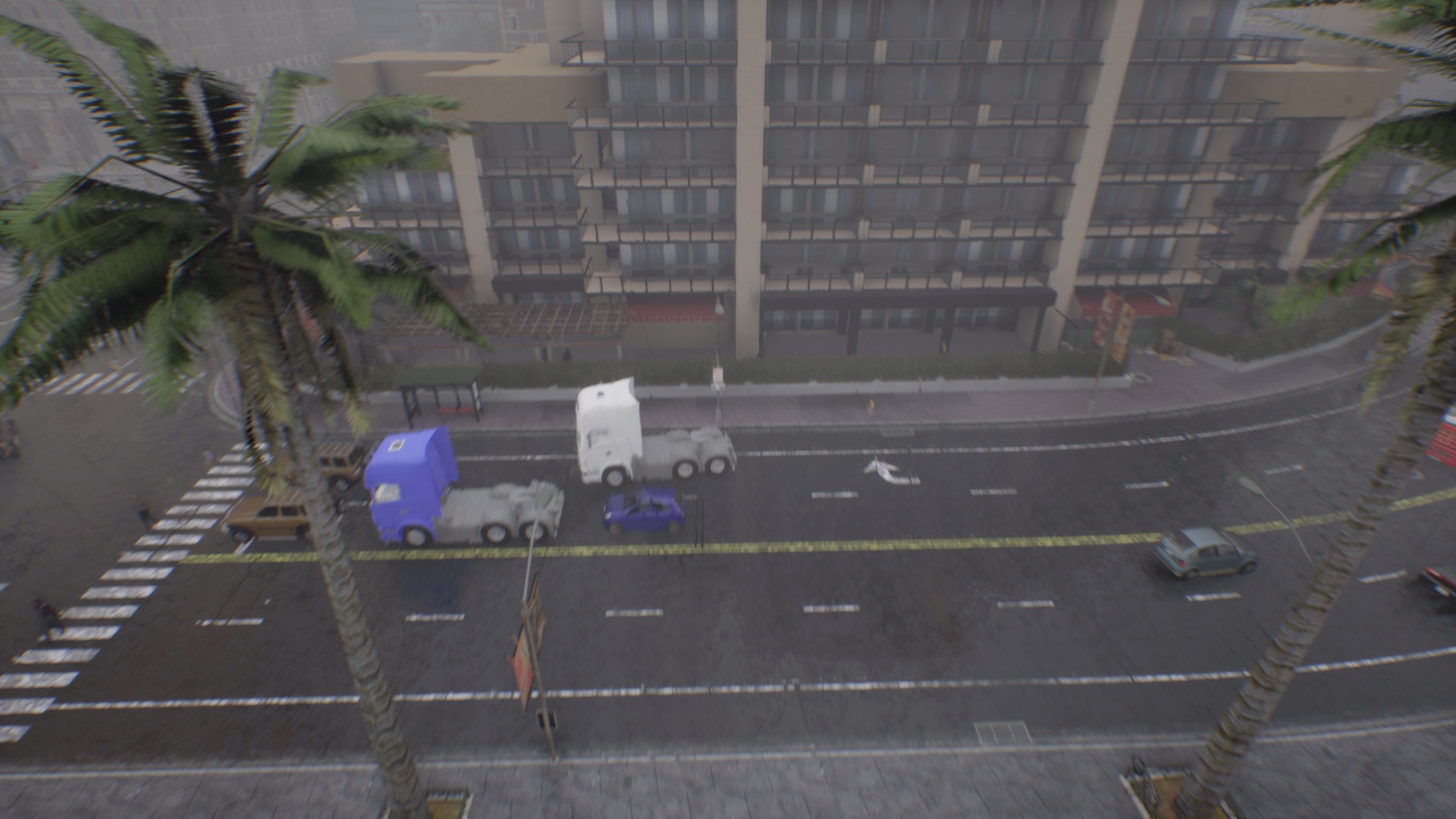}
        \end{subfigure}
        \begin{subfigure}{.245\textwidth}
            \centering
            \includegraphics[width=\textwidth]{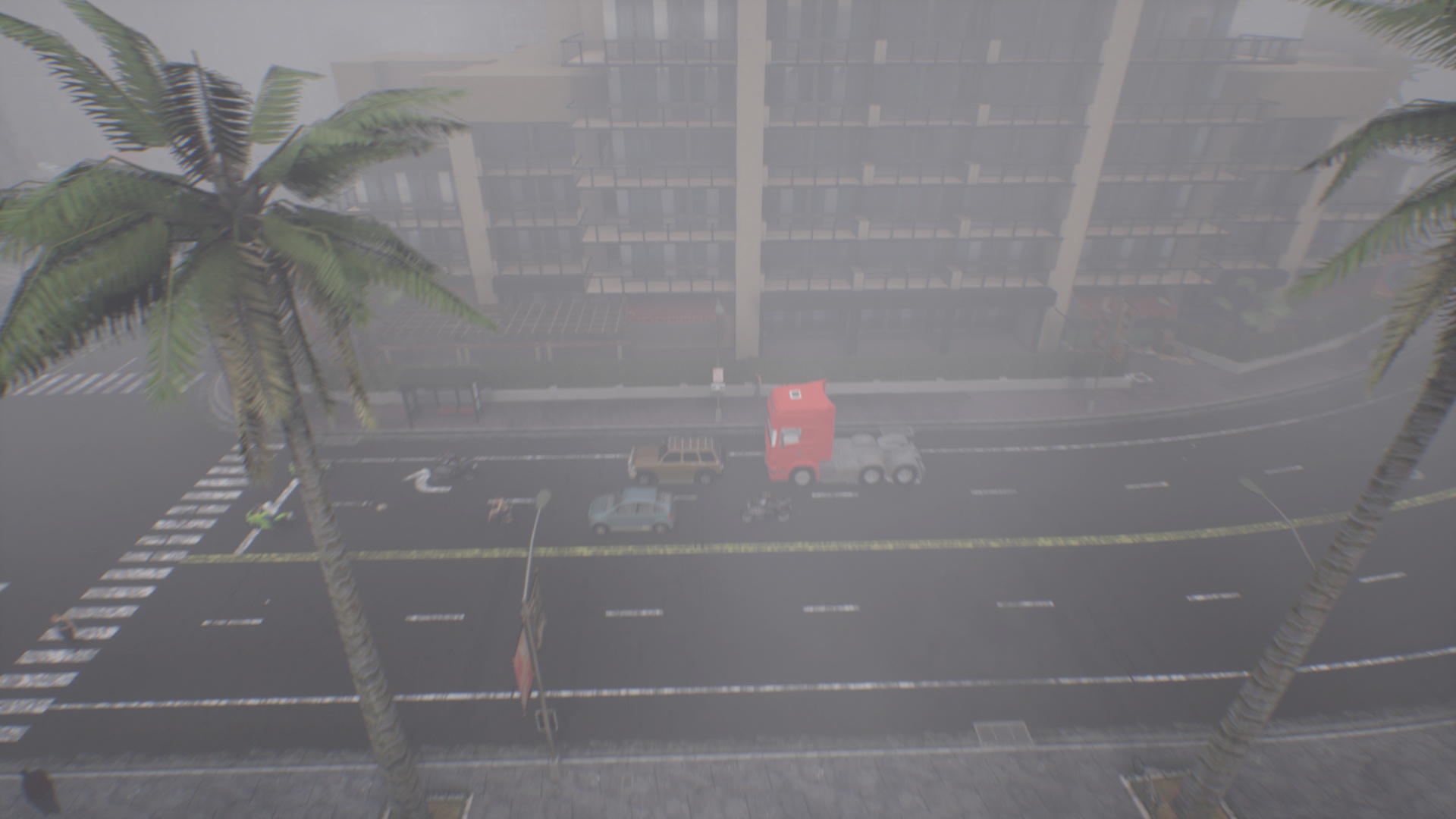}
        \end{subfigure}
    \end{subfigure}
    \begin{subfigure}{\textwidth}
        \begin{subfigure}{.245\textwidth}
            \centering
            \includegraphics[width=\textwidth]{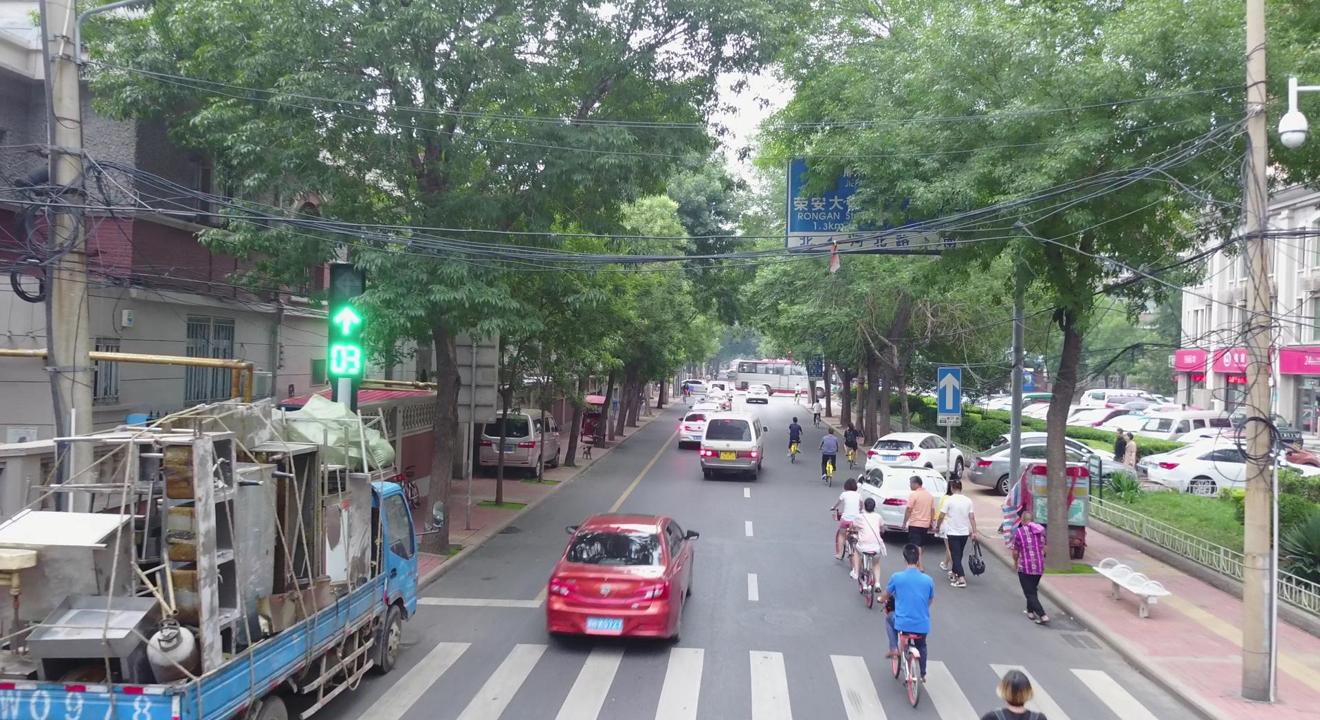}
        \end{subfigure}
        \begin{subfigure}{.245\textwidth}
            \centering
            \includegraphics[width=\textwidth]{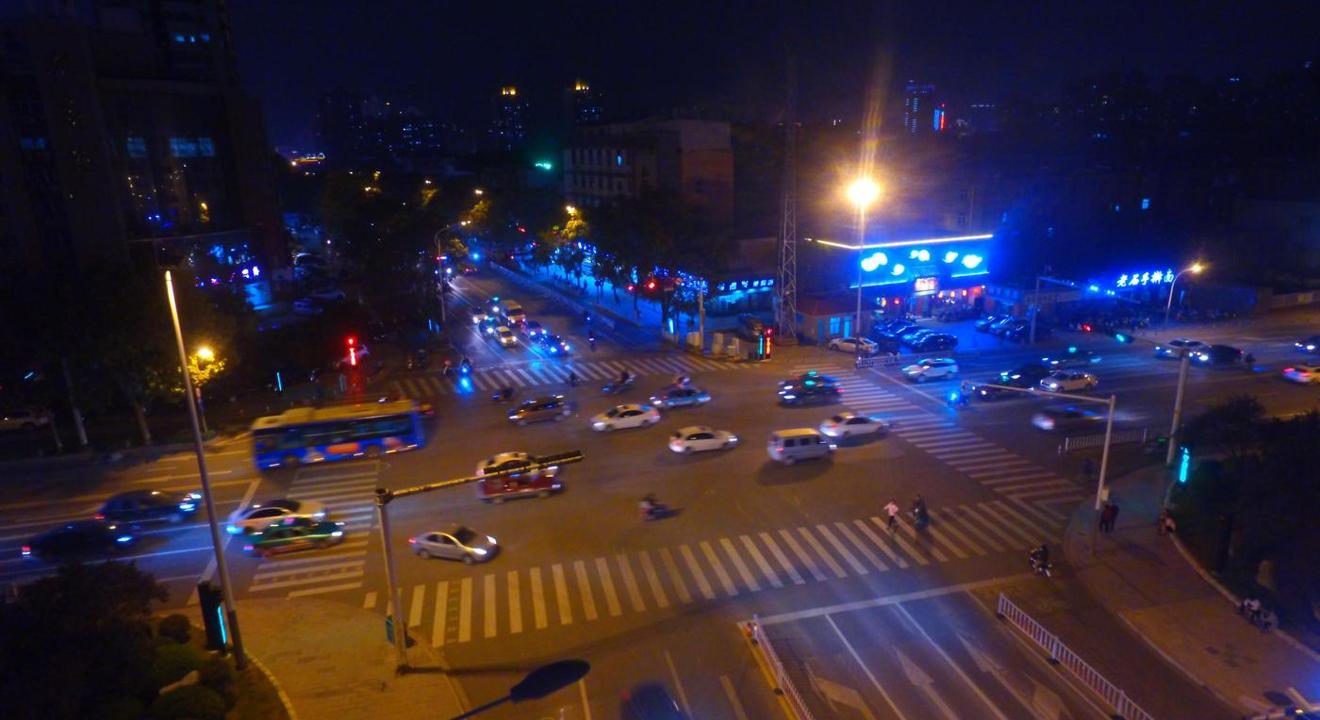}
        \end{subfigure}
        \begin{subfigure}{.245\textwidth}
            \centering
            \includegraphics[width=\textwidth]{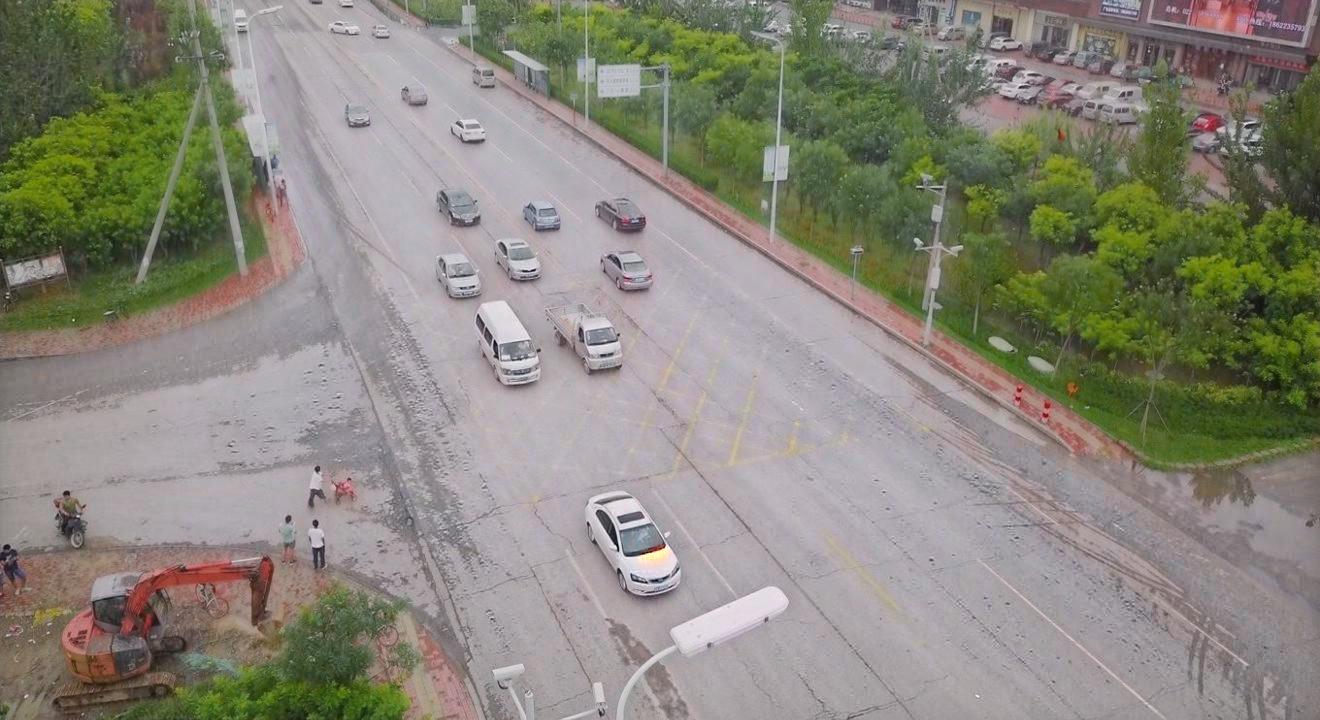}
        \end{subfigure}
        \begin{subfigure}{.245\textwidth}
            \centering
            \includegraphics[width=\textwidth]{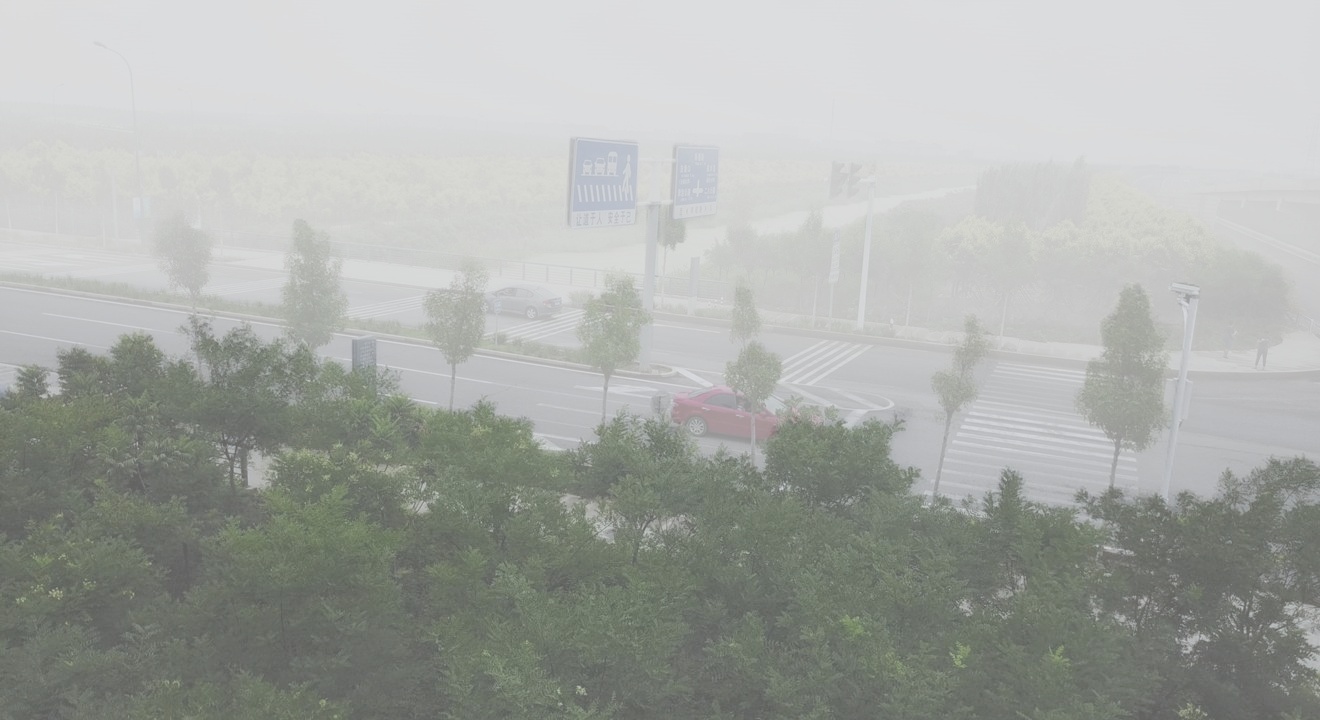}
        \end{subfigure}
    \end{subfigure}
    \caption{The FlyAwareV2 dataset provides data in variable weather and daytime conditions.}
    \label{fig:weathers}
\end{figure*}

The generation of various weather conditions for the synthetic dataset was achieved using the Unreal Engine (UE)~\cite{unrealengine} integrated within CARLA. By programmatically modifying the environmental configuration parameters, we produced photorealistic images under a wide range of adverse conditions. The physics-based rendering capabilities of UE ensure that these simulated weather effects closely approximate their real-world counterparts (see Figure \ref{fig:weathers} for some visual examples).
    
To further enhance data realism and diversity, we customized the CARLA source code at multiple levels. 
First, we extended and refined the predefined environmental settings by adjusting the atmospheric scattering parameters, fog density, and solar elevation and azimuth angles. These modifications increased both the diversity and the realism of lighting and visibility conditions. We also introduced new configurations, including an additional nighttime setting and a new weather profile, hard fog.

\begin{table}[h]
    \centering
    \begin{tabular}{c|c:c}
        Weather & Synthetic Config. & Real Data Aug. Strategy \\
        \hline
        Day & ClearNoon & \textit{Unchanged} \\
        Night & ClearNight & \texttt{img2img-turbo} \\
        Rain & HardRainNoon & \texttt{img2img-turbo + recolor} \\
        Fog & MidFoggyNoon & \texttt{FoHIS + recolor} \\
    \end{tabular}
    \caption{Weather configuration settings and strategies for  synthetic and real data.}
    \label{tab:config}
\end{table}

\subsection{Weather conditions for real data} \label{sub:augment:real_weather}
Collecting real images under adverse weather conditions proved to be extremely challenging. Operating a drone during heavy rain or dense fog is very challenging and impractical due to reduced visibility and flight instability. Consequently, such imagery is largely unavailable in the literature. Nevertheless, achieving consistency between the real and synthetic datasets was crucial for our study, particularly in terms of the inclusion of adverse weather scenarios.

After evaluating multiple alternatives, we determined that augmenting real clear-weather images to simulate different weather conditions was the only feasible solution.
Although the augmentation process varied slightly across weather types, in all cases we focused on preserving the structural realism and label alignment of the original images. A visual example of the final result is shown in the bottom row of Figure \ref{fig:weathers}.

\paragraph{\textbf{Fog}}
Simulating realistic fog needs to take into account that natural fog exhibits complex interactions with scene depth and light scattering.
We adopted an analytical approach that computes fog intensity per pixel using physics-based models. Specifically, we used the \texttt{FoHIS} algorithm~\cite{zhang2017towards} to generate depth-dependent fog effects. To tailor the method to our use case, we updated the original implementation and extended it with color manipulation functionality. This addition allowed us to reproduce the cooler, desaturated tones typically associated with foggy and winter conditions. 
We name this operation \texttt{recoloring}, as it consists of darkening and desaturation operations applied to the colorspace. More in detail, we first shift the white-point of the image $\mathbf{X}$ from the original warm-reddish
color to a toneless one, then we desaturate the image by computing the weighted average with the grayscale counterpart with weight $r_d = 0.7$, \ie, $\mathbf{X}_{\text{desat}} = r_s\mathbf{X} + (1-r_s)\mathbf{X}_{\text{gray}}$.
Finally, we darken the scene by re-scaling the RGB values by $r_l = 0.8$, \ie, $\mathbf{X}' = r_l\mathbf{X}_{\text{desat}}$.

The algorithm takes in input the RGB image, its corresponding depth map, and scene-level parameters such as the camera position and the visibility range. Depth maps were obtained as described in Sec.~\ref{sub:augment:depth_est}, and environment-specific profiles were manually defined for each dataset.

\paragraph{\textbf{Rain and night}}
Rain and night scenes cannot be reproduced analytically, as they are highly dependent on image-specific features such as light sources, reflections, and object materials. For example, generating a realistic night image requires simulating illuminated streetlights, active vehicle headlights, and lit windows in surrounding buildings.

Although diffusion-based generative models can produce visually convincing results, they often alter the structure of the scene, leading to inconsistencies between the generated images and ground-truth annotations. 
To avoid such distortions, we employed a U-Net~\cite{ronneberger2015u}-based model, \texttt{img2img-turbo}~\cite{parmar2024one}, which enables pixel-level transformations while preserving spatial and content integrity. Using this approach, we generated realistic rainy and nighttime variants of daytime images at the same resolution, that preserve perfect alignment with their original labels. For rainy images only, we have coupled img2img-turbo with the recoloring step described above to achieve better fidelity with the real counterpart.

It is important to note that for the test set, night images were synthetically generated from daytime counterparts, whereas for the training set, we included real nighttime images from existing datasets whenever available.

    \section{Experimental Evaluation} \label{sec:results}
We performed an extensive set of experiments using the FlyAwareV2 dataset to provide valuable insights into how it allows efficient training of deep learning models for multimodal semantic segmentation in urban environments tailored to UAV imagery. 
In this section, we start with the implementation details (Sec. \ref{sub:results:impl}), then we discuss the performances on synthetic data (Sec.~\ref{sub:results:exp_synth}). 
We continue analyzing how a model trained on the FlyAwareV2 synthetic data can perform on real-world data, firstly using it ``as is'' (Sec.~\ref{sub:results:exp_real}) and then employing also Unsupervised Domain Adaptation (UDA) techniques (Sec.~\ref{sub:results:exp_uda}).
Finally, we also discuss the performances of multimodal strategies that also exploit depth information (Sec.~\ref{sub:results:exp_mm}).

\subsection{Implementation Details} \label{sub:results:impl}
For the experimental evaluation, we employ an encoder–decoder architecture composed of a MobileNetV3+~\cite{howard2019searching} backbone integrated with a DeepLabV3~\cite{chen2017rethinking} decoder. This design choice, widely used in semantic segmentation literature, provides an effective balance between computational efficiency and segmentation accuracy. 

Model training is conducted using a single NVIDIA L40s GPU with a batch size of $16$ and full-HD images in input. Each training run spans $30$k iterations. For experiments involving multi-modal architectures, we utilize two L40s GPUs to cope with the memory requirements arising from the increased model complexity and input dimensionality. 
Optimization is carried out using the Adam algorithm, with an initial learning rate of $2.5 \times 10^{-4}$. The learning rate follows a cosine annealing schedule that decays to zero, preceded by a linear warm-up phase during the first $2000$ iterations. 

The data augmentation pipeline is designed to improve generalization by introducing controlled variations in image appearance and structure. 
Specifically, we apply random horizontal flipping with probability $p = 0.5$, as well as brightness, contrast, saturation, and hue jittering with a rate of $r = 0.5$, following the implementation in \cite{torchvision}. 
In addition, a Gaussian blur with $\sigma_b = 1.5$ and additive Gaussian noise with standard deviation $\sigma_n = 1.5$ are applied to simulate sensor noise and slight defocus. 
These enhancements collectively improve the resilience to illumination changes, color variability, and moderate image degradation.

\subsection{Synthetic Data Segmentation Experiments} \label{sub:results:exp_synth}
\begin{table*}[t]
\setlength{\tabcolsep}{.15em}
\begin{subtable}{.48\textwidth}
    \centering
    \begin{tabular}{c|cccc:c}
        \diagbox{Train}{Test} & Day & Night & Rain & Fog & All \\
        \hline
        Day   & \textbf{57.9} & \dpad1.7 & 11.8 & 1.9 & 17.0 \\
        Night & \dpad2.2 & \textbf{30.6} & \dpad1.7 & \dpad0.7 & \dpad8.4 \\
        Rain  & \dpad9.3 & \dpad1.7 & \textbf{53.5} & \dpad9.7 & \underline{17.1} \\
        Fog   & \dpad4.3 & \dpad0.8 & 18.2 & \underline{34.0} & 15.1 \\
        \hdashline
        All   & \underline{51.7} & \underline{29.8} & \underline{49.9} & \textbf{39.8} & \textbf{42.5}
    \end{tabular}
    \caption{mIoU on the full 28 classes set.}
    \label{tab:rgb_fine_synth_weather}
\end{subtable}
\begin{subtable}{.48\textwidth}
    \centering
    \begin{tabular}{c|cccc:c}
        \diagbox{Train}{Test} & Day & Night & Rain & Fog & All \\
        \hline
        Day   & \textbf{75.2} & 13.7 & 33.9 & 10.3 & 31.3 \\
        Night & 13.9 & \textbf{53.4} & \dpad8.8 & \dpad5.3 & 19.3 \\
        Rain  & 32.3 & \dpad9.5 & \textbf{72.3} & 27.8 & \underline{34.1} \\
        Fog   & 18.9 & 10.6 & 42.5 & \textbf{63.7} & 30.5 \\
        \hdashline
        All   & \underline{72.1} & \underline{55.0} & \underline{70.6} & \underline{61.6} & \textbf{64.5}
    \end{tabular}
    \caption{mIoU on the coarse 5 classes set.}
    \label{tab:rgb_coarse_synth_weather}
\end{subtable}
\caption{Training and testing on synthetic data with varying weather conditions in the train and test sets.}
\label{tab:rgb_synth_weather}
\end{table*}

\begin{figure*}[tbh]
    \centering
    \begin{subfigure}{\textwidth}
        \centering
        \begin{subfigure}{.24\textwidth}
            \centering
            \caption*{Day}
            \includegraphics[width=\textwidth]{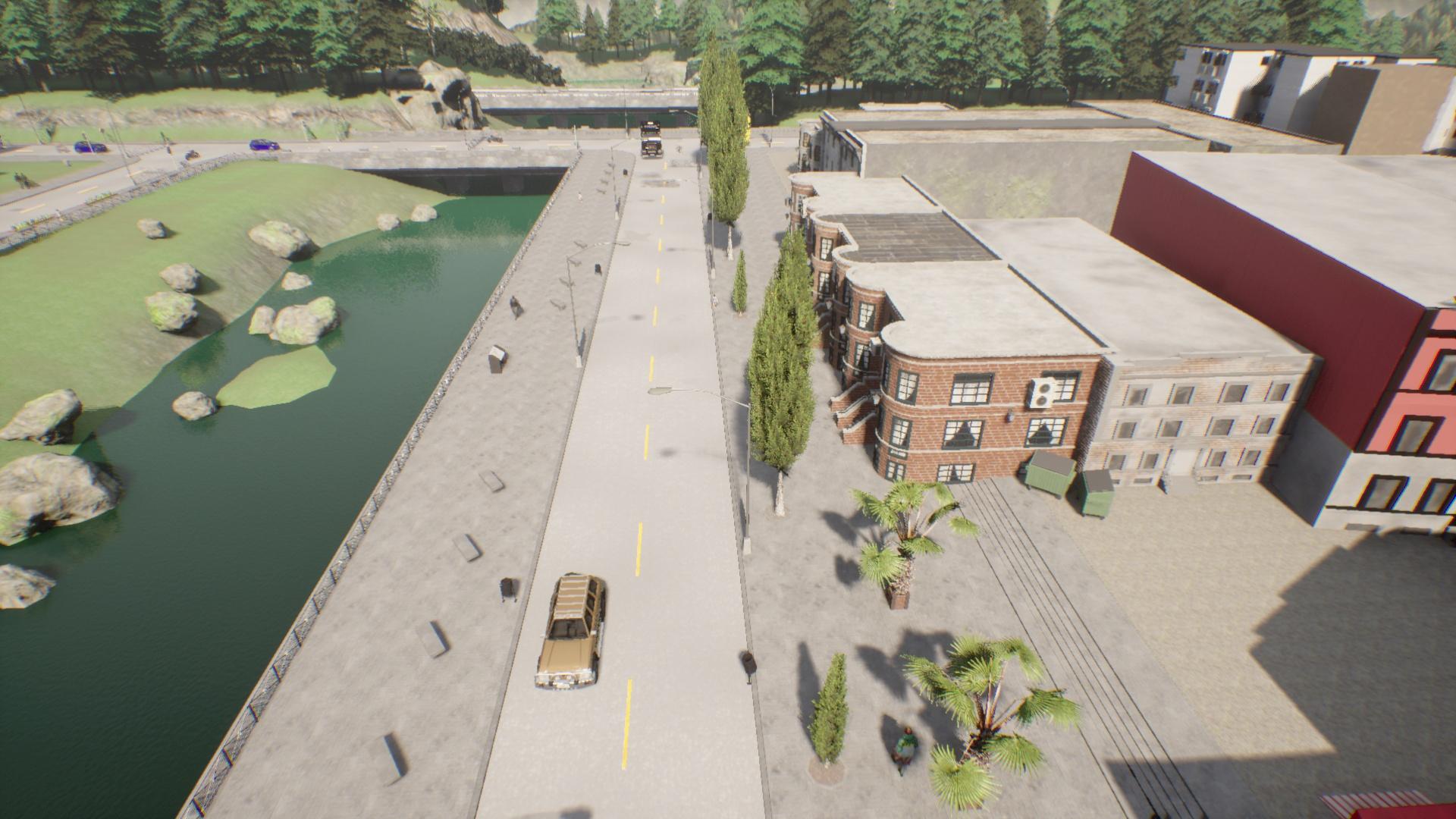}
        \end{subfigure}
        \begin{subfigure}{.24\textwidth}
            \centering
            \caption*{Night}
            \includegraphics[width=\textwidth]{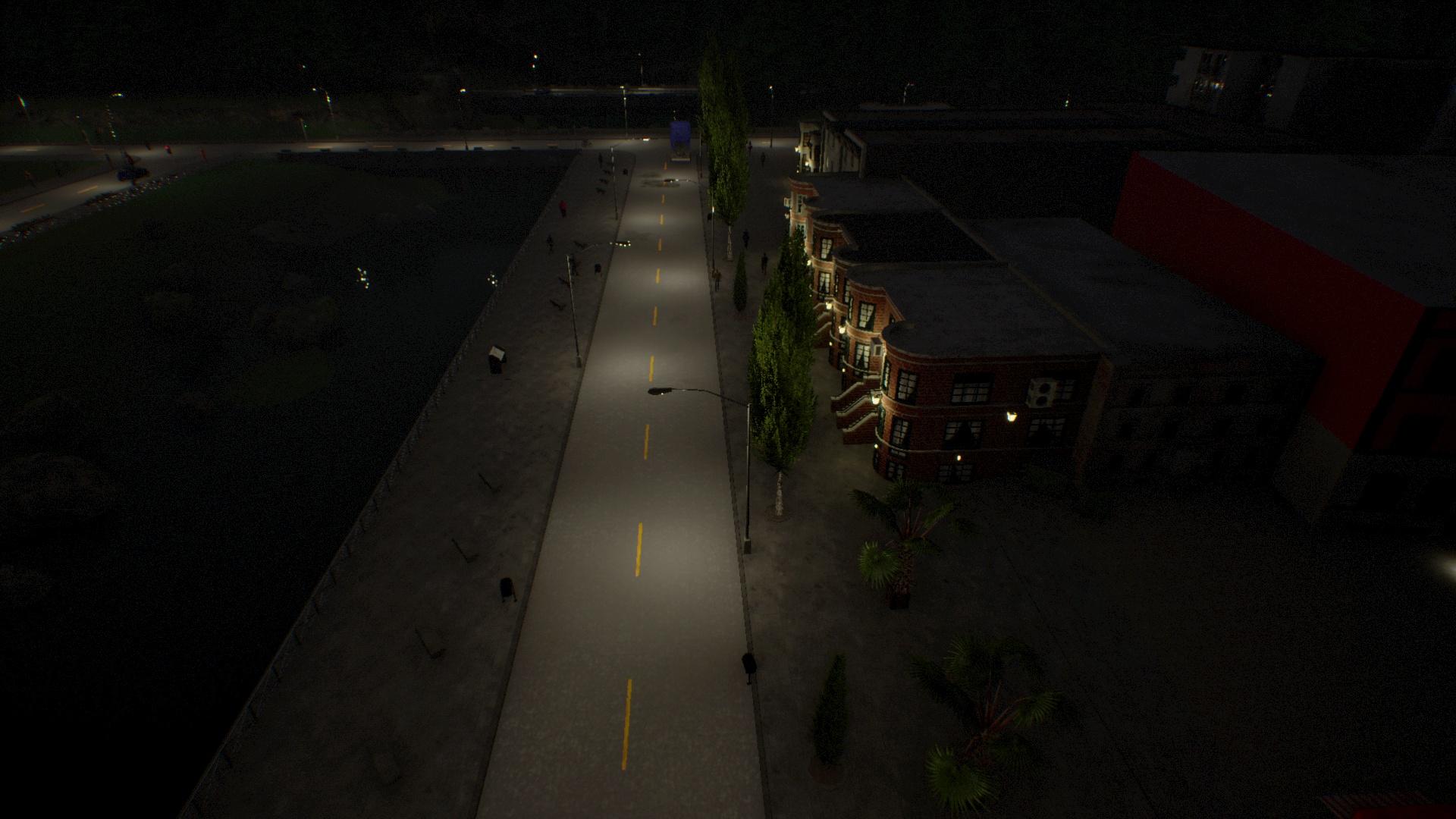}
        \end{subfigure}
        \begin{subfigure}{.24\textwidth}
            \centering
            \caption*{Rain}
            \includegraphics[width=\textwidth]{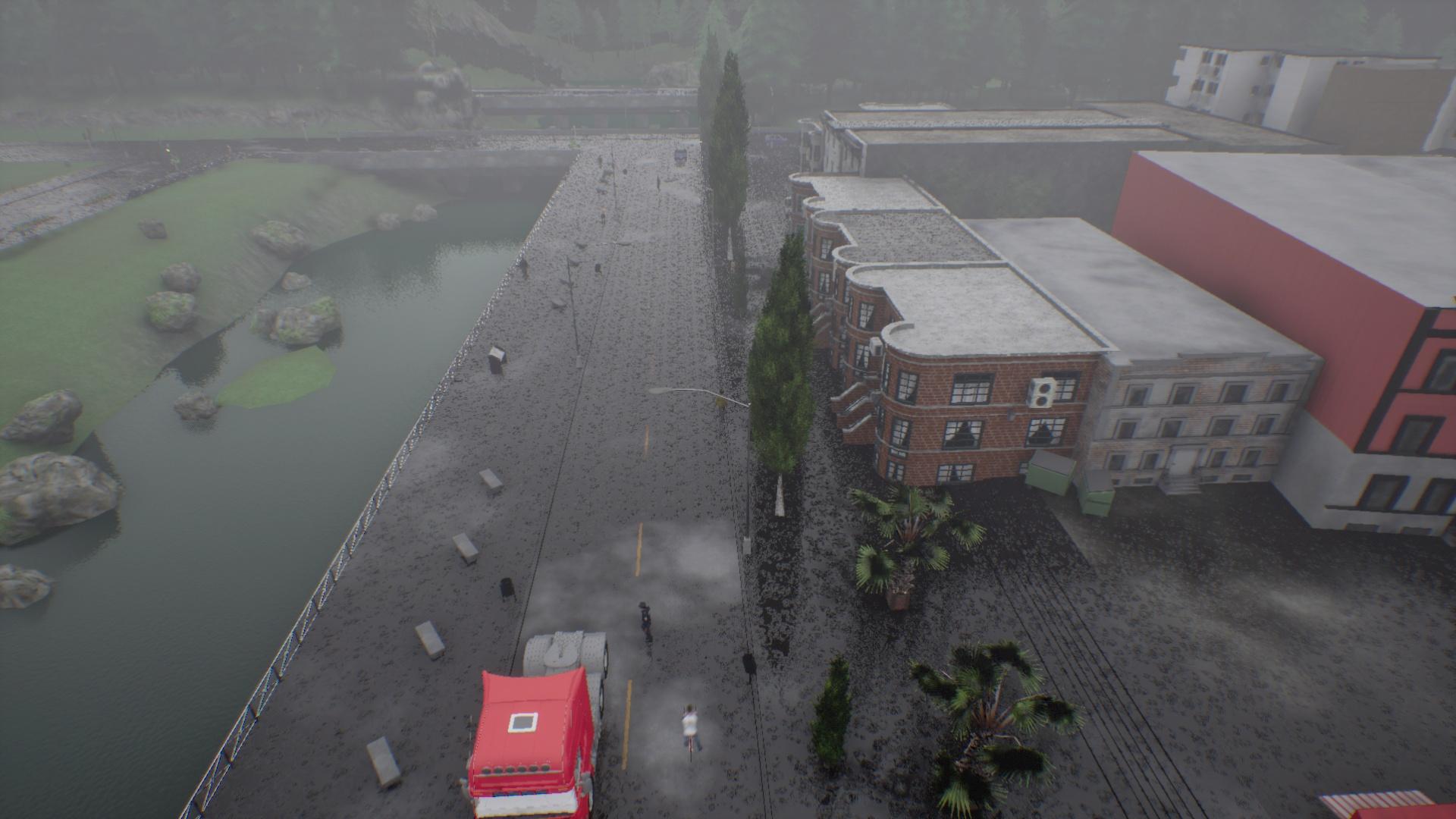}
        \end{subfigure}
        \begin{subfigure}{.24\textwidth}
            \centering
            \caption*{Fog}
            \includegraphics[width=\textwidth]{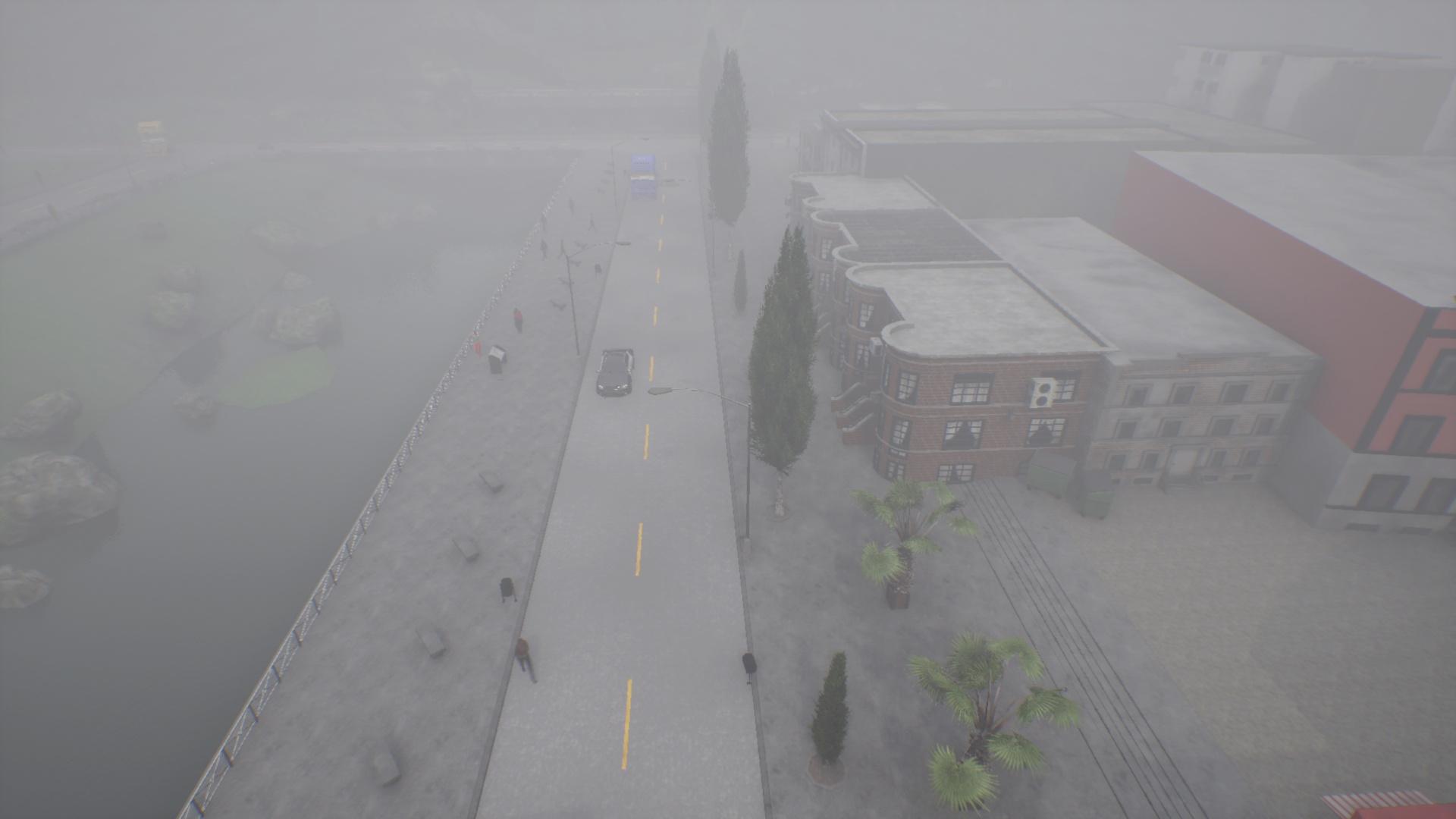}
        \end{subfigure}
    \end{subfigure}
    \begin{subfigure}{\textwidth}
        \centering
        \begin{subfigure}{.24\textwidth}
            \centering
            \includegraphics[width=\textwidth]{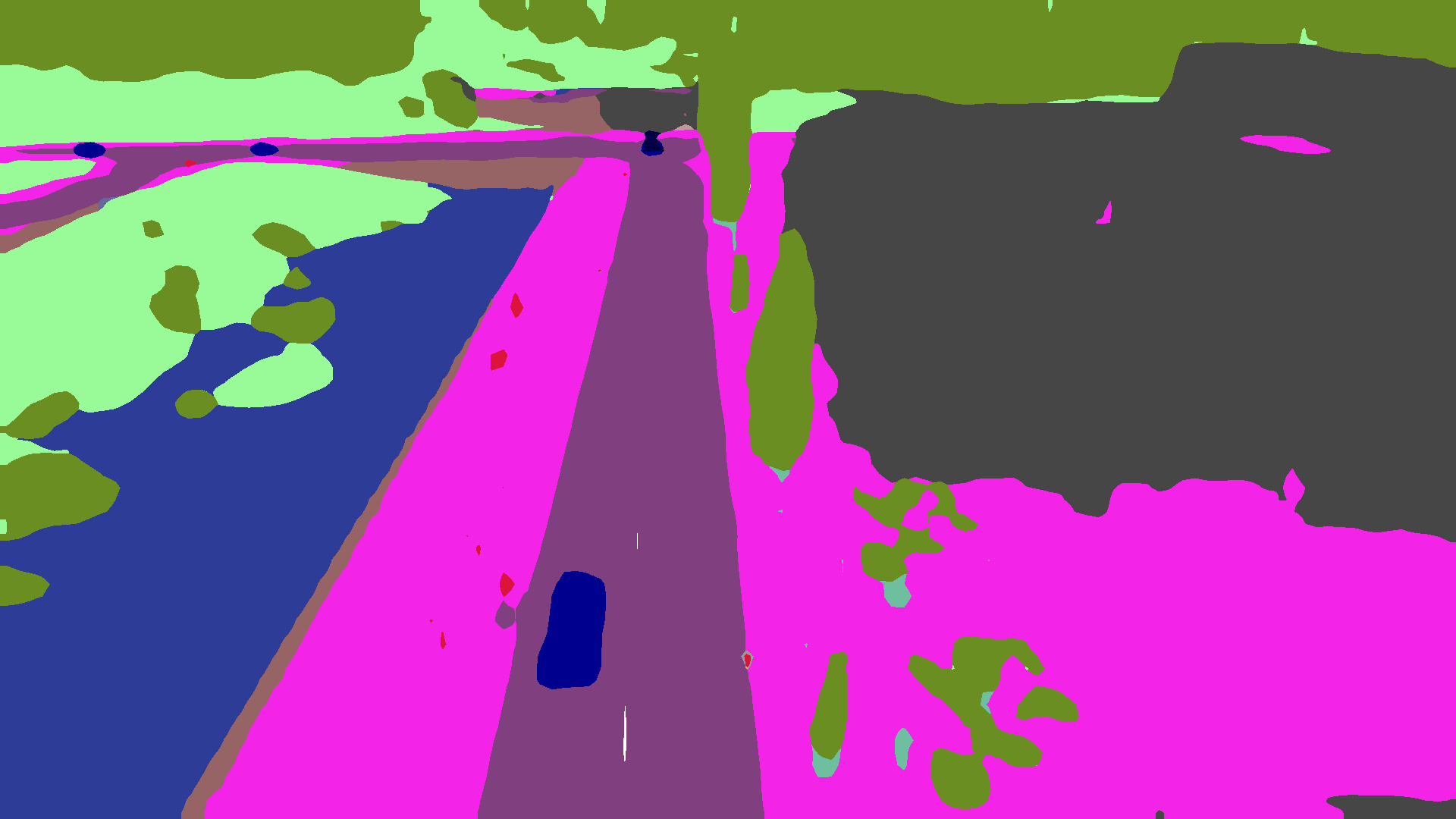}
        \end{subfigure}
        \begin{subfigure}{.24\textwidth}
            \centering
            \includegraphics[width=\textwidth]{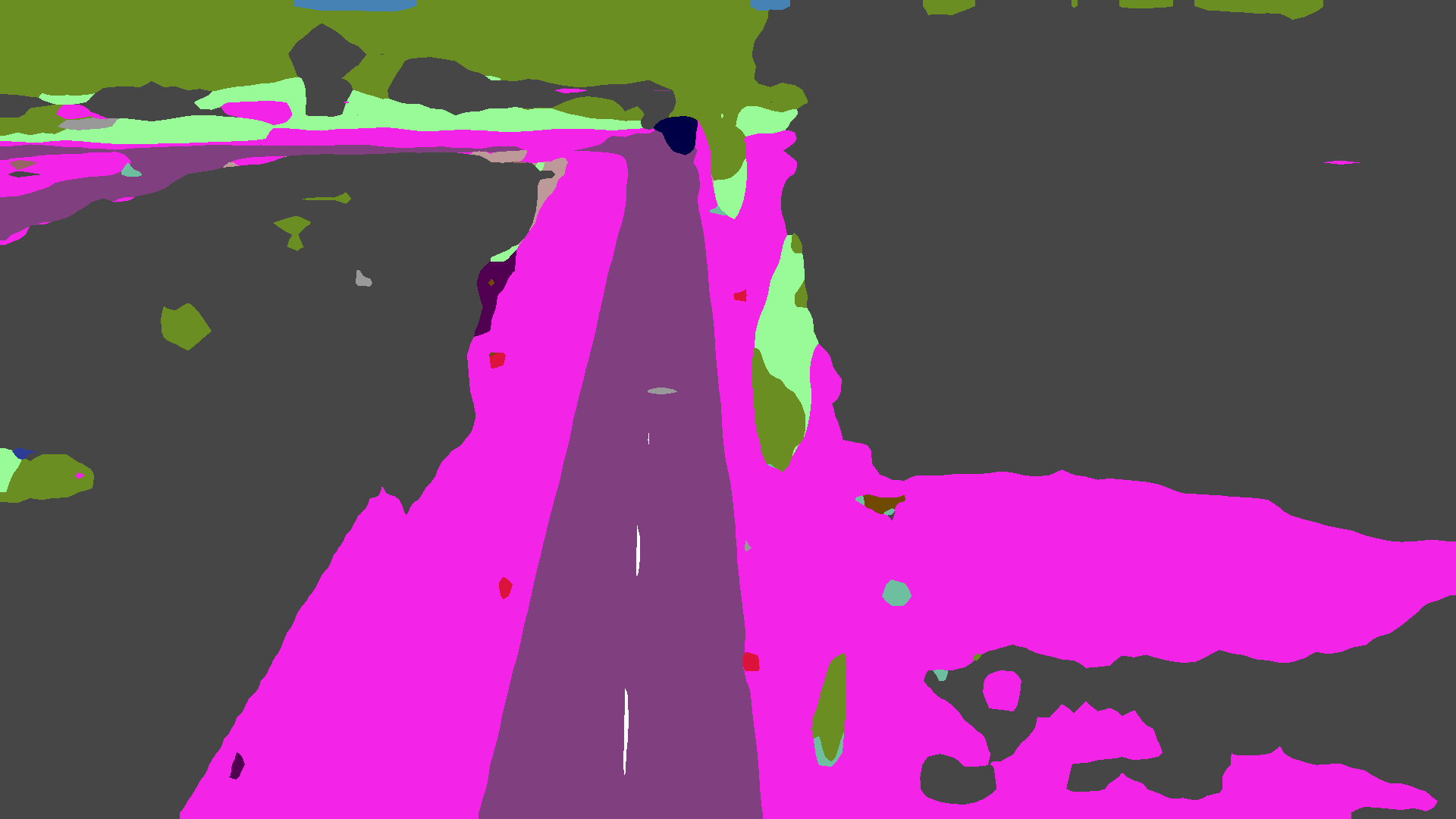}
        \end{subfigure}
        \begin{subfigure}{.24\textwidth}
            \centering
            \includegraphics[width=\textwidth]{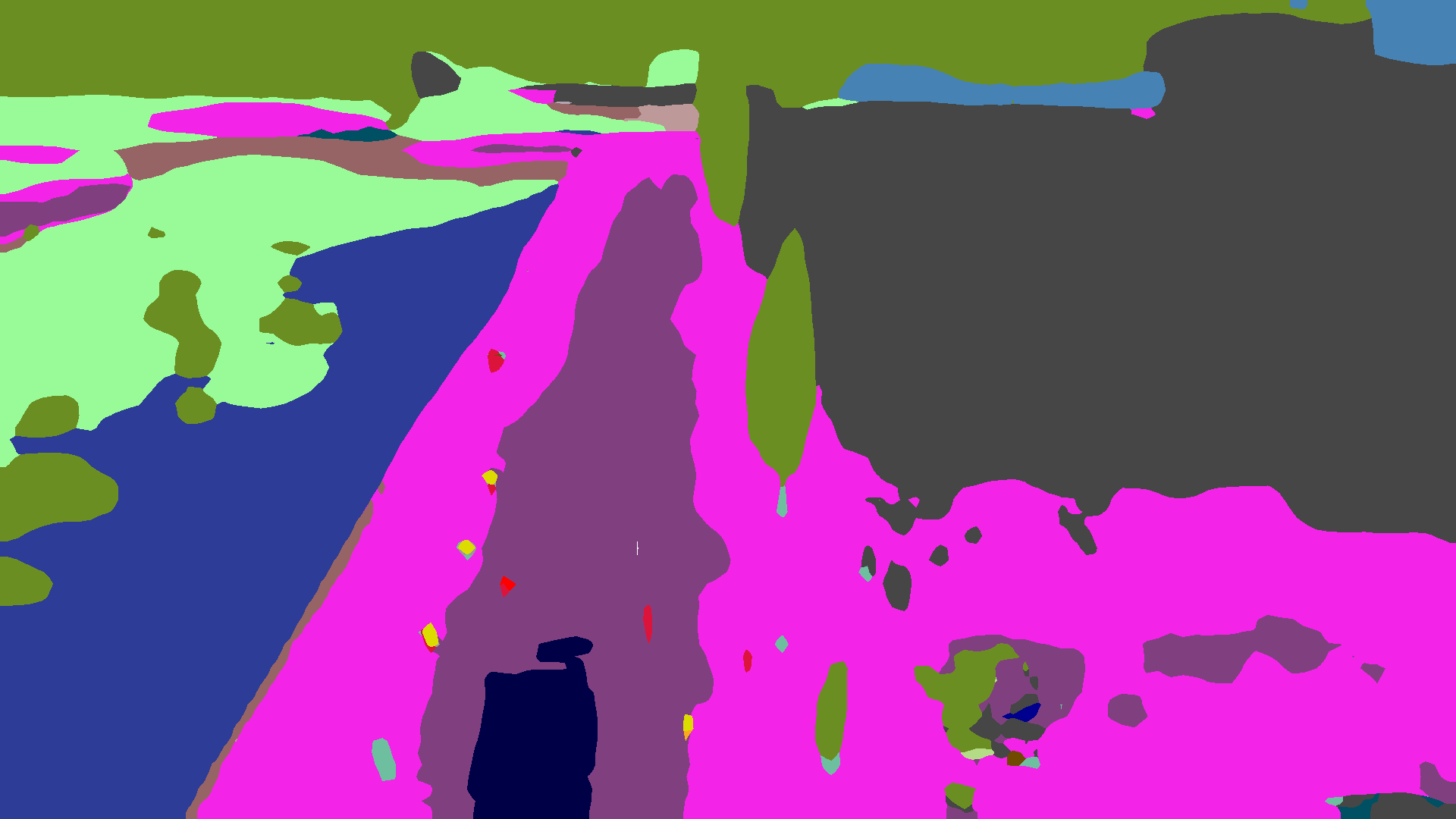}
        \end{subfigure}
        \begin{subfigure}{.24\textwidth}
            \centering
            \includegraphics[width=\textwidth]{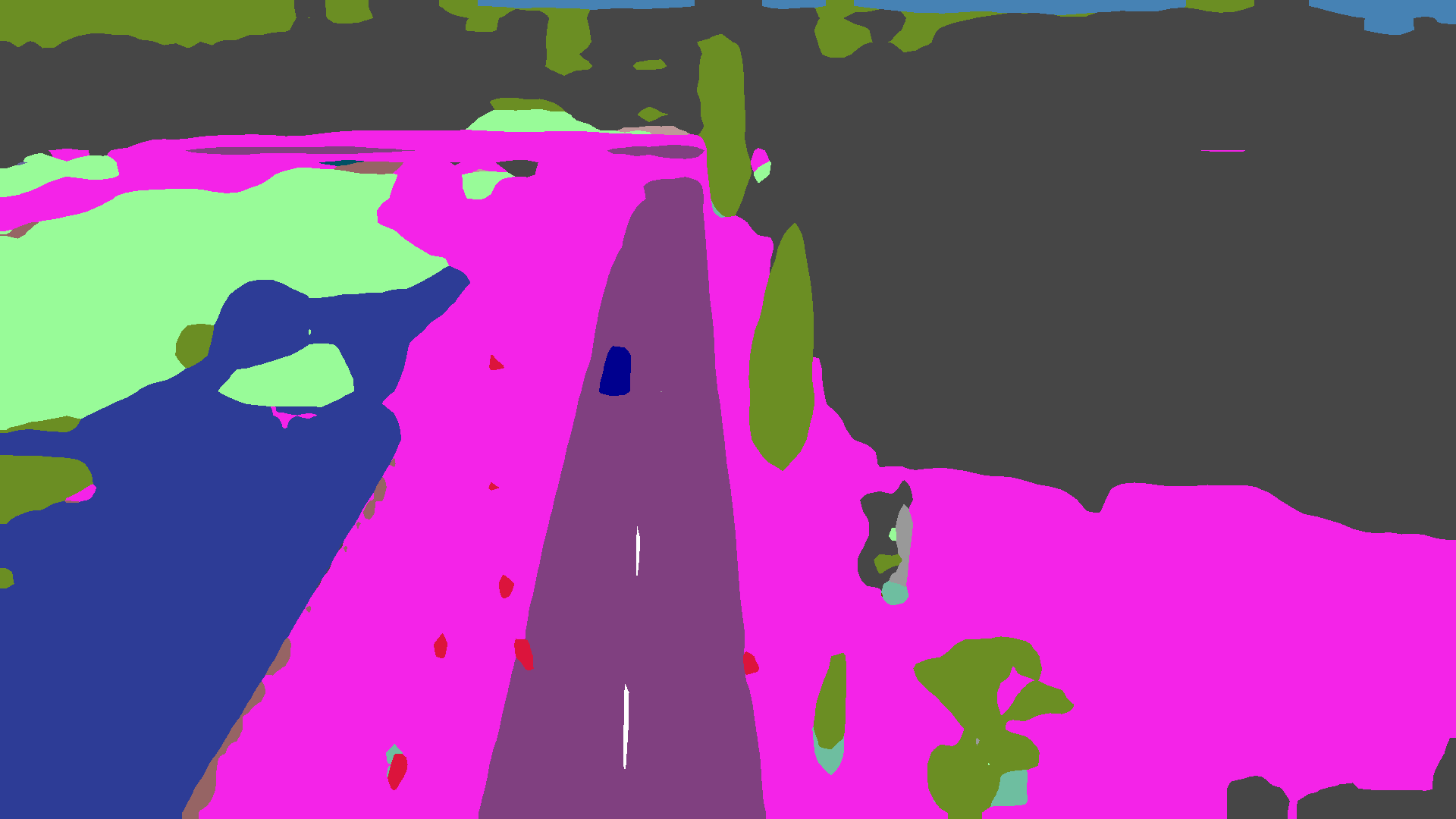}
        \end{subfigure}
    \end{subfigure}
    \begin{subfigure}{\textwidth}
        \centering
        \begin{subfigure}{.24\textwidth}
            \centering
            \includegraphics[width=\textwidth]{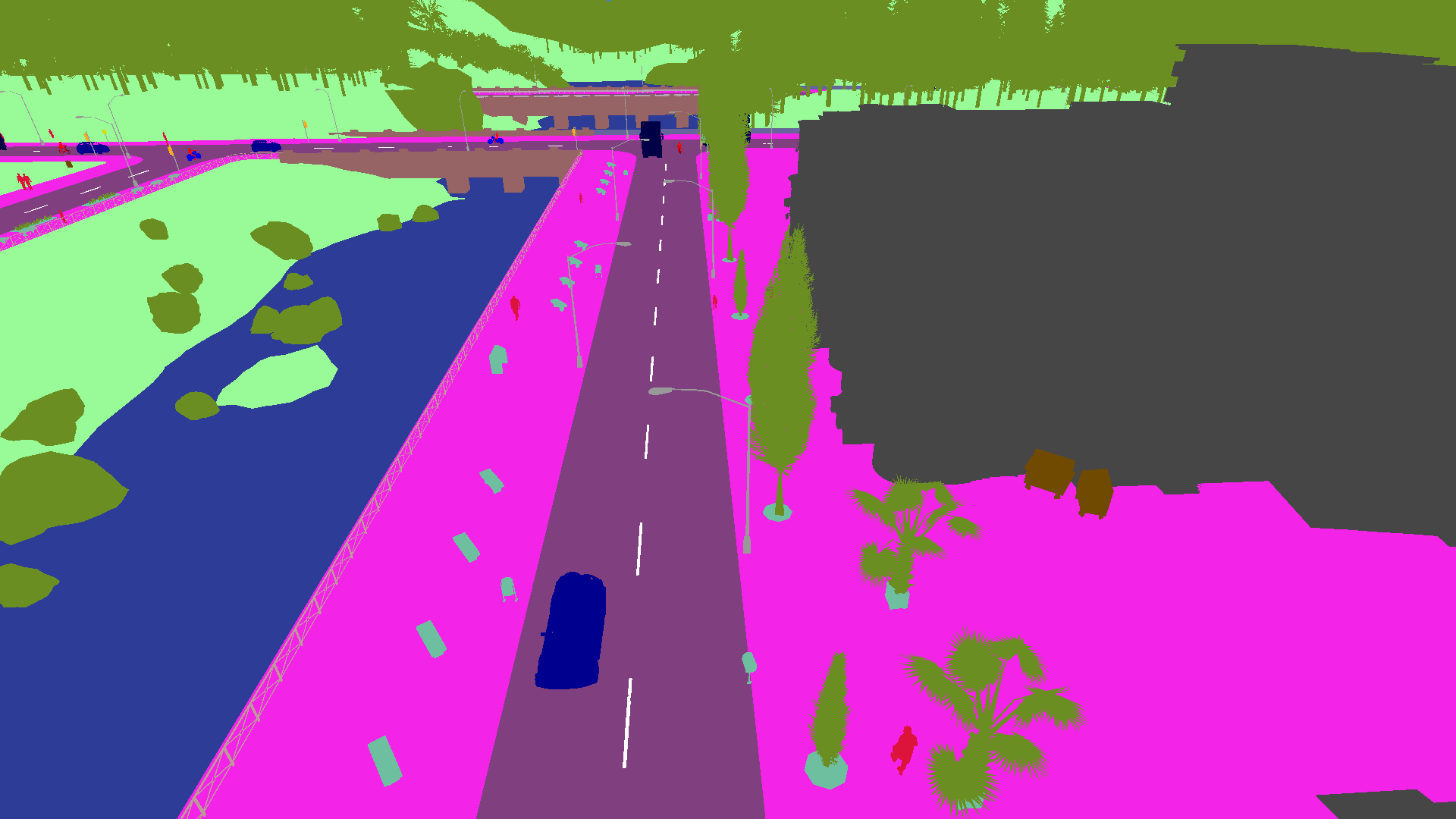}
        \end{subfigure}
        \begin{subfigure}{.24\textwidth}
            \centering
            \includegraphics[width=\textwidth]{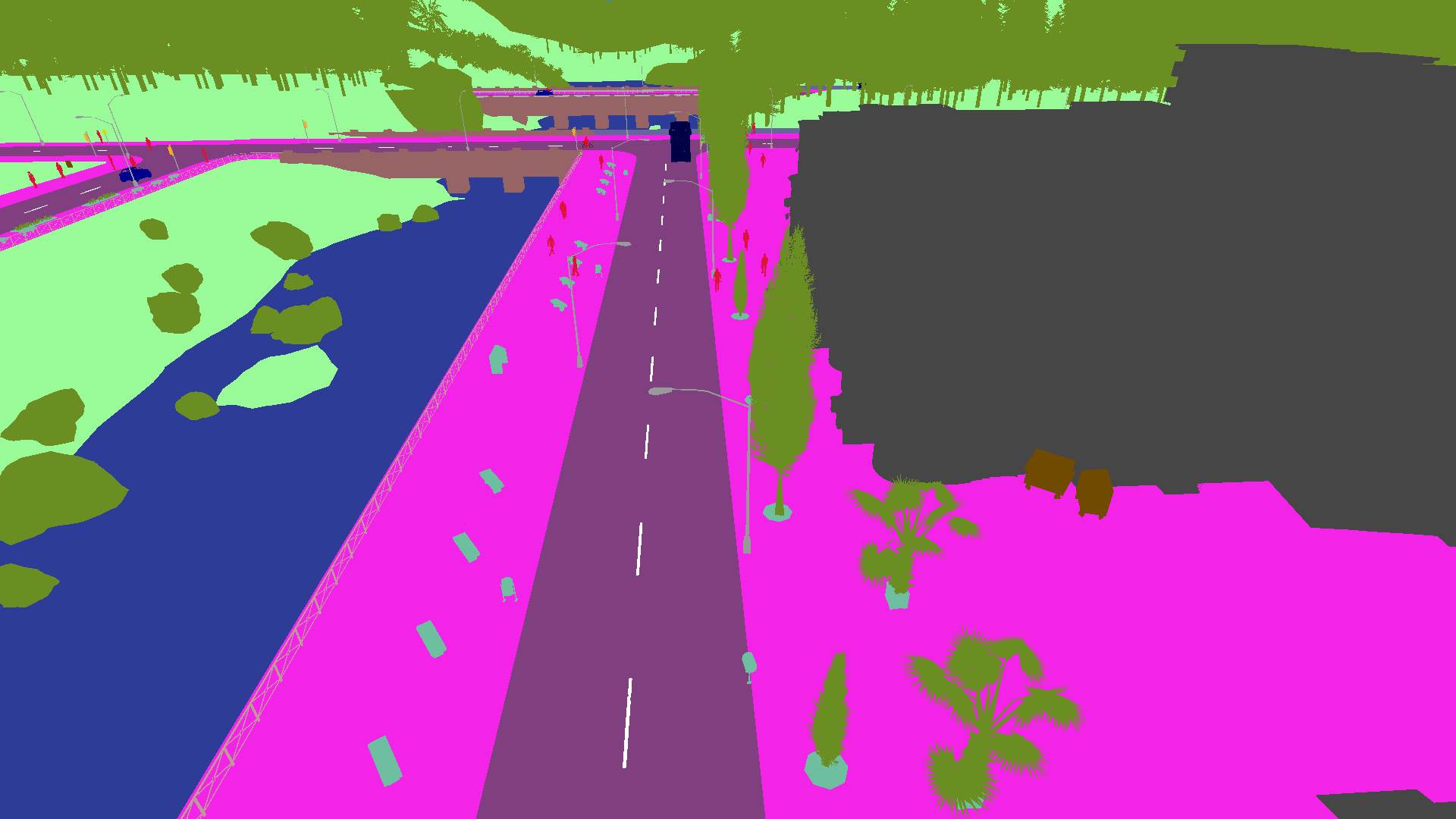}
        \end{subfigure}
        \begin{subfigure}{.24\textwidth}
            \centering
            \includegraphics[width=\textwidth]{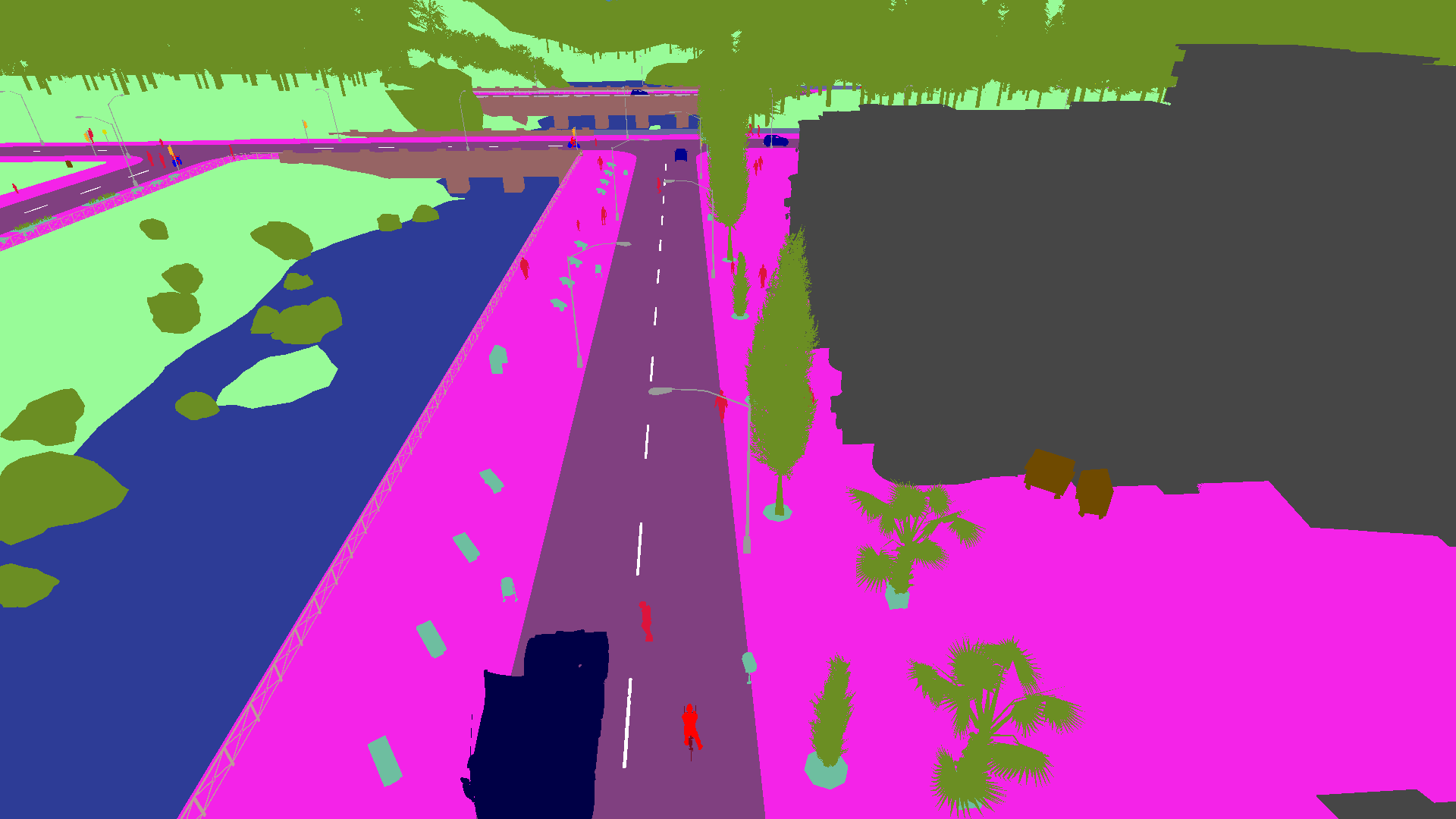}
        \end{subfigure}
        \begin{subfigure}{.24\textwidth}
            \centering
            \includegraphics[width=\textwidth]{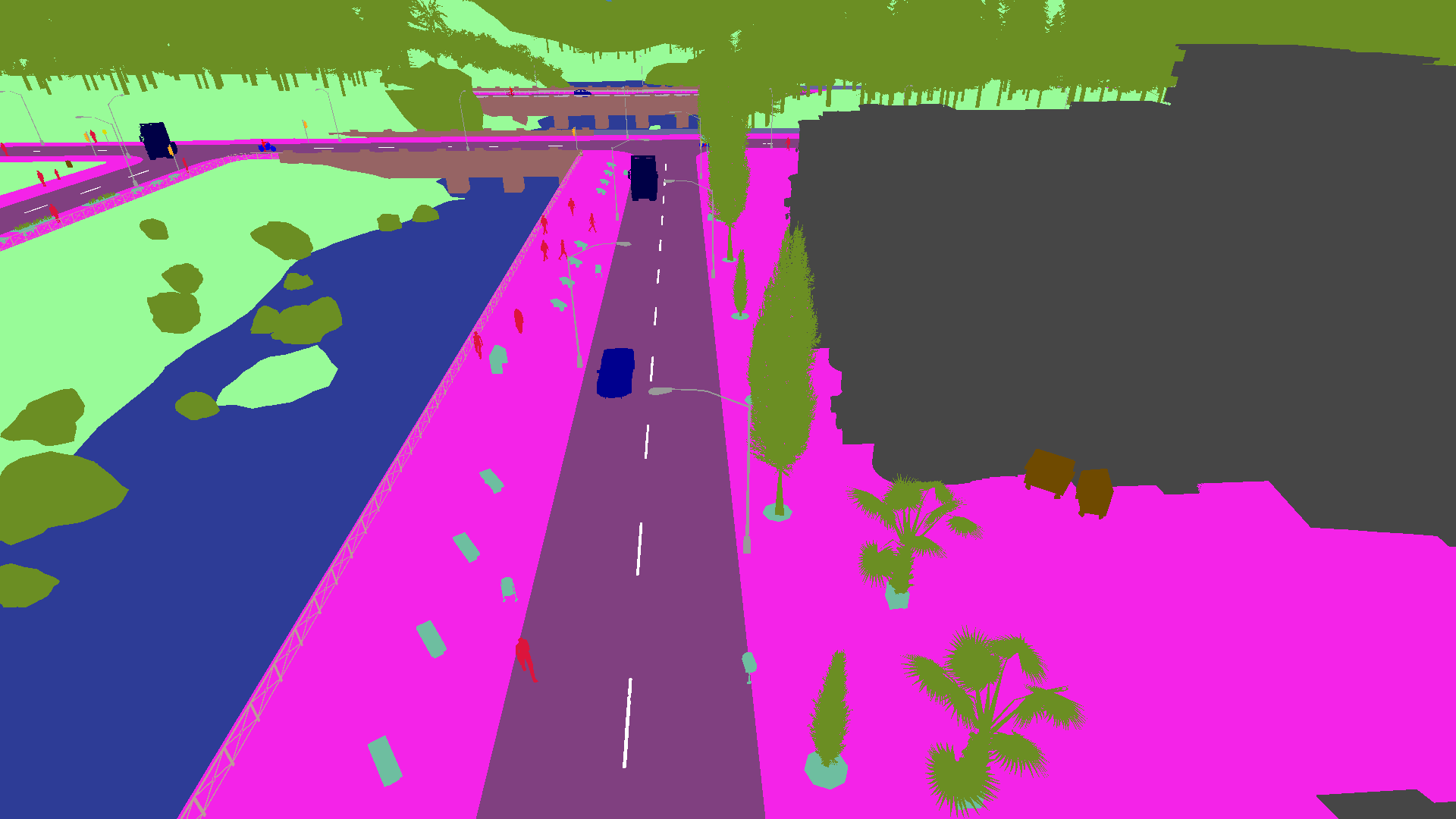}
        \end{subfigure}
    \end{subfigure}
    \caption{Qualitative results: model trained on synthetic samples and tested on synthetic samples with varying weather conditions. First row: Input, Second row: Model Prediction, Third Row: Ground Truth. }
    \label{fig:quali_synth_weathers}
\end{figure*}

We start by training the network on the synthetic data in FlyAwareV2 and evaluating the performance of the network on the synthetic test set, \ie, on the same domain used for training. 
The results are reported in Table \ref{tab:rgb_synth_weather} and Figure \ref{fig:quali_synth_weathers}. %
Table \ref{tab:rgb_coarse_synth_weather} reports the results on the reduced class set used by the synthetic-to-real experiments; details of the mapping are listed in Table \ref{tab:cmap}.

As a starting point, Table \ref{tab:rgb_fine_synth_weather} shows the performance on the full set of 28 classes. Using the model trained on all the training data (of all weathers) and evaluated on all testing data leads to a mIoU of $42.5\%$. 
From the table, it can be seen that the daytime data is easier ($51.7\%$), while the night and fog settings proved to be more challenging, as expected (29.8\% and 39.8\%).
Training only on a specific weather condition leads to overfitting that specific setting, with improved performance on the chosen setting at the price of strong degradation on the others (in many cases, the model simply does not work, leading to accuracies below $2\%$).

A similar situation is noticeable in Figure \ref{fig:quali_synth_weathers}, the daytime prediction is clearly of higher accuracy with respect to all other weather conditions, while fog and rain lag behind.
The nighttime prediction is the worst, with the network confusing the water in the river for a building and completely missing the vegetation on the sidewalk.

Using the coarser class set (Table \ref{tab:rgb_coarse_synth_weather}), the task becomes easier and the accuracy improves. The all-weather experiment leads to an accuracy of $64.5\%$, but the general trend remains the same. The only difference is that, in settings where train and test data do not match, performance is low but not unacceptable as before.
This is probably because \eg, fog or rain makes it particularly difficult to recognize challenging small classes, while larger and simpler things like the road or a building can still be recognized.

\begin{table}[t]
    \centering
    \newcommand{\rotangle}{30}
    \renewcommand{\arraystretch}{.9}
    \resizebox{\textwidth}{!}{%
    \begin{tabular}{c|W{c}{2em}W{c}{2em}W{c}{2em}W{c}{2em}W{c}{2em}W{c}{2em}W{c}{2em}W{c}{4em}:c}
        \diagbox{Train}{Test} & \rotatebox[origin=lb]{\rotangle}{Town01(25)} & \rotatebox[origin=lb]{\rotangle}{Town02(21)} & \rotatebox[origin=lb]{\rotangle}{Town03(26)} & \rotatebox[origin=lb]{\rotangle}{Town04(25)} & \rotatebox[origin=lb]{\rotangle}{Town05(27)} & \rotatebox[origin=lb]{\rotangle}{Town06(24)} & \rotatebox[origin=lb]{\rotangle}{Town07(24)} & \rotatebox[origin=lb]{\rotangle}{T.10HD(26)} & All \\
        \hline
        Town01    & \textbf{38.7} & 14.9 & \dpad6.1 & \dpad9.9 & \dpad7.6 & 10.4 & 10.3 & 5.3 & 13.5 \\
        Town02   & 15.2 & \textbf{45.1} & \dpad5.5 & \dpad6.8 & \dpad6.9 & \dpad6.2 & \dpad7.5 & \dpad5.1 & 10.6 \\
        Town03   & \dpad8.8 & \dpad8.8 & \textbf{50.7} & 12.1 & 14.5 & 13.8 & \dpad7.3 & 10.0 & \underline{18.6} \\
        Town04   & 10.8 & \dpad9.0 & 11.4 & \textbf{37.3} & 16.0 & 14.6 & 10.9 & \dpad6.2 & 16.1 \\
        Town05   & \dpad7.5 & \dpad7.5 & 14.7 & 14.1 & \textbf{37.8} & 12.1 & \dpad8.9 & \dpad8.5 & 14.9 \\
        Town06   & \dpad8.1 & \dpad6.0 & 10.2 & 11.3 & 11.8 & \textbf{33.2} & \dpad9.3 & \dpad6.0 & 10.9 \\
        Town07   & \dpad7.3 & \dpad5.2 & \dpad7.4 & 10.0 & \dpad9.8 & 12.6 & \textbf{40.1} & \dpad3.4 & 10.3 \\
        Town10HD & \dpad7.5 & \dpad7.8 & \dpad9.3 & \dpad6.7 & \dpad7.5 & \dpad4.2 & \dpad3.5 & \textbf{39.9} & 12.2 \\
        \hdashline
        All      & \underline{30.9} & \underline{27.5} & \underline{39.0} & \underline{31.9} & \underline{33.8} & \underline{25.0} & \underline{27.3} & \underline{26.6} & \textbf{42.5} \\[.1em]
        \hline\vspace*{.3em}
        All* (town cl.)     & 34.6 & 36.7 & 42.0 & 35.7 & 35.1 & 29.2 & 31.9 & 28.6 & 42.5
    \end{tabular}}
    \caption{mIoU varying training and testing towns. The results are on the full 28 classes set. Note how not all classes are present in all towns (the number of classes per town is in parentheses close to the town name). \\
    *: The mIoU computed only on the classes present in each town is reported in the last row.}
    \label{tab:rgb_synth_towns}
\end{table}

Then, in Table \ref{tab:rgb_synth_towns} we analyze the impact of training or testing in different environments (in this case, the different \textit{Towns} of the CARLA simulator). A model trained on all cities can generalize quite well to the various environments corresponding to the different towns. Training on a single town instead leads to a model that is not able to generalize well due to the different appearance (some classes are unavailable in certain towns, making it impossible to train there). 

\begin{wraptable}{r}{.45\textwidth}
    \setlength{\tabcolsep}{.3em}
    \setlength{\belowcaptionskip}{-1em}
    \centering
    \begin{tabular}{c|ccc:c}
        \diagbox{Train}{Test} & 20m & 50m & 80m & All \\
        \hline
        20m   & \textbf{48.6} & 21.5 & 10.7 & 28.5 \\
        50m   & 21.6 & \textbf{44.4} & 31.0 & \underline{31.0} \\
        80m   & \dpad9.6 & 30.0 & \textbf{44.0} & 25.4 \\
        \hdashline
        All   & \underline{42.9} & \underline{41.4} & \underline{40.5} & \textbf{42.5}
    \end{tabular}
    \caption{mIoU varying training and testing height (using synthetic data from all weathers and towns for both training and testing).}
    \label{tab:rgb_synth_height}
\end{wraptable}

Finally, we focus on the impact of flying height. Table \ref{tab:rgb_synth_height} shows how a model trained on drones flying at different heights is able to generalize well to this aspect, obtaining a stable accuracy ranging between around $40\%$ and $43\%$ (as expected, higher heights are slightly more challenging since objects appear smaller). However, training using only data at a specific height leads to weaker models that are not able to generalize well to a change of viewpoint. This is consistent with the results found for weather and towns, confirming the validity of heterogeneous data.

\subsection{Real world evaluation of models trained on synthetic data.} \label{sub:results:exp_real}

\begin{table}[b]
    \centering
    \begin{tabular}{c|cccc:c}
        \diagbox{Train}{Test} & Day & Night & Rain & Fog & All \\
        \hline
        Day   & \underline{48.7} & 21.3 & 37.6 & 12.9 & \underline{33.8} \\
        Night & 12.7 & \underline{22.7} & \dpad7.4 & \dpad8.6 & 13.4 \\
        Rain  & 38.5 & 17.7 & \underline{45.0} & 18.4 & 31.4 \\
        Fog   & 28.6 & 14.7 & 32.7 & \underline{19.7} & 24.7 \\
        \hdashline
        All   & \textbf{49.7} & \textbf{23.4} & \textbf{48.4} & \textbf{38.4} & \textbf{42.3}
    \end{tabular}
    \caption{Synthetic-to-Real adaptation across weather conditions, RGB only, Coarse class-set.}
    \label{tab:rgb_real_weather}
\end{table}

\begin{figure*}[t]
    \centering
    \begin{subfigure}{\textwidth}
        \centering
        \begin{subfigure}{.24\textwidth}
            \centering
            \caption*{Day}
            \includegraphics[trim={0 1.3cm 0 1.3cm},clip,width=\textwidth]{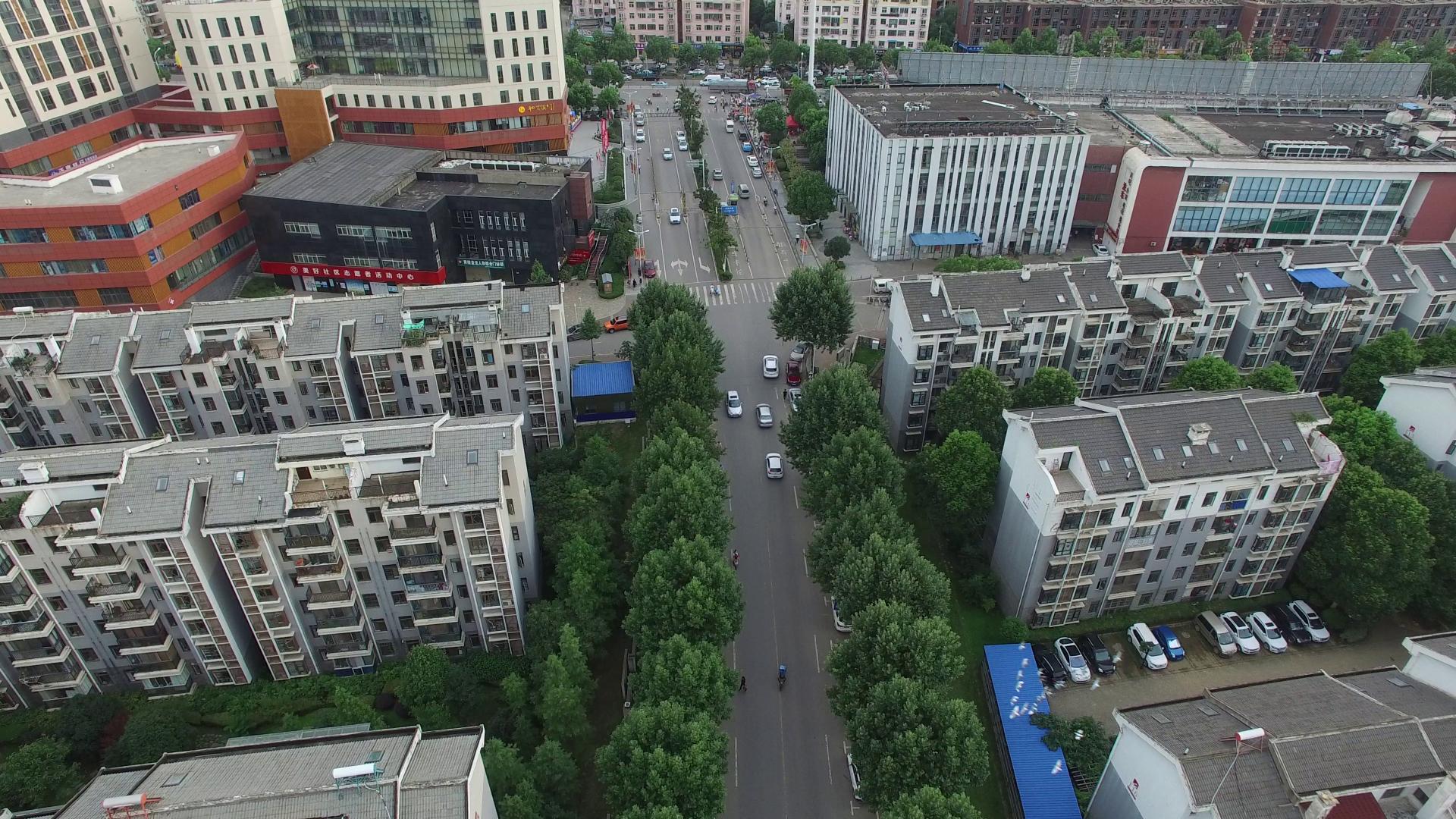}
        \end{subfigure}
        \begin{subfigure}{.24\textwidth}
            \centering
            \caption*{Night}
            \includegraphics[width=\textwidth]{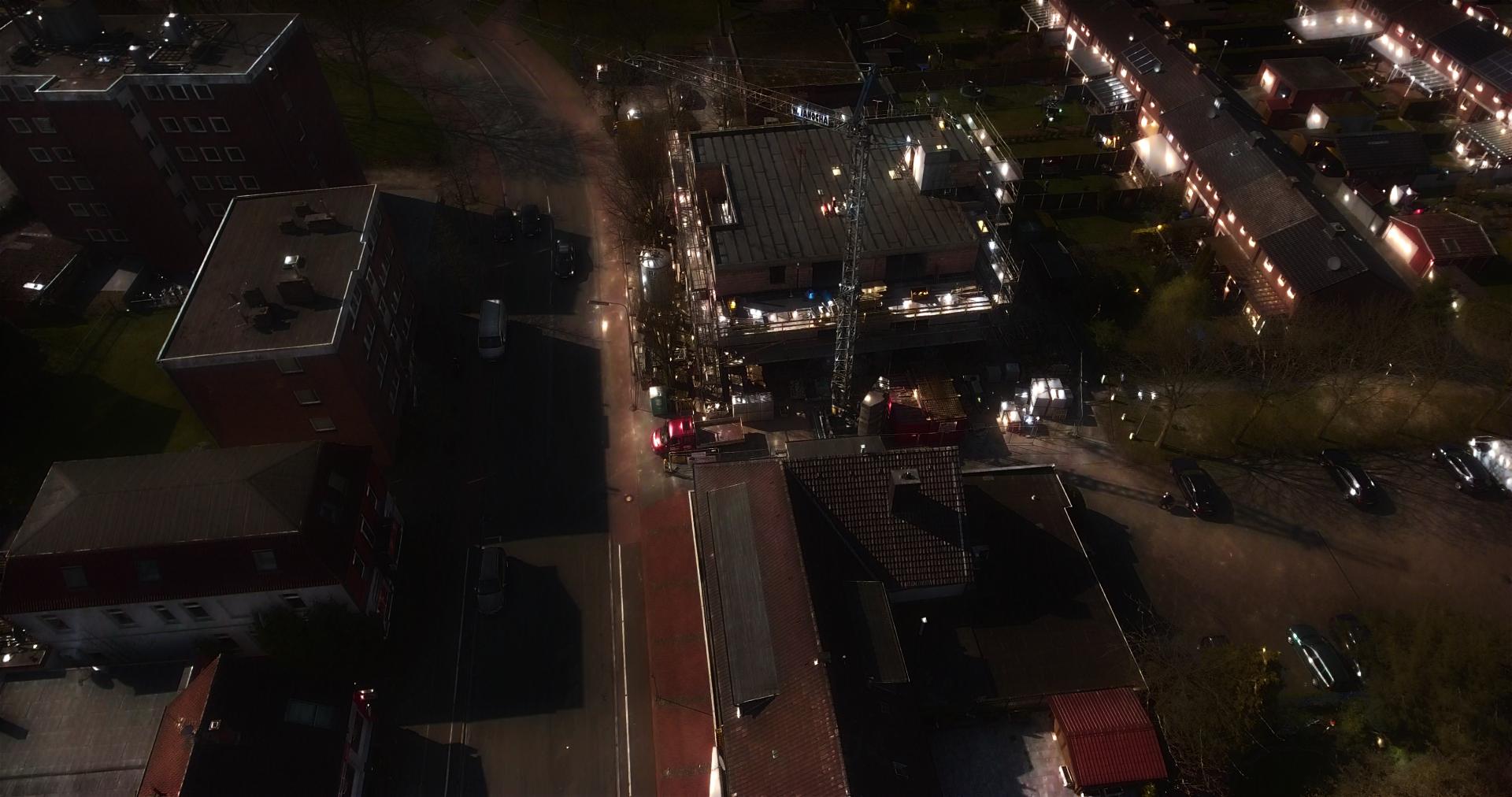}
        \end{subfigure}
        \begin{subfigure}{.24\textwidth}
            \centering
            \caption*{Rain}
            \includegraphics[trim={0 1.3cm 0 1.3cm},clip,width=\textwidth]{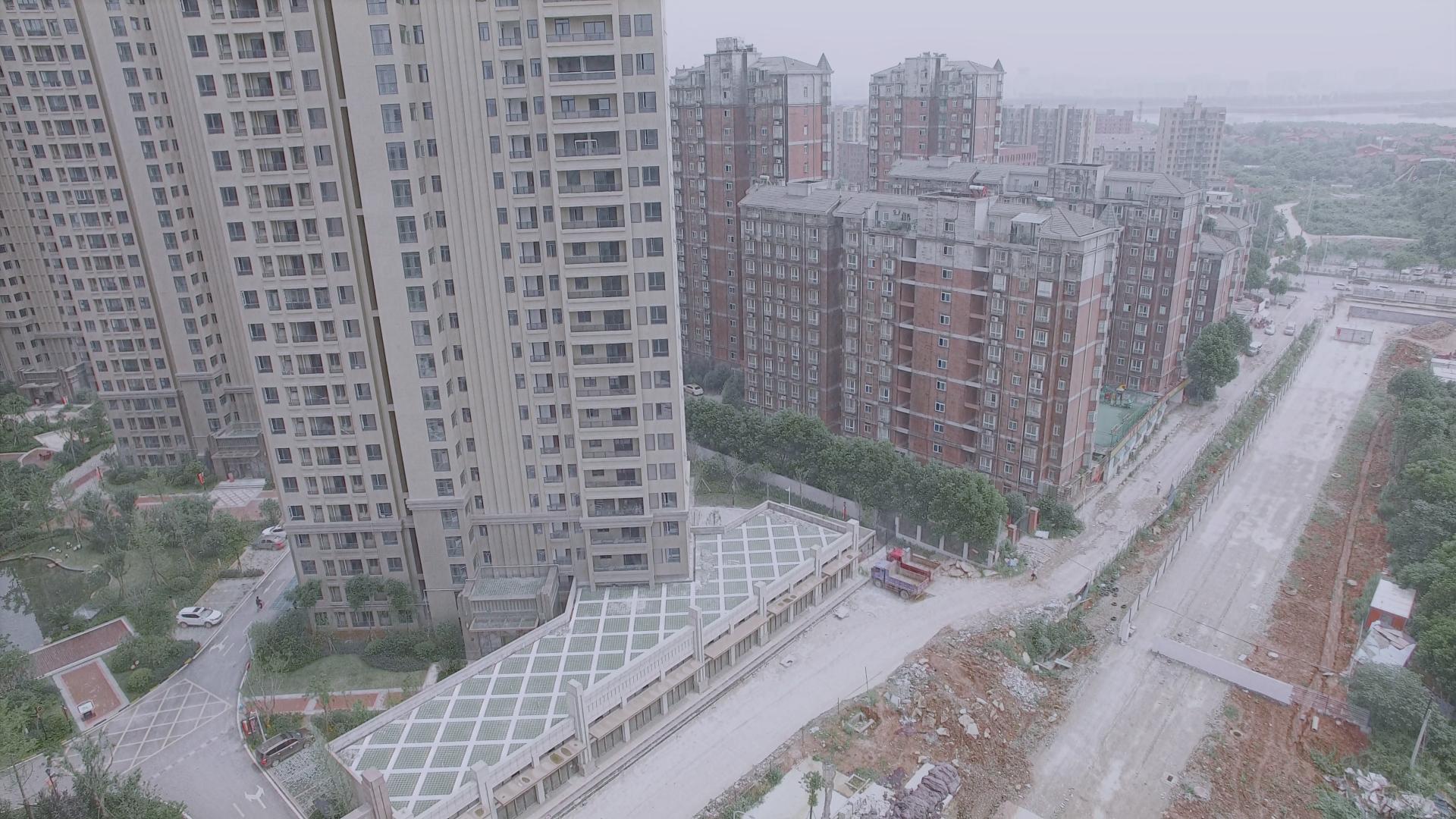}
        \end{subfigure}
        \begin{subfigure}{.24\textwidth}
            \centering
            \caption*{Fog}
            \includegraphics[trim={0 1.3cm 0 1.3cm},clip,width=\textwidth]{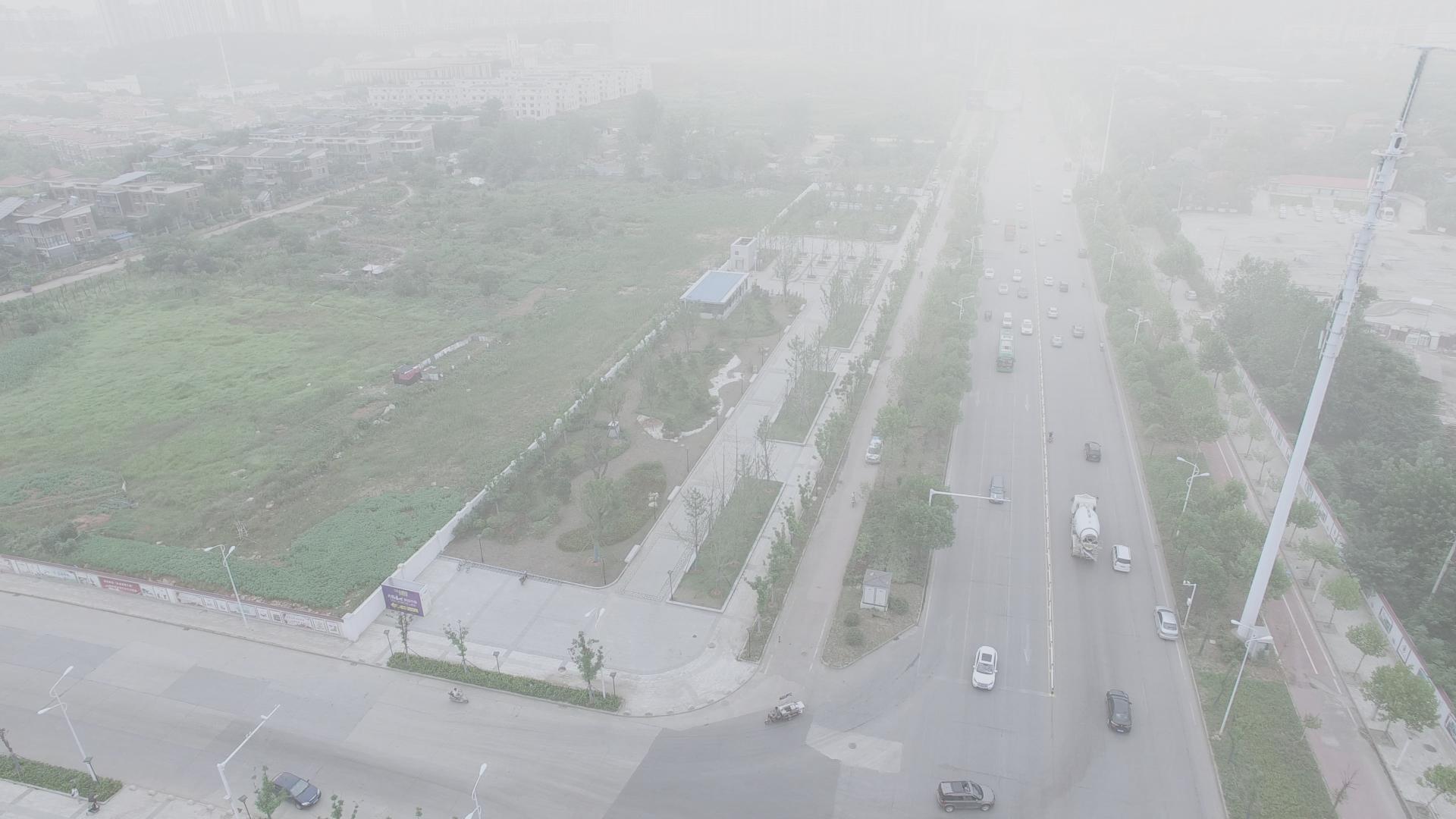}
        \end{subfigure}
    \end{subfigure}
    \begin{subfigure}{\textwidth}
        \centering
        \begin{subfigure}{.24\textwidth}
            \centering
            \includegraphics[trim={0 1.3cm 0 1.3cm},clip,width=\textwidth]{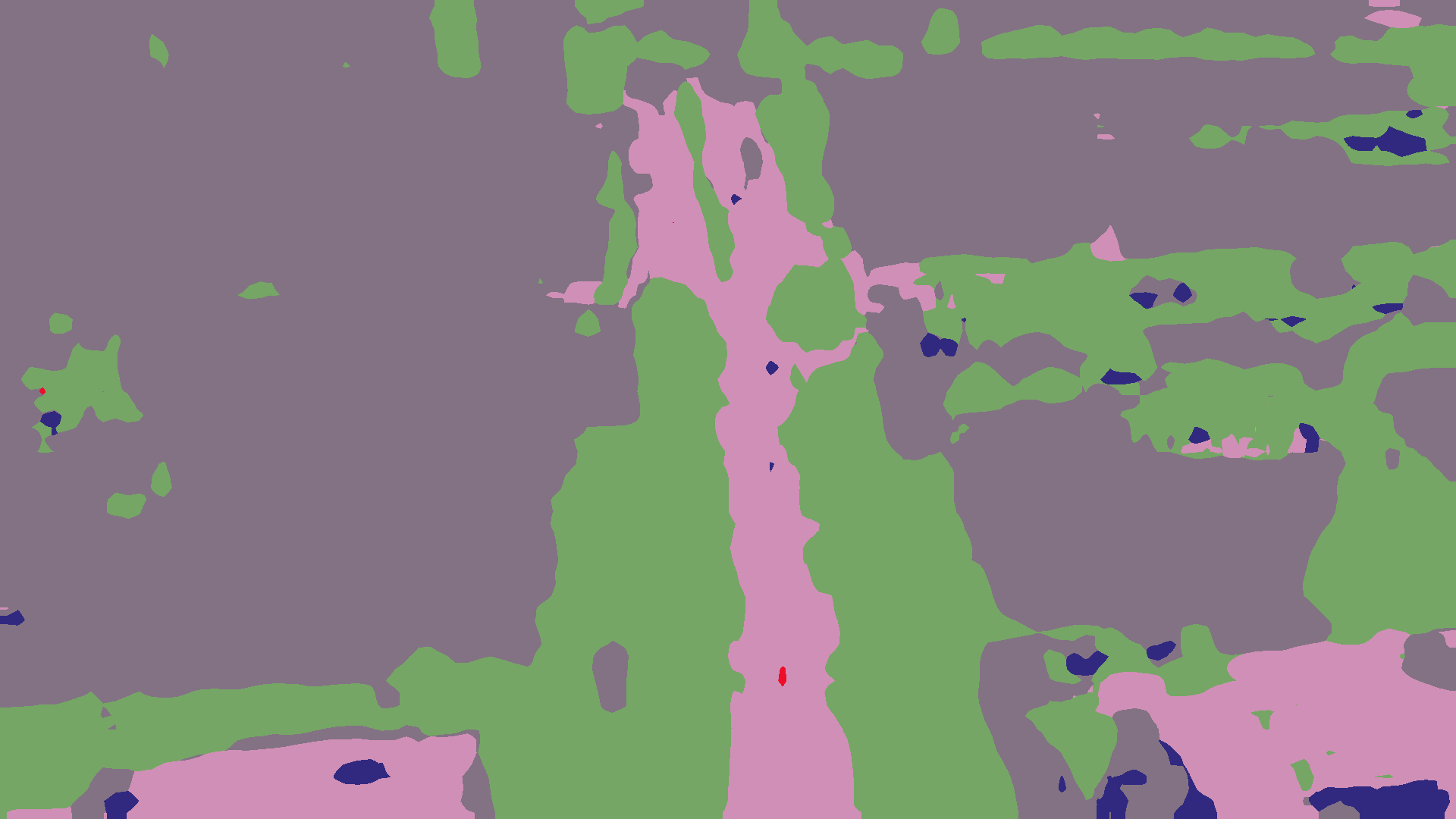}
        \end{subfigure}
        \begin{subfigure}{.24\textwidth}
            \centering
            \includegraphics[width=\textwidth]{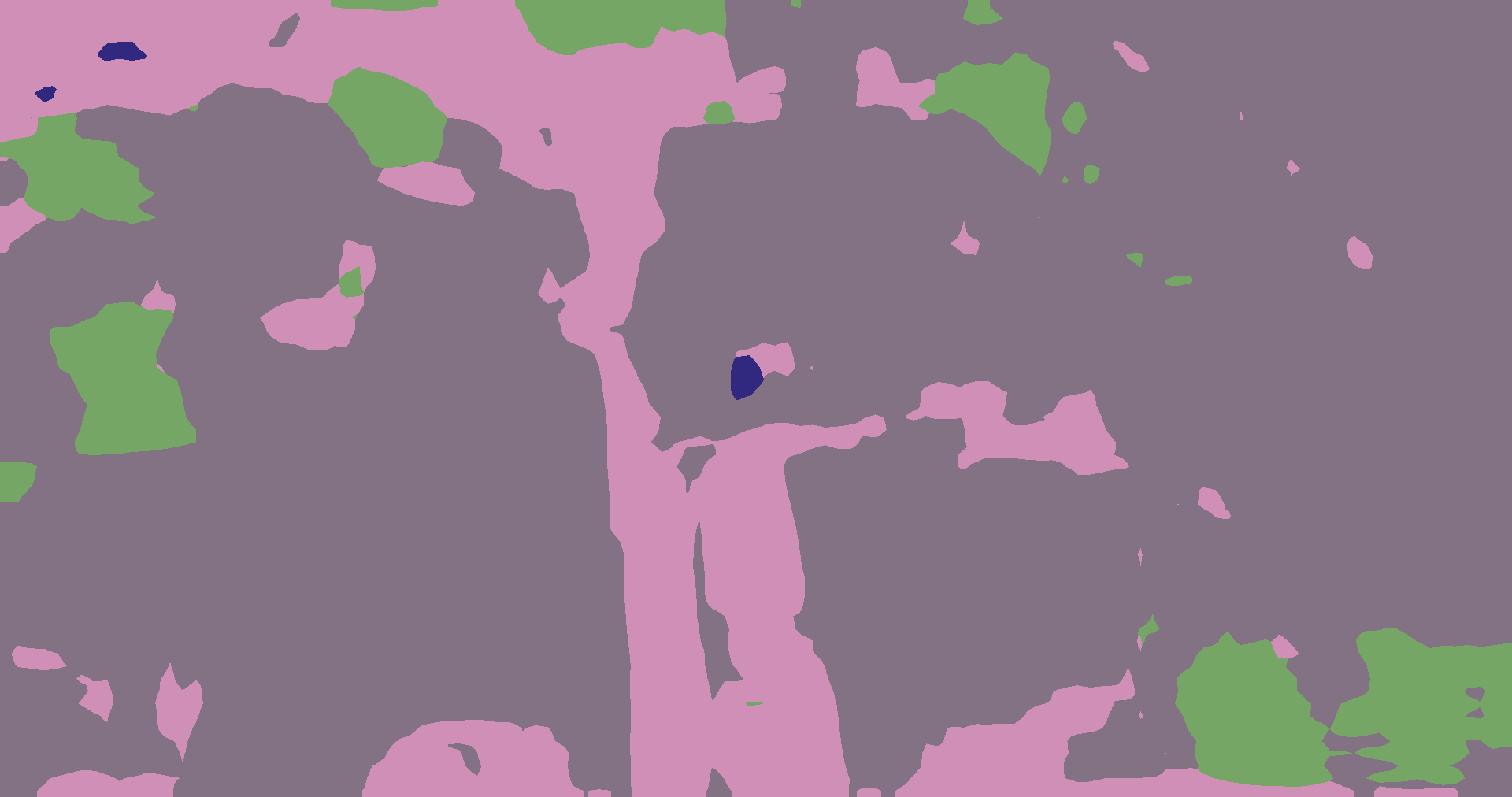}
        \end{subfigure}
        \begin{subfigure}{.24\textwidth}
            \centering
            \includegraphics[trim={0 1.3cm 0 1.3cm},clip,width=\textwidth]{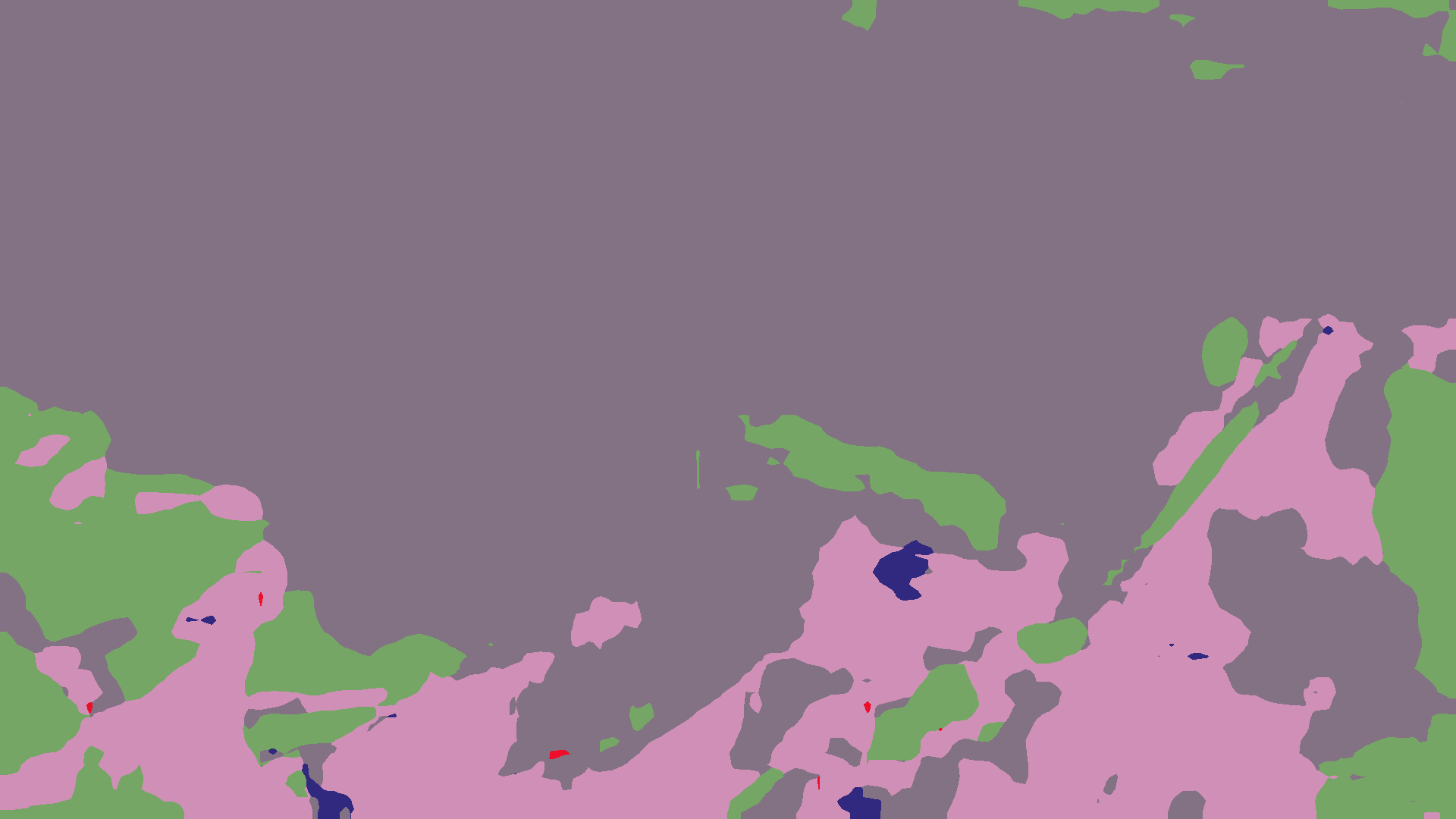}
        \end{subfigure}
        \begin{subfigure}{.24\textwidth}
            \centering
            \includegraphics[trim={0 1.3cm 0 1.3cm},clip,width=\textwidth]{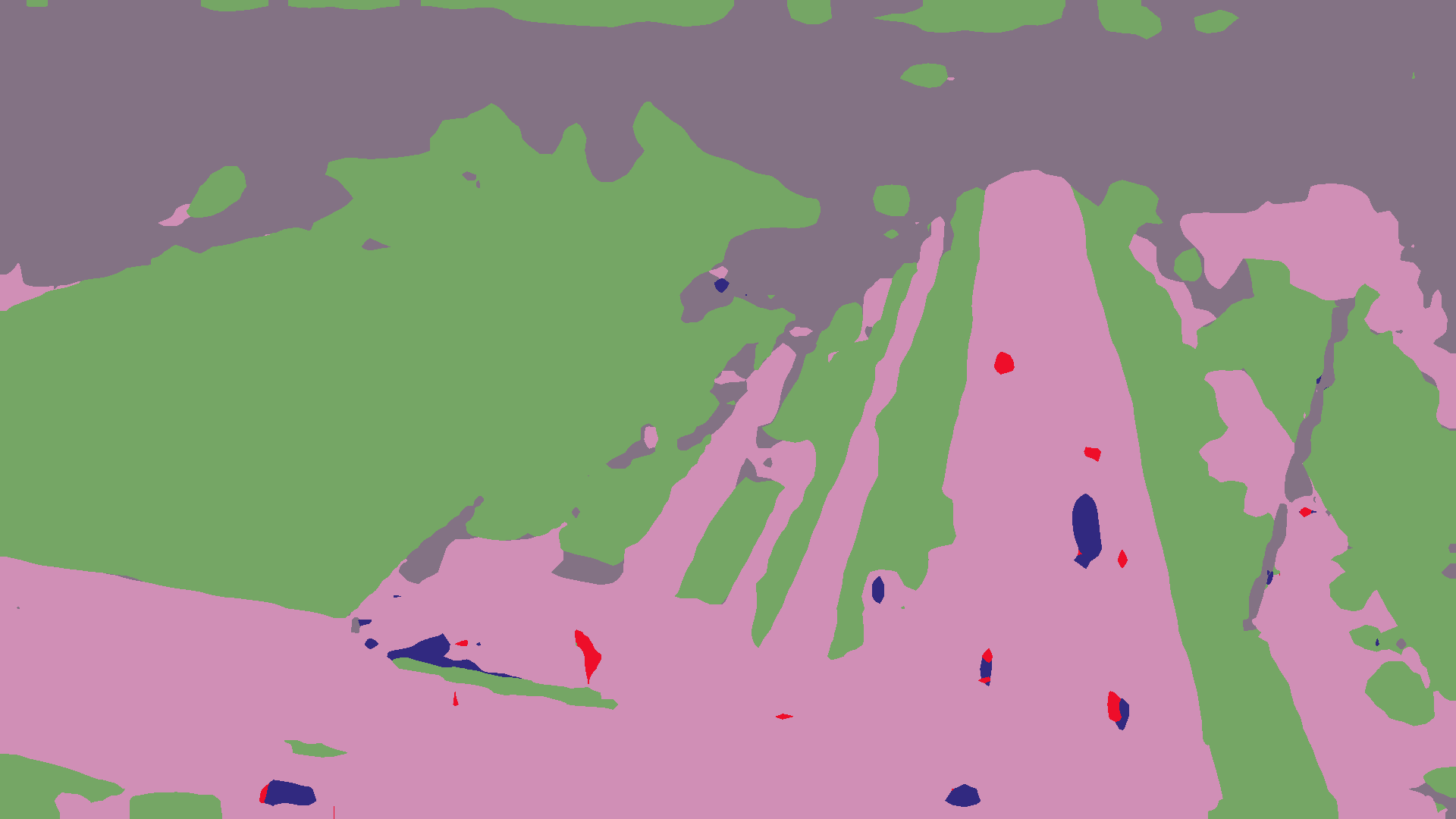}
        \end{subfigure}
    \end{subfigure}
    \begin{subfigure}{\textwidth}
        \centering
        \begin{subfigure}{.24\textwidth}
            \centering
            \includegraphics[trim={0 1.3cm 0 1.3cm},clip,width=\textwidth]{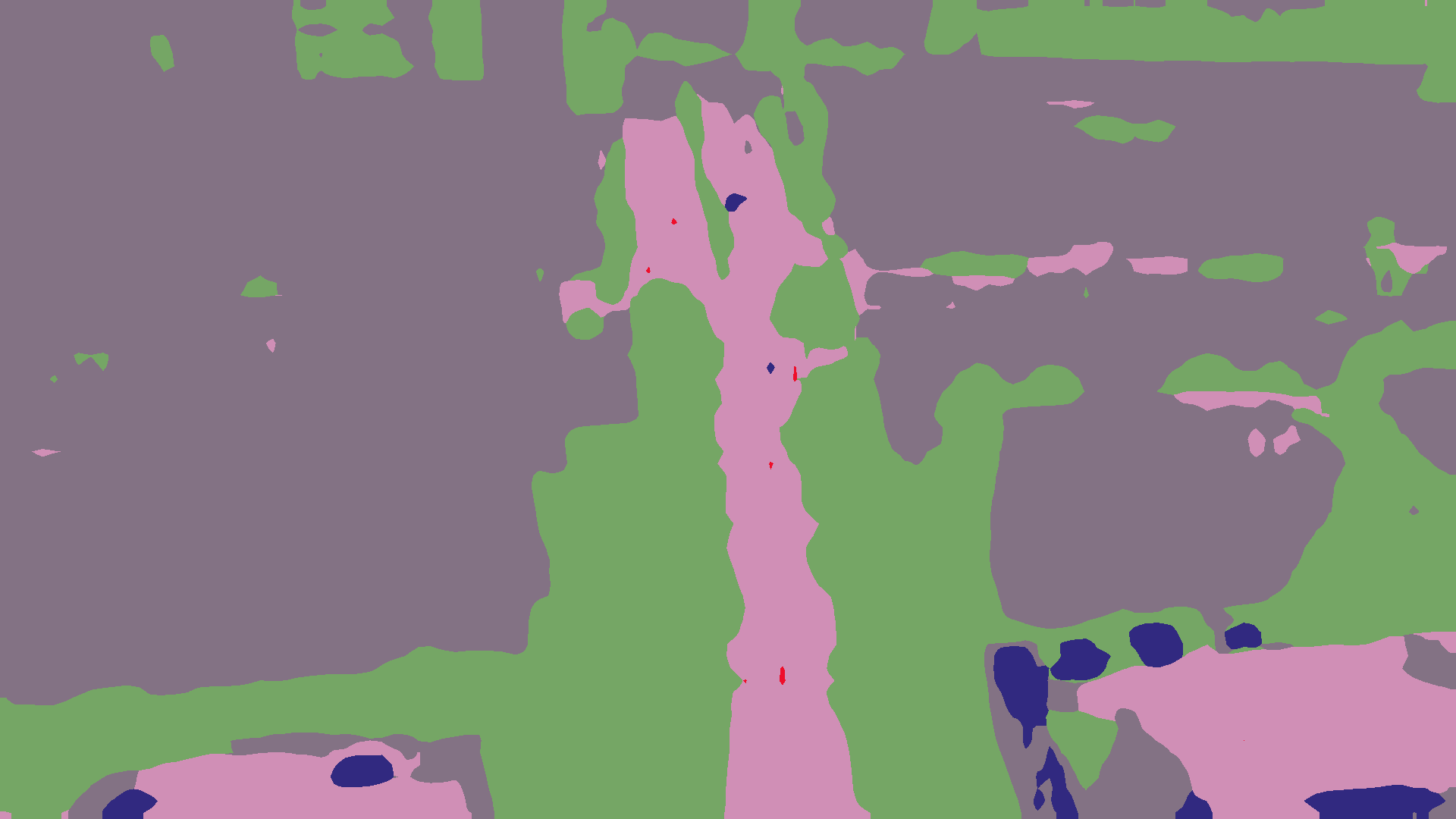}
        \end{subfigure}
        \begin{subfigure}{.24\textwidth}
            \centering
            \includegraphics[width=\textwidth]{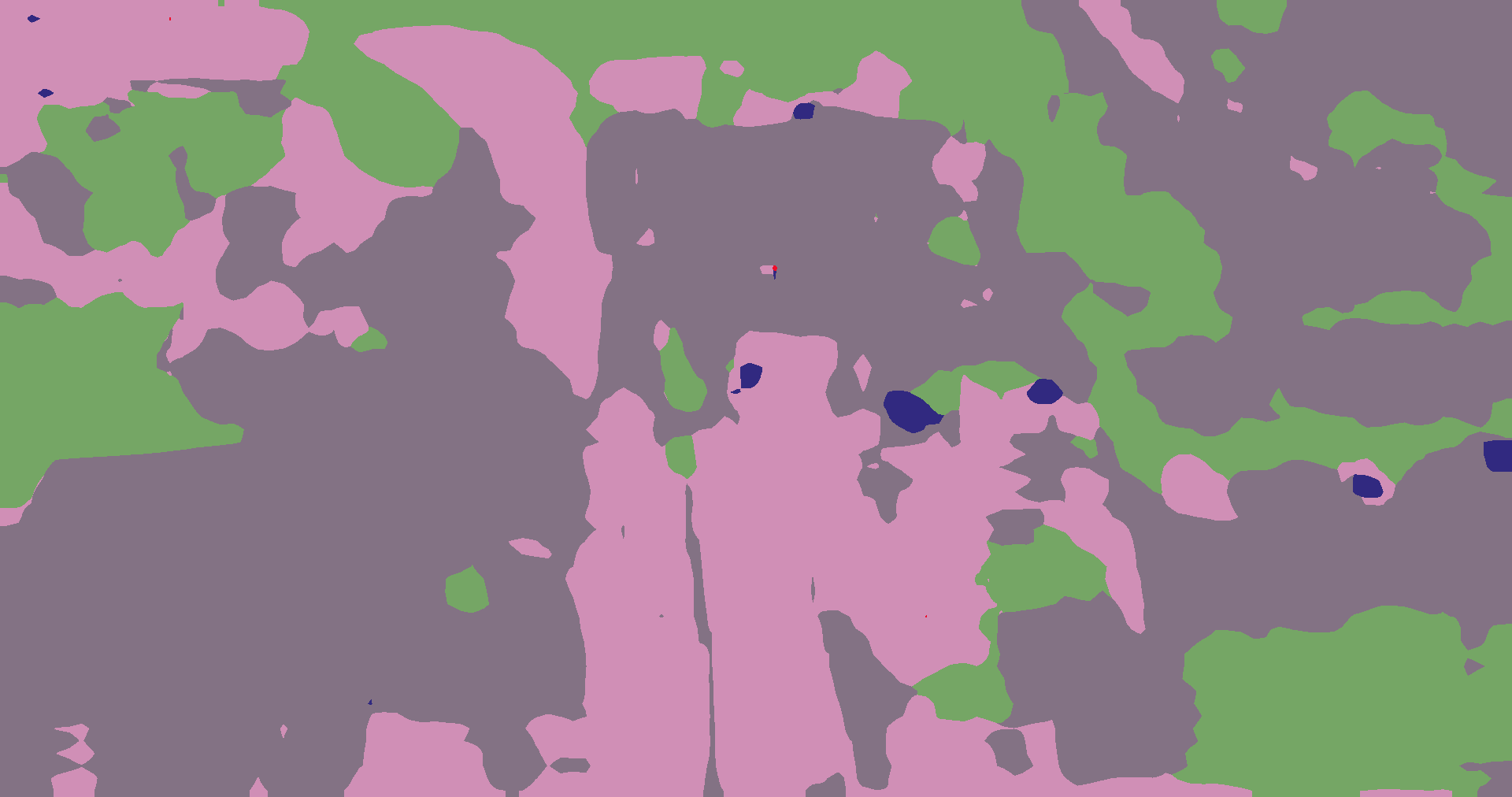}
        \end{subfigure}
        \begin{subfigure}{.24\textwidth}
            \centering
            \includegraphics[trim={0 1.3cm 0 1.3cm},clip,width=\textwidth]{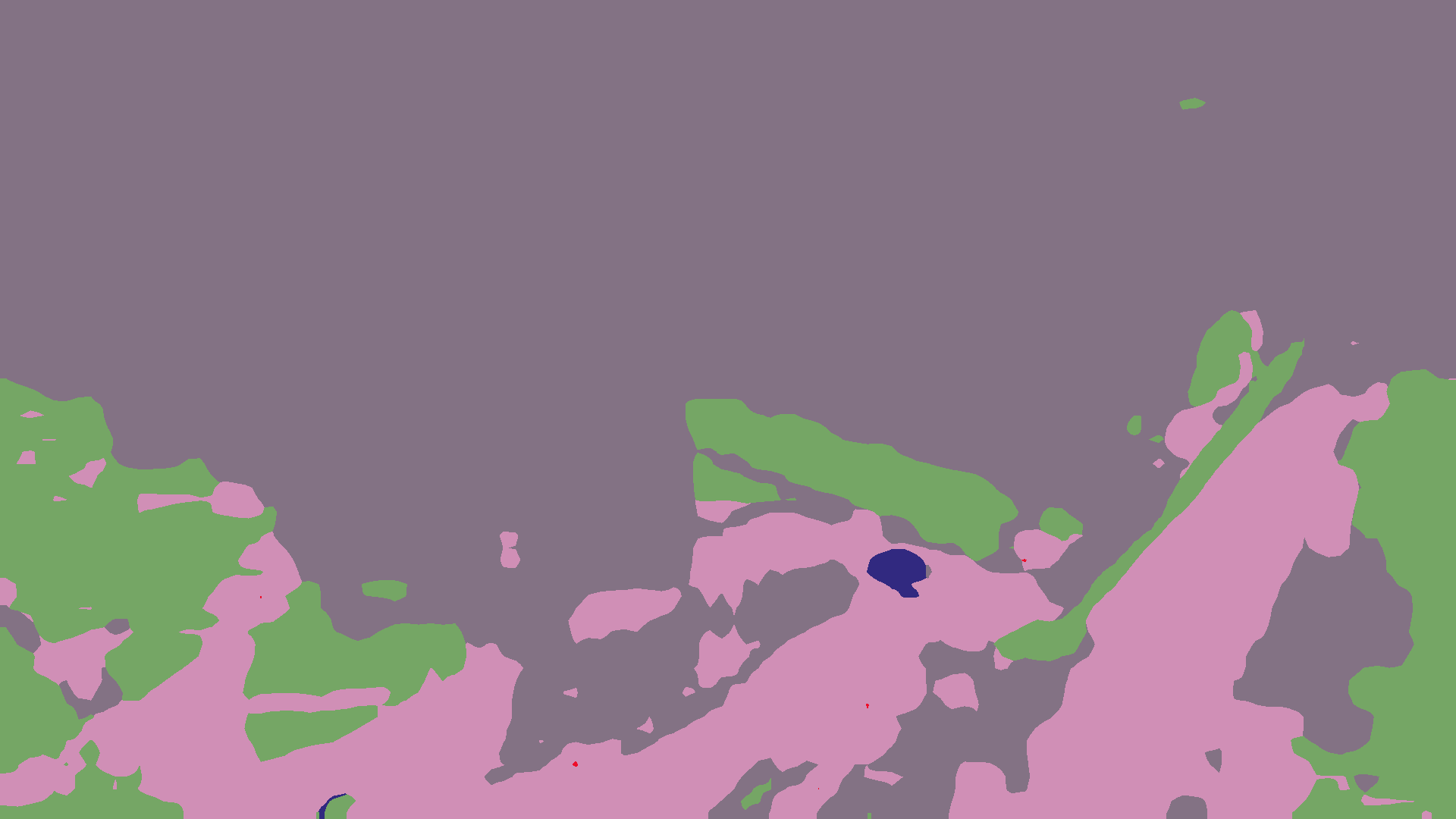}
        \end{subfigure}
        \begin{subfigure}{.24\textwidth}
            \centering
            \includegraphics[trim={0 1.3cm 0 1.3cm},clip,width=\textwidth]{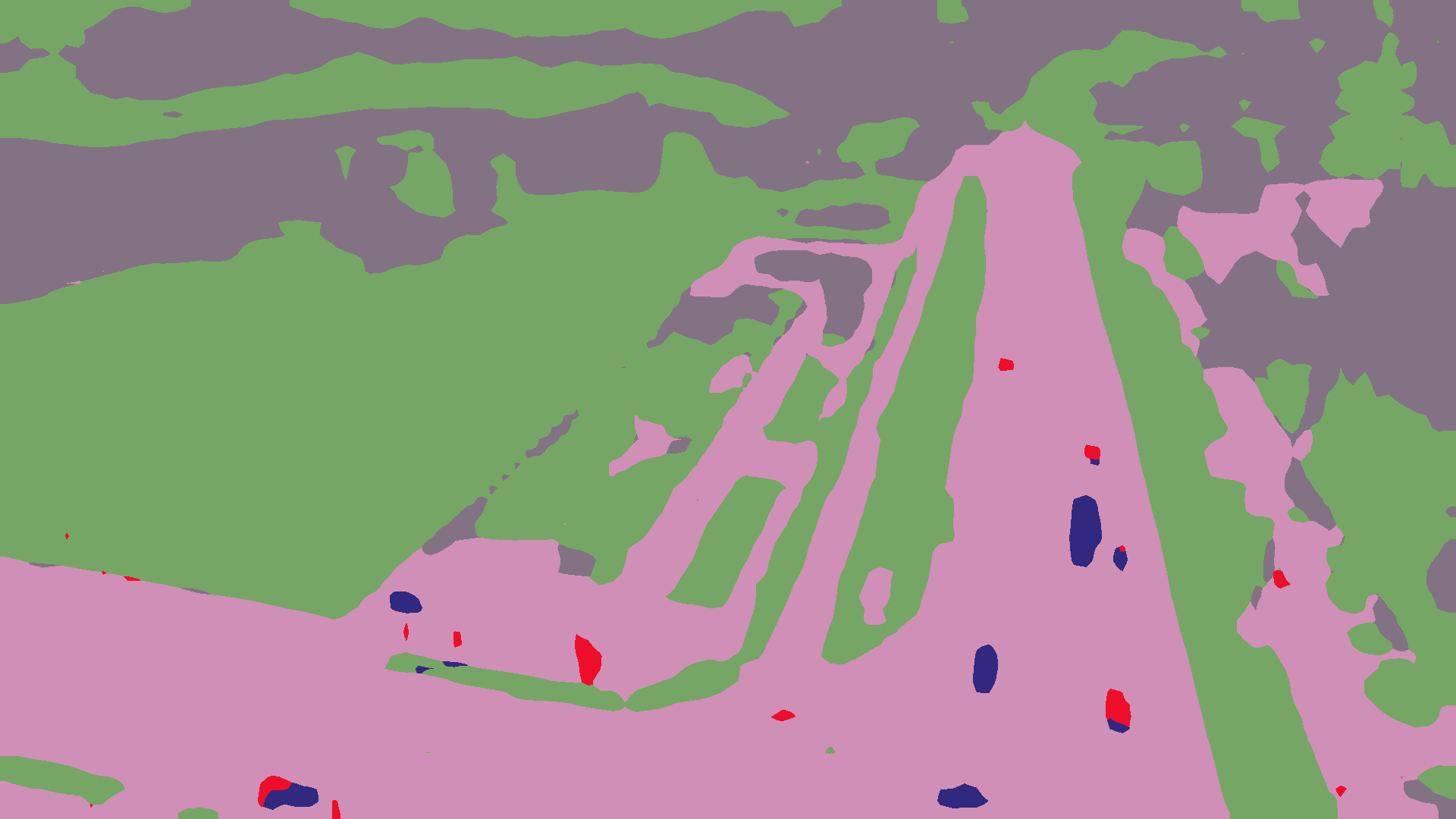}
        \end{subfigure}
    \end{subfigure}
    \begin{subfigure}{\textwidth}
        \centering
        \begin{subfigure}{.24\textwidth}
            \centering
            \includegraphics[trim={0 1.3cm 0 1.3cm},clip,width=\textwidth]{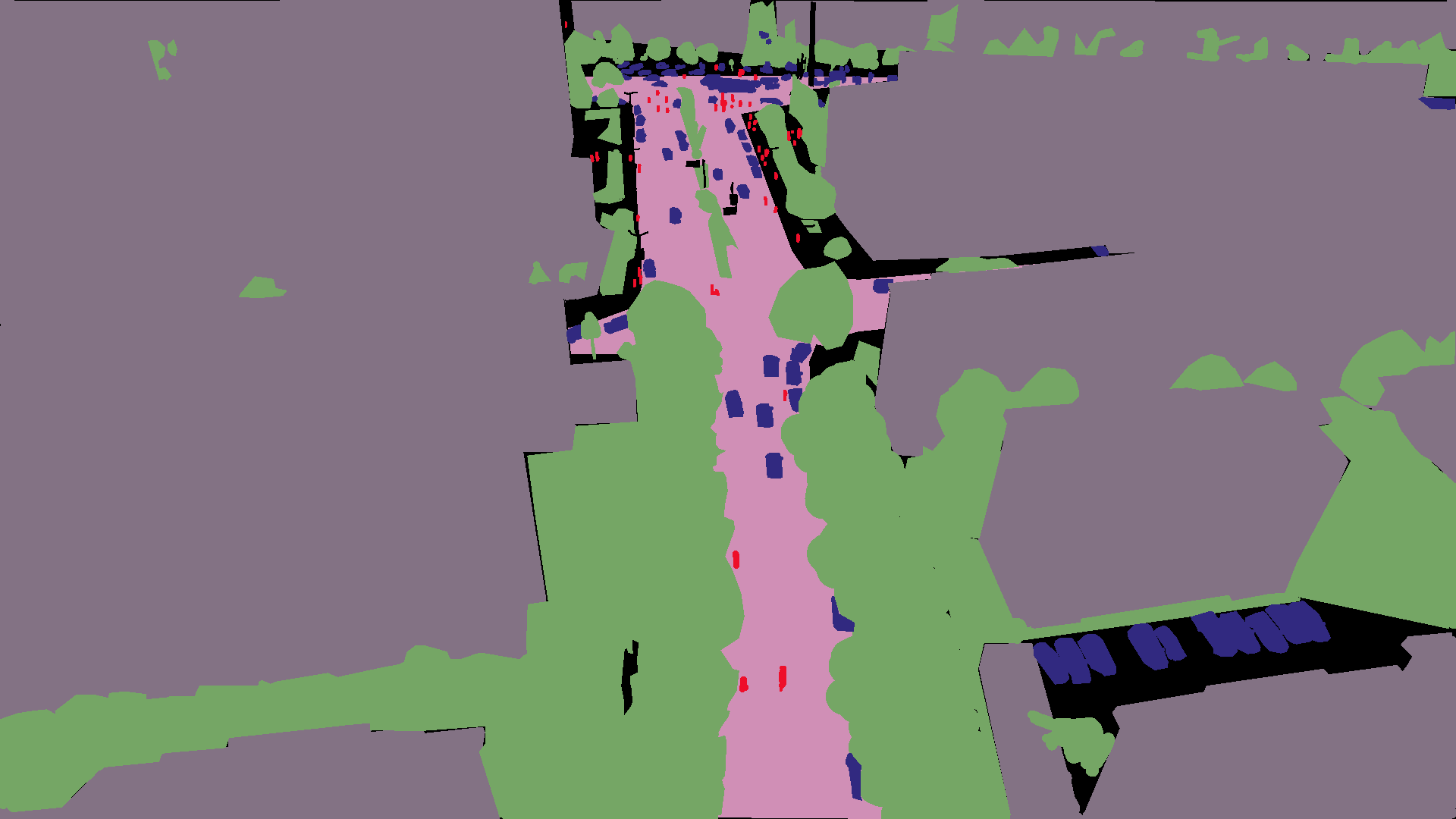}
        \end{subfigure}
        \begin{subfigure}{.24\textwidth}
            \centering
            \includegraphics[width=\textwidth]{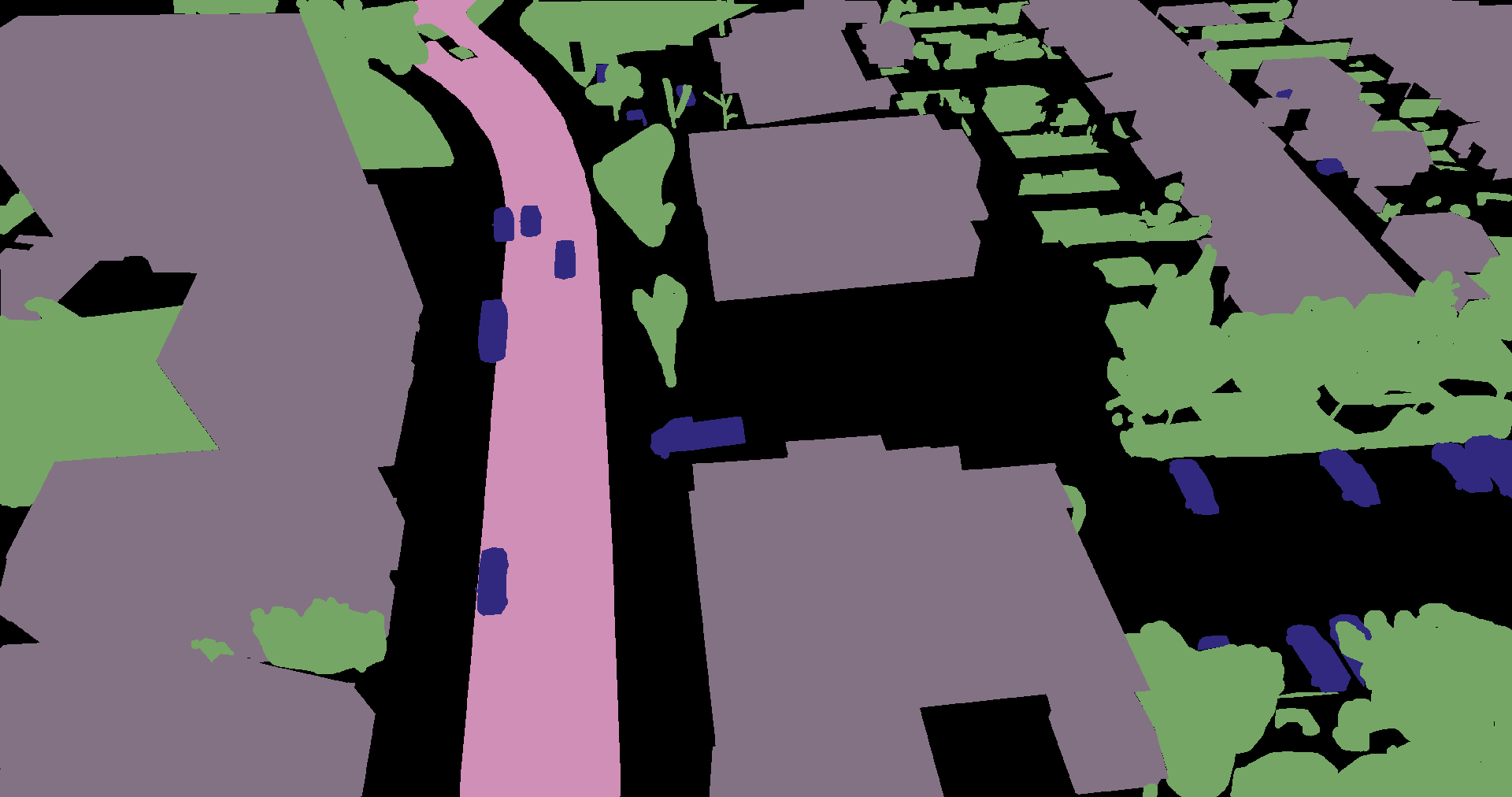}
        \end{subfigure}
        \begin{subfigure}{.24\textwidth}
            \centering
            \includegraphics[trim={0 1.3cm 0 1.3cm},clip,width=\textwidth]{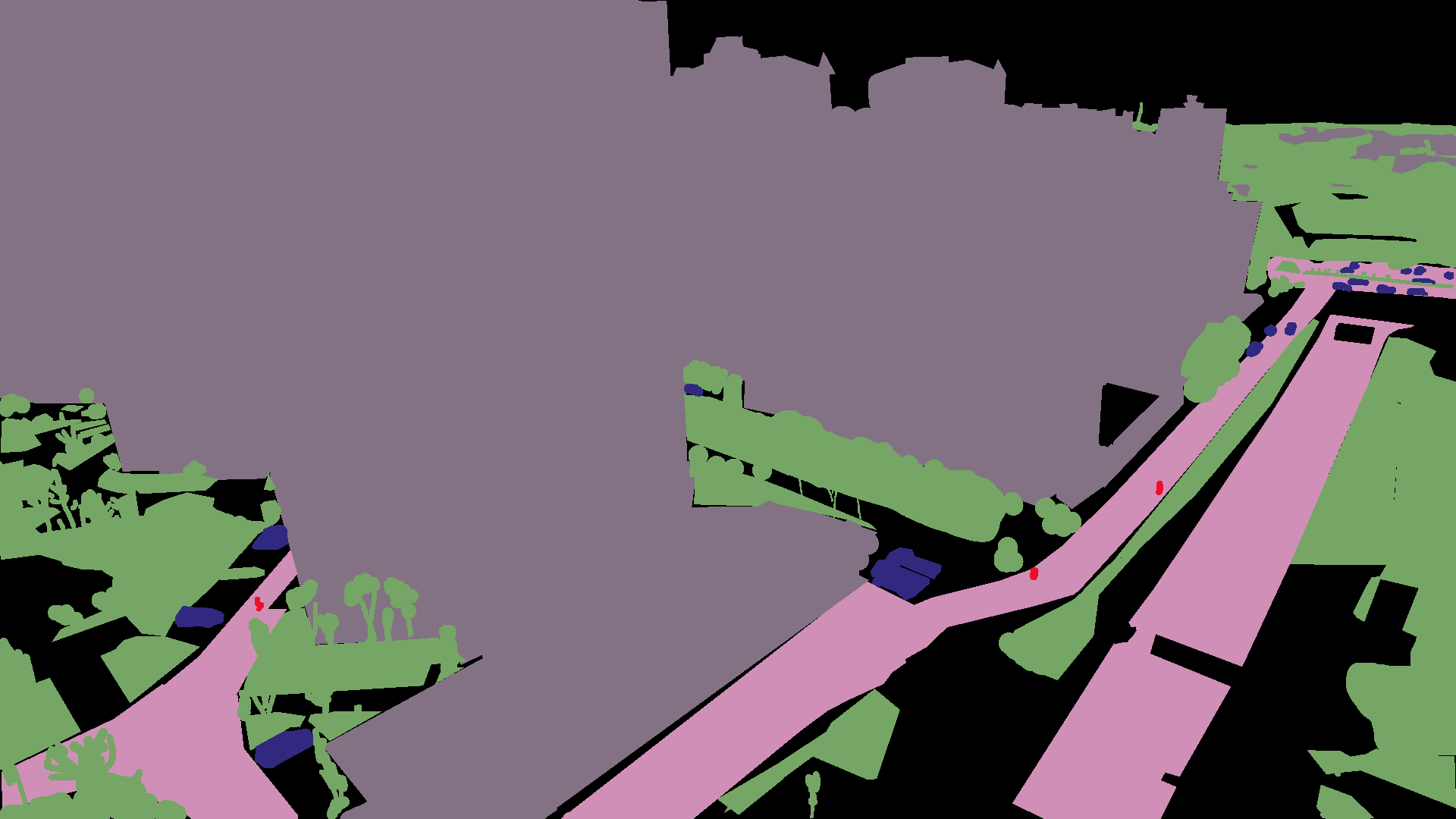}
        \end{subfigure}
        \begin{subfigure}{.24\textwidth}
            \centering
            \includegraphics[trim={0 1.3cm 0 1.3cm},clip,width=\textwidth]{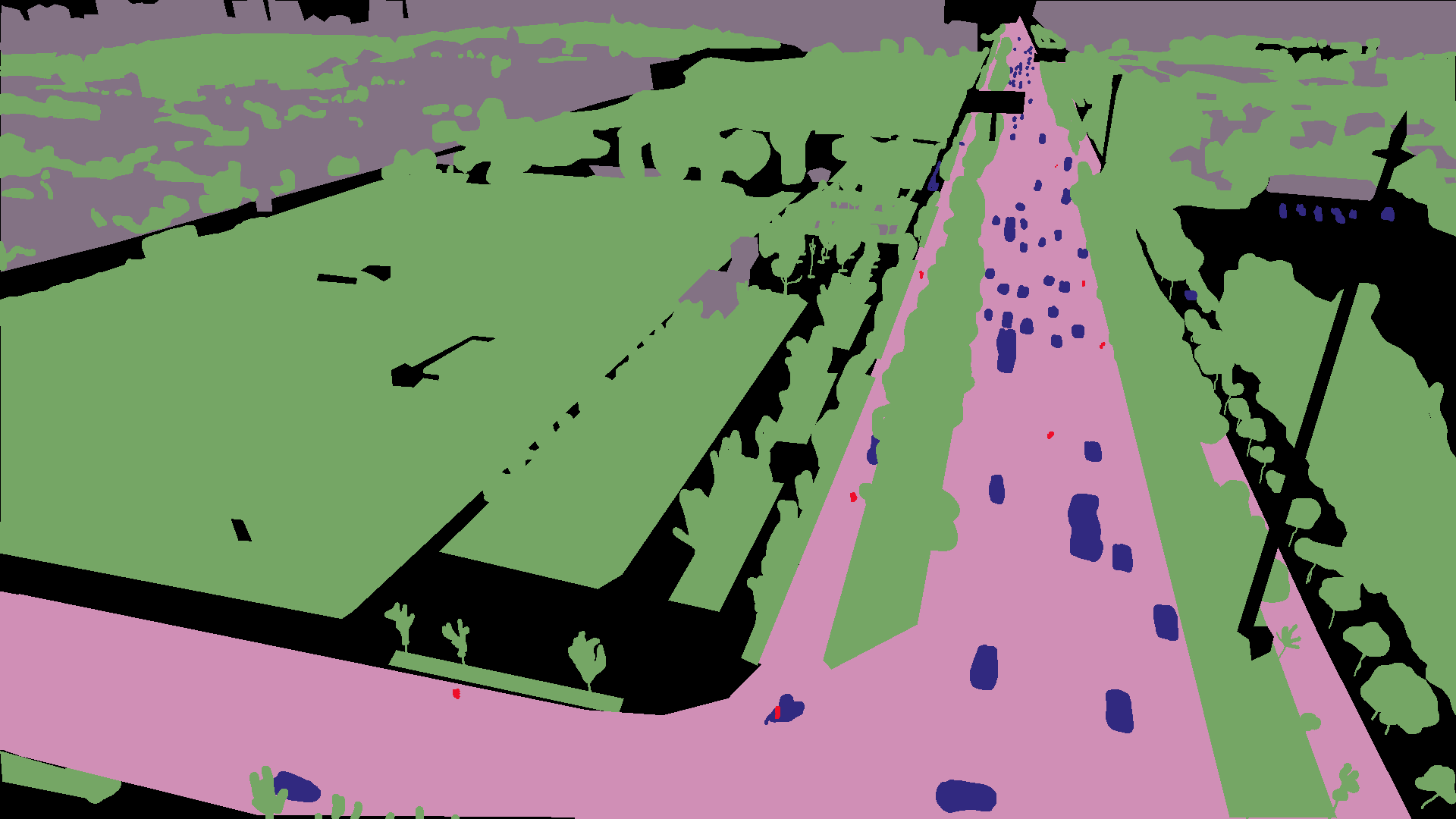}
        \end{subfigure}
    \end{subfigure}
    \caption{Qualitative experiments: model trained on synthetic samples and tested on real samples with varying weather conditions. First row: Input, Second row: Model Prediction (no adaptation), Third Row: Model  Prediction (UDA), Fourth Row: Ground Truth. }
    \label{fig:quali_real_weathers}
\end{figure*}

The next step is to deploy the models trained on the synthetic FlyAwareV2 data in the real world.
We start by showing the results achieved by training on synthetic data and testing on real-world data across varying weather conditions in Table~\ref{tab:rgb_real_weather} and Figure~\ref{fig:quali_real_weathers}.
Note that, for the synthetic pretraining, we trained the DeepLabV3/MobileNetV3-large segmenter using only color data and the fine-level class-set. At test time, we map the fine-level class-set into the corresponding coarse-level one (refer to Table~\ref{tab:cmap} for more details).
As expected, when considering all the data, it can be observed that there is a significant drop in accuracy compared to synthetic test data (on the coarse set, the accuracy was $64.5\%$), but a reasonably good accuracy of $42.3\%$ can be achieved.

Working in clear daytime conditions is, again, easier with an accuracy that reaches almost $50\%$, however, models trained on clear weather data struggle to generalize when tested on real-world nighttime and foggy weather scenarios: using only clear weather data for training, the overall mIoU drops significantly from $48.7\%$ on daytime test data to $21.3\%$ for nighttime and $12.9\%$ for fog, while rain is slightly better at $37.4\%$. Training in a mixture of all weather conditions leads to much better performances, especially in rainy and foggy data ($48.4\%$ and $38.4\%$), while the night setting remains the most challenging at $23.4\%$. 
This highlights the importance of incorporating diverse environmental variations during training to enhance the robustness of UAV perception systems.

\begin{wraptable}{r}{.45\textwidth}
    \setlength{\tabcolsep}{.3em}
    \setlength{\belowcaptionskip}{-1em}
    \centering
    \begin{tabular}{c|c:c}
        Height & Real mIoU & Synth mIoU \\
        \hline
        20m & \textbf{44.5} & 50.3 \\
        50m & 37.2 & \underline{56.7} \\
        80m & 29.4 & 50.5 \\
        \hdashline
        All & \underline{42.3} & \textbf{64.5} \\
    \end{tabular}
    \caption{Synthetic-to-Real adaptation varying training height, RGB only, Coarse class-set.}
    \label{tab:rgb_real_height}
\end{wraptable}

Like before, the qualitative results reported in Figure \ref{fig:quali_real_weathers} support the quantitative experiments. For this discussion, we focus on the predictions of the second row; the ones in the third will be discussed in the Unsupervised Domain Adaptation section. 
The daytime prediction offers the highest accuracy, especially on small segments like \textit{person}. 
Rain and fog, again, offer suboptimal but acceptable results, given the relative similarity to the daytime conditions. 
On the other hand, the nighttime environment results in highly degraded performance, with significant confusion between classes (\eg, \textit{Building} vs. \textit{Vegetation}).

\begin{figure*}[b]
    \centering
    \begin{subfigure}{\textwidth}
        \centering
        \begin{subfigure}{.19\textwidth}
            \centering
            \caption*{Input}
            \includegraphics[width=\textwidth]{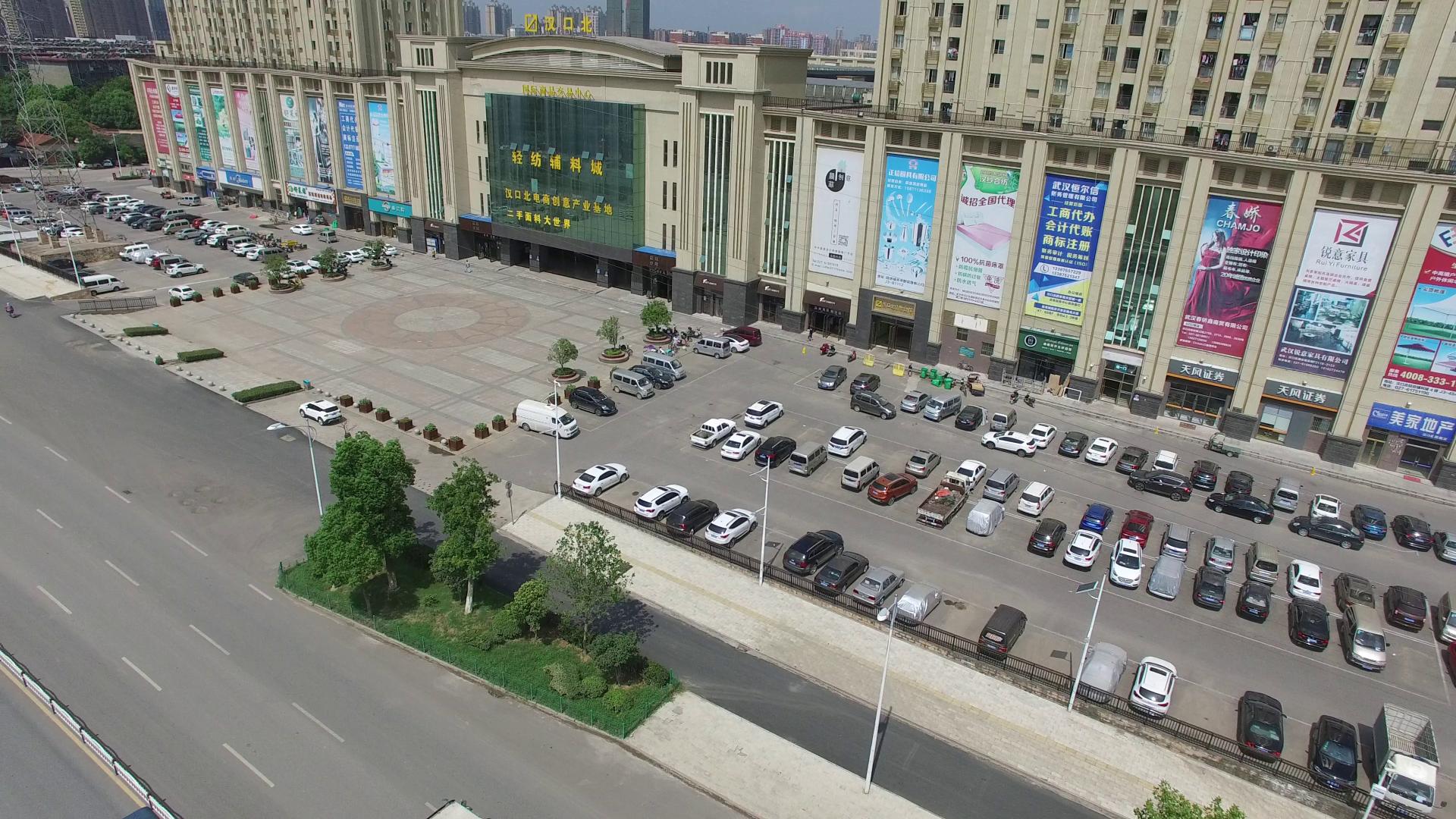}
        \end{subfigure}
        \begin{subfigure}{.19\textwidth}
            \centering
            \caption*{GT}
            \includegraphics[width=\textwidth]{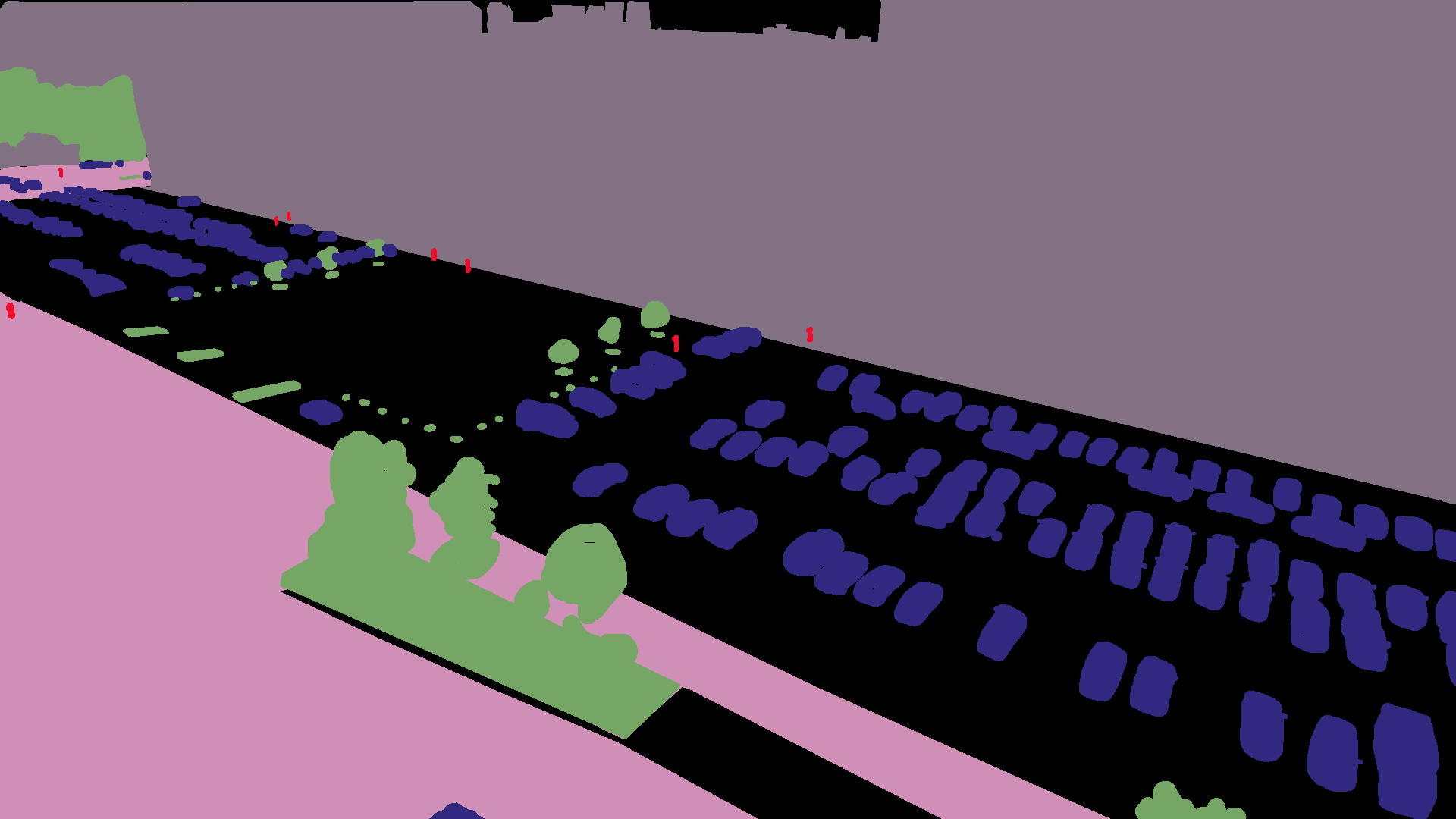}
        \end{subfigure}
        \begin{subfigure}{.19\textwidth}
            \centering
            \caption*{20m}
            \includegraphics[width=\textwidth]{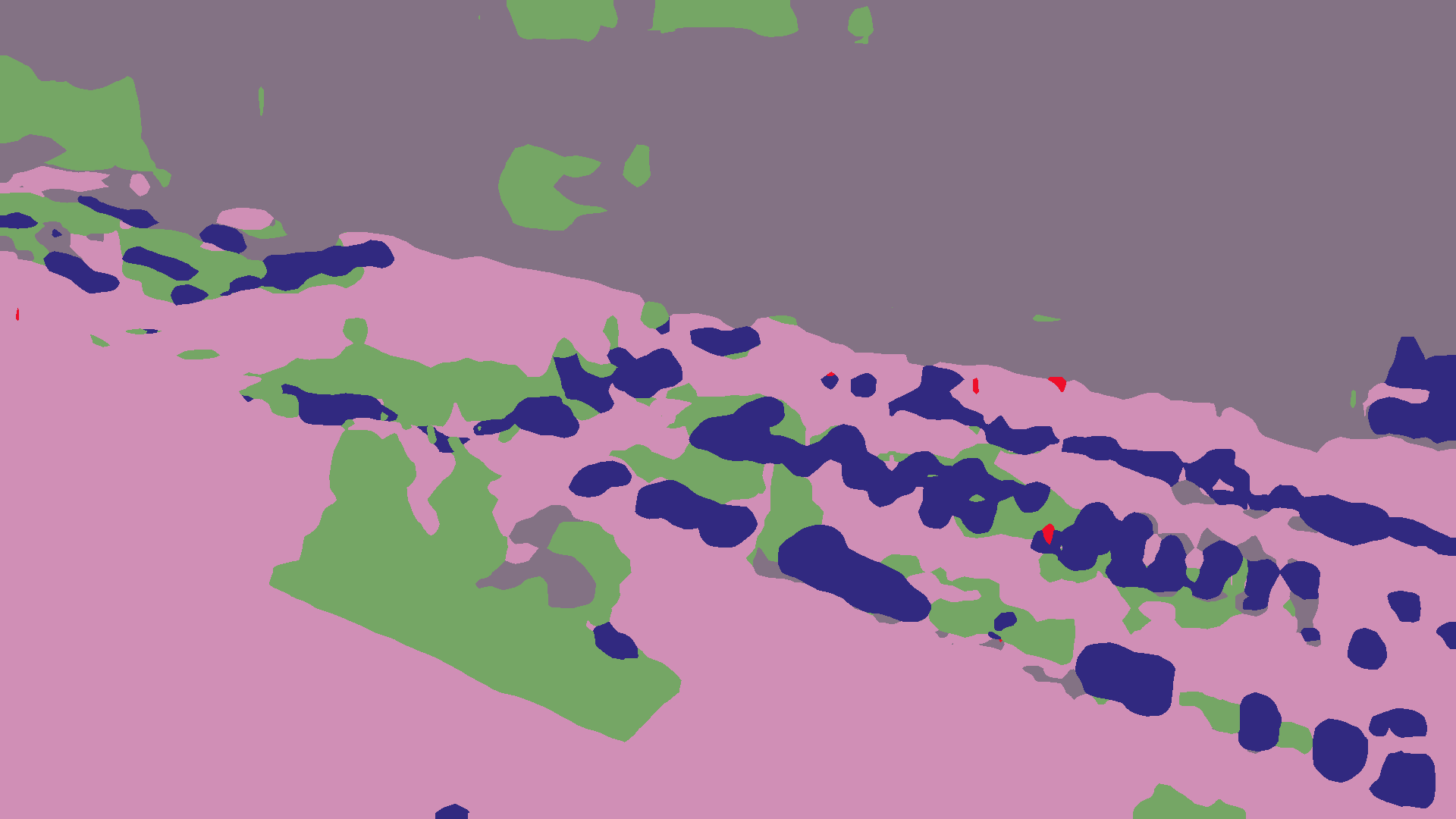}
        \end{subfigure}
        \begin{subfigure}{.19\textwidth}
            \centering
            \caption*{50m}
            \includegraphics[width=\textwidth]{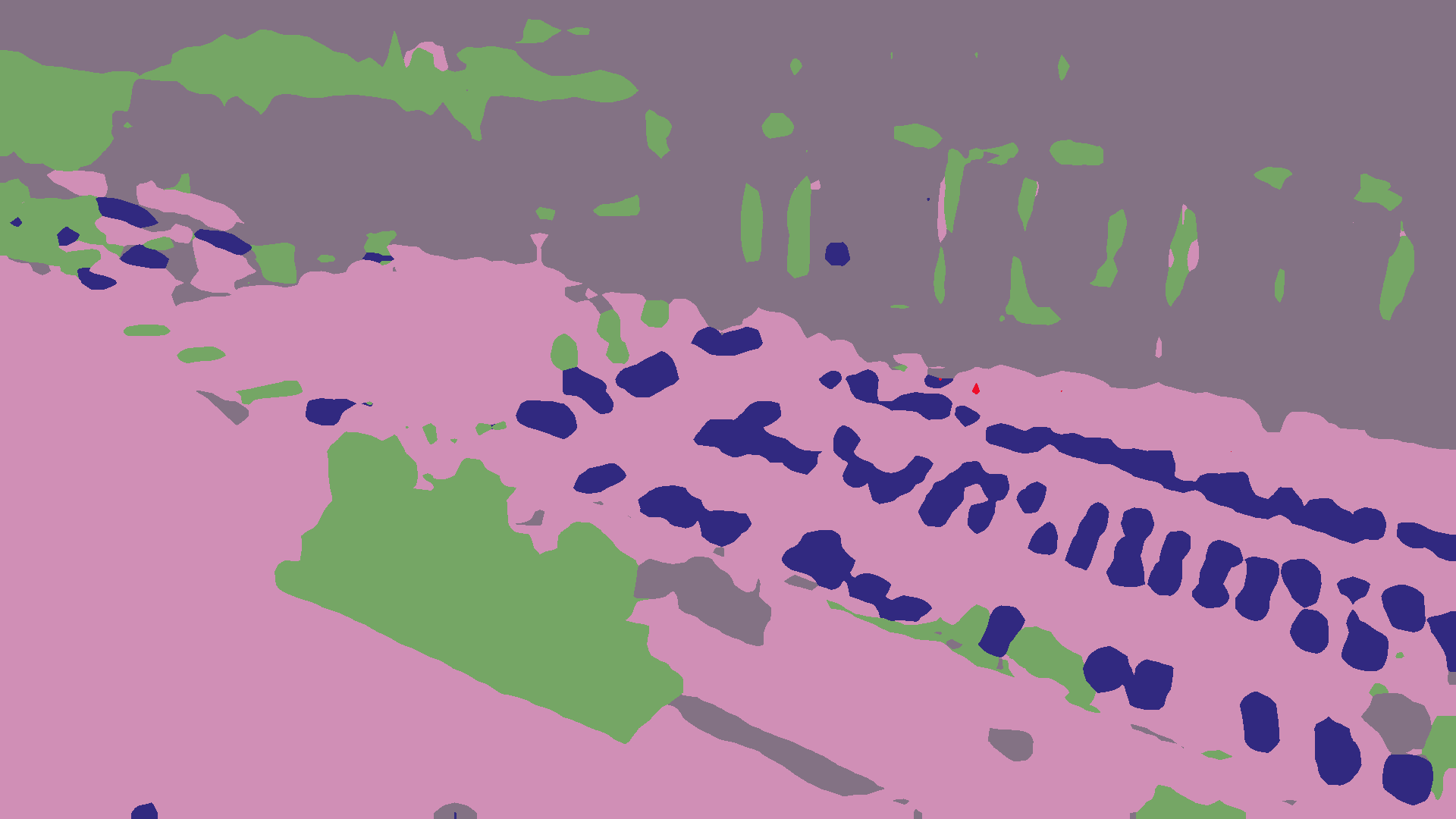}
        \end{subfigure}
        \begin{subfigure}{.19\textwidth}
            \centering
            \caption*{80m}
            \includegraphics[width=\textwidth]{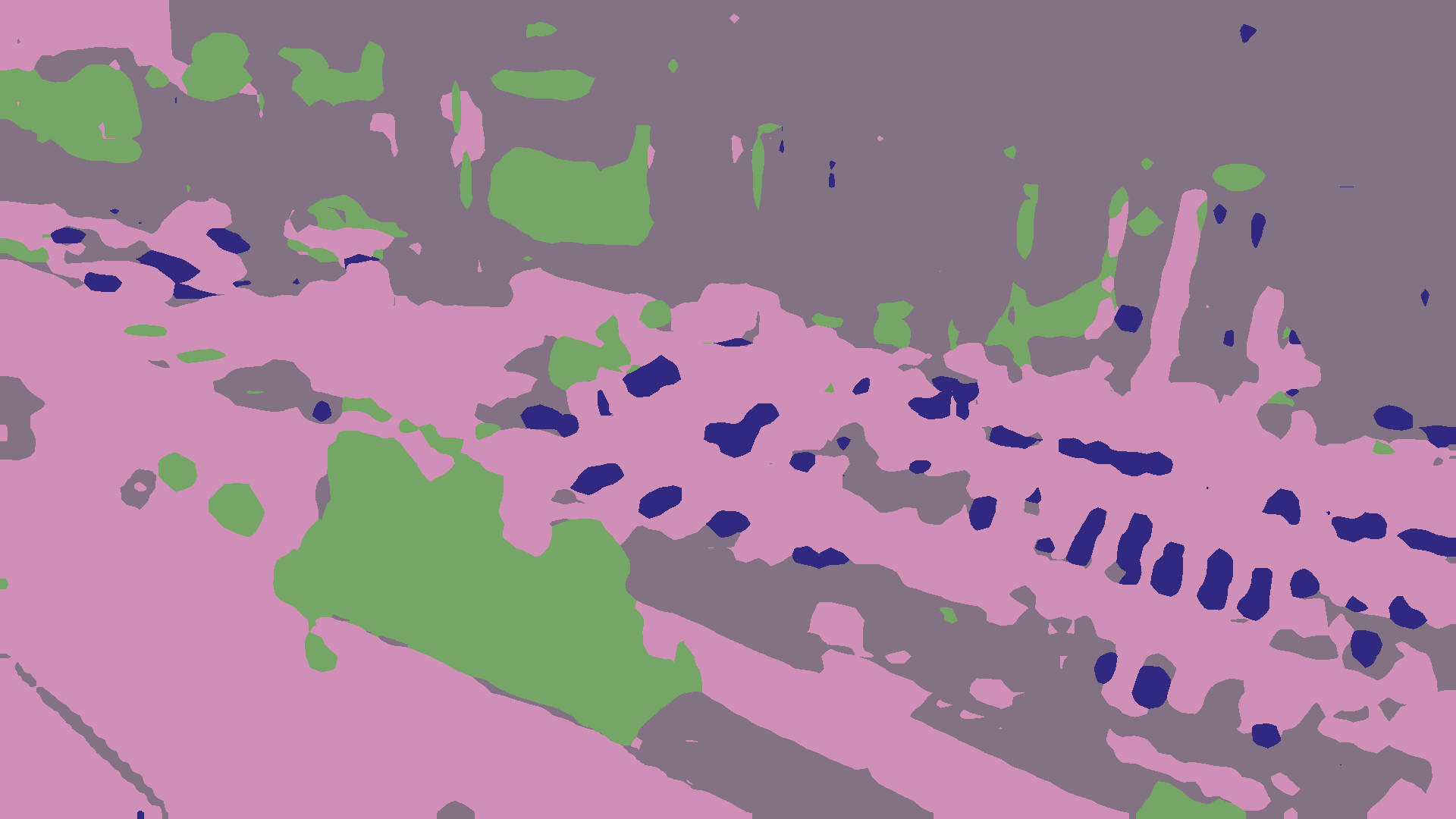}
        \end{subfigure}
    \end{subfigure}
    \caption{Qualitative experiments, Model trained at different heights and tested on real samples.}
    \label{fig:quali_heights}
\end{figure*}

The impact of varying UAV flight height during training on synthetic-to-real adaptation performance is investigated instead in Table~\ref{tab:rgb_real_height} and Figure~\ref{fig:quali_heights} (note that, in this case, the test samples have been taken at different heights and height data is not provided in the source real-world datasets). 
As expected, models trained at lower altitudes ($20$m) achieve better generalization, with a mIoU of $44.5\%$, compared to $37.2\%$ at heights of $50$m and $29.4\%$ at heights of $80$m. 
This can be attributed to the increased level of detail and resolution available in lower-altitude images, which aids in learning more discriminative features for semantic segmentation.
In the Figure, we show how the architectures trained at different heights predict the input sample. We can observe a strong correlation between altitude and performance degradation, highlighting the differences in PoV between real and synthetic samples. 
This, of course, depends on the particular height of our real samples; different environments or applications may be closer to other synthetic configurations, highlighting the importance of data heterogeneity.

\begin{wraptable}{r}{.45\textwidth}
    \setlength{\tabcolsep}{.3em}
    \setlength{\belowcaptionskip}{-1em}
    \centering
    \begin{tabular}{c|c:c}
        Town & Real mIoU & Synth mIoU \\
        \hline
        Town01 & 34.9 & 41.9 \\
        Town02 & \underline{36.1} & 41.7 \\
        Town03 & 35.1 & \underline{44.9} \\
        Town04 & 32.4 & 43.2 \\
        Town05 & 34.3 & 42.2 \\
        Town06 & 26.0 & 38.8 \\
        Town07 & 27.9 & 33.9 \\
        Town10HD & 34.3 & 36.1 \\
        \hdashline
        All & \textbf{42.3} & \textbf{64.5}
    \end{tabular}
    \caption{Synthetic-to-Real adaptation varying training town, RGB only, Coarse class-set.}
    \label{tab:rgb_real_town}
\end{wraptable}  

The influence of the specific synthetic urban environment used for training is analyzed in Table \ref{tab:rgb_real_town}. Although there are variations in performance when employing different virtual towns, the general trend suggests that models trained on diverse synthetic environments can generalize reasonably well to real-world data, with mIoU scores ranging from $26.0\%$ to $36.1\%$. However, training with all towns together leads to a score of $42.3\%$, much better than using each town alone. 
This underscores the importance of leveraging diverse synthetic data environments to improve generalization capabilities.

\subsection{Synthetic-to-real Unsupervised Domain Adaptation results} \label{sub:results:exp_uda}

As discussed in Section \ref{sub:results:exp_real}, the domain shift between the synthetic and real data causes a degradation of performance. Since FlyAwareV2 also provides a large set of unlabeled real-world samples, a viable solution is to use them to apply Unsupervised Domain Adaptation (UDA) strategies.

Taking inspiration from the wide literature in the field \cite{Schwonberg23,Toldo20}, we tested 2 different classic UDA approaches, which can be considered as benchmarks: 
\begin{enumerate}
    \item A min-entropy UDA strategy (MaxSquareIW, MSIW \cite{chen2019domain}); in this case, the fine-tuning starts from the architectures pre-trained on the synthetic samples.
    \item A Min-entropy and multi-batch normalization combined strategy: here, we mix the MSIW approach with a common strategy for test-time domain adaptation, \ie, using different sets of batch normalization layers, one for each domain. 
    In essence, we add a set of batch-norms (initialized with the original values) and fine-tune them to estimate the real samples distribution (the supervision comes only from the MSIW loss). 
    The hypothesis behind these methods is that convolutional layers generalize across domains, while the BNs are domain-specific. 
    After training, we obtain a network with two sets of normalization layers (we denote the ones adapted to the source and target domains as BN-S and BN-T, respectively), leaving us with a choice between them at evaluation time. For completeness, we evaluate on the real samples using both, confirming an improvement over single-BNs architectures in all cases.
\end{enumerate}

The results are shown in Table \ref{tab:uda}: UDA strategies consistently improve performance across all weather conditions compared to the no-adaptation baseline.
\begin{wraptable}{r}{.5\textwidth}
    \setlength{\tabcolsep}{.3em}
    \setlength{\belowcaptionskip}{-1em}
    \centering
    \begin{tabular}{c|c|c|c:c}
        \multirow{2}{*}{Weather} & \multirow{2}{*}{no-UDA} & \multirow{2}{*}{UDA} & \multicolumn{2}{c}{UDA-BN} \\
        & & & BN-S & BN-T \\
        \hline
        Day & 49.7 & 53.2 & \textbf{54.2} & \underline{53.5} \\
        Night & 23.4 & \textbf{29.1} & \underline{28.5} & 27.5 \\
        Rain & 48.4 & 48.0 & \textbf{55.3} & \underline{48.5} \\
        Fog & 38.4 & 36.1 & \textbf{43.8} & \underline{38.5} \\
        \hline
        All & 42.3 & 44.0 & \textbf{47.1} & \underline{44.3} \\
    \end{tabular}
    \caption{Results after adapting on real (unlabeled) training FlyAwareV2 data. (Row-wise comparisons)}
    \label{tab:uda}
\end{wraptable}
The best overall mIoU of $47.1\%$ is achieved using the UDA-BN-S method, demonstrating the effectiveness of domain adaptation in bridging the synthetic-to-real gap for UAV scene understanding tasks. A visual example is also shown in the third row of Figure \ref{fig:quali_real_weathers}, which highlights the effectiveness of UDA strategies. Note how in the daytime sample all objects are much better defined, how in nighttime and rain conditions the vegetation is restored, and in foggy environments the vehicles are better identified.

\subsection{Multimodal Segmentation Experiments}
\label{sub:results:exp_mm}

Following \cite{rizzoli2023syndrone}, we evaluated our multimodal data using two benchmark architectures, one for early fusion and one for late (output-level) fusion.

\begin{wraptable}{r}{.45\textwidth}
    \setlength{\tabcolsep}{.3em}
    \setlength{\belowcaptionskip}{-1em}
    \centering
    \begin{tabular}{c|c:c|c}
        \multirow{2}{*}{Modality} & \multicolumn{2}{c|}{Synth} & \multirow{2}{*}{Real}\\
        & Fine & Coarse \\
        \hline
        RGB & 42.5 & 64.5 & \underline{42.3} \\
        D & \underline{67.1} & \underline{82.3} & 17.0 \\
        RGBD Early & 63.5 & 80.0 & \textbf{47.8} \\
        RGBD Late & \textbf{68.4} & \textbf{82.8} & 25.6 \\
    \end{tabular}
    \caption{Multimodal experiments. Real results in coarse class-set.}
    \label{tab:all_real}
\end{wraptable}

They represent standard baseline approaches to multimodal fusion, highlighting the data quality and generalizability, rather than focusing on the achievements possible with highly complex state-of-the-art multimodal schemes.
In the first, we simply concatenate RGB data and (normalized) Depth at the input level, obtaining a 4-channel-input architecture.
In the second, we merge the multimodal information at the output level: we duplicate the segmentation architecture, computing a prediction from RGB and Depth independently, before concatenating them together and merging them into a single prediction using a $1 \times 1$ Convolution without bias that maps the two outputs to a single one with the same number of channels. %

\begin{figure*}[t]
    \centering
    \begin{subfigure}{\textwidth}
        \centering
        \begin{subfigure}{.3\textwidth}
            \centering
            \caption*{RGB}
            \includegraphics[width=\textwidth]{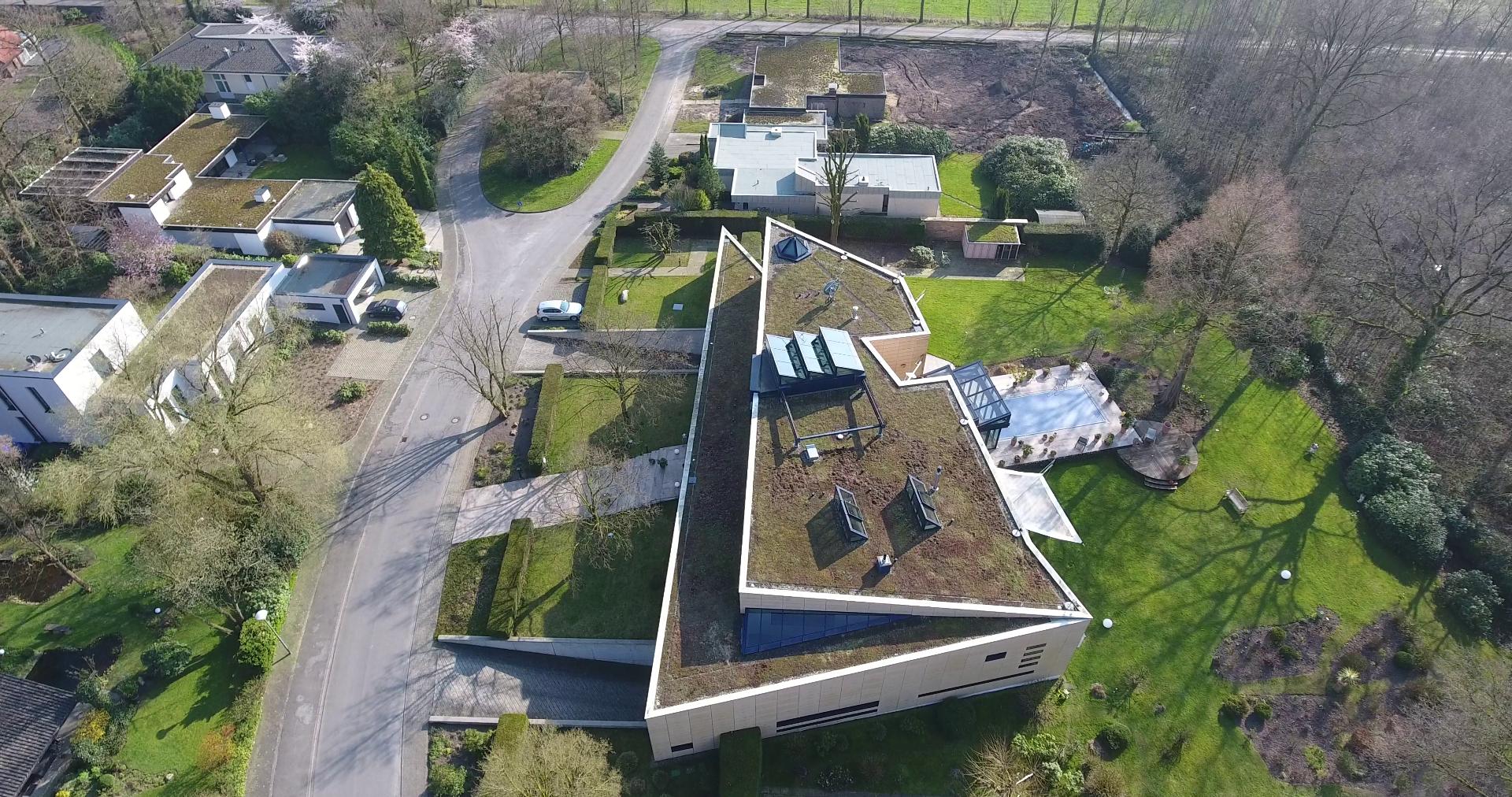}
        \end{subfigure}
        \begin{subfigure}{.3\textwidth}
            \centering
            \caption*{Depth}
            \includegraphics[width=\textwidth]{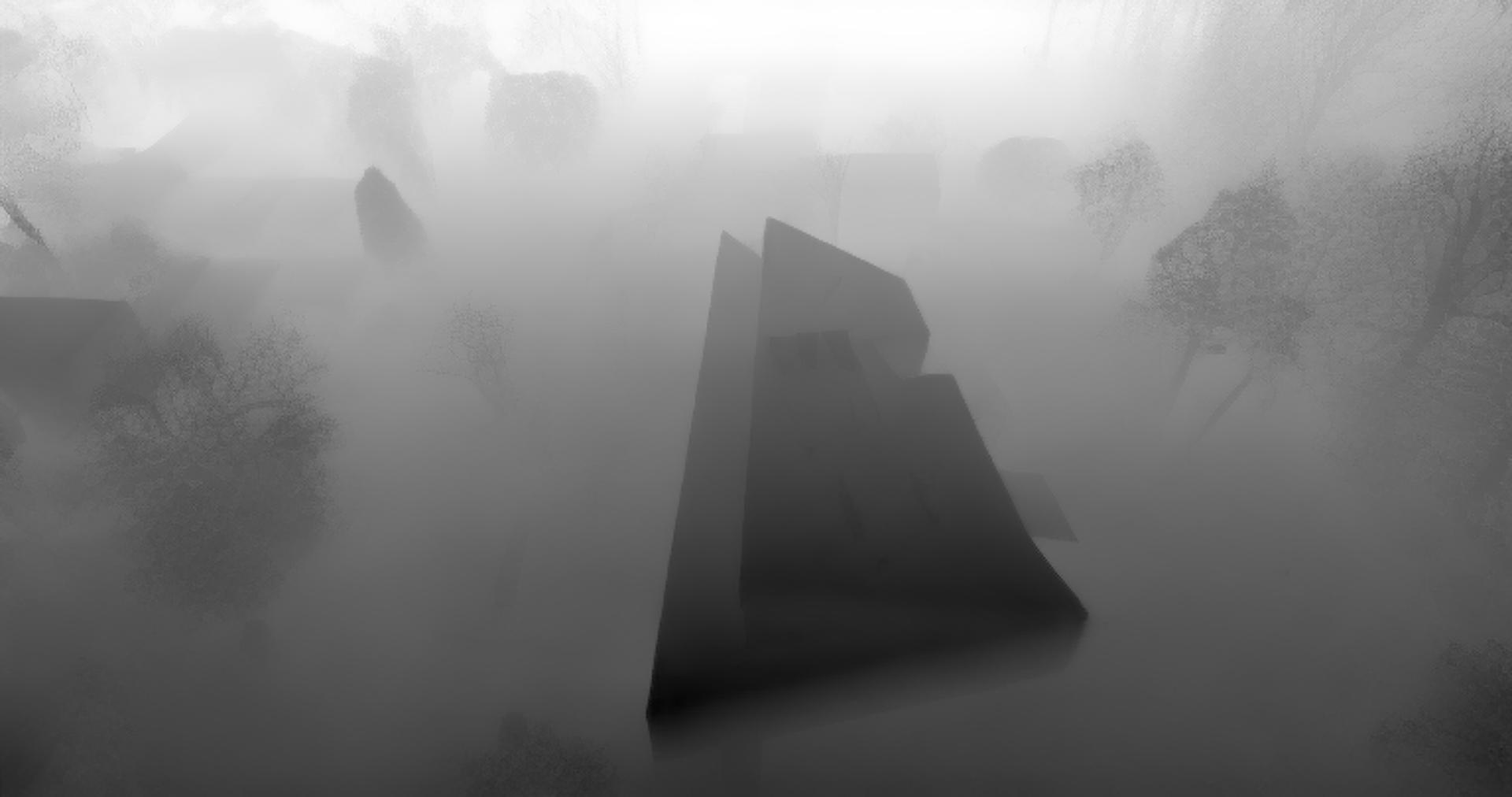}
        \end{subfigure}
        \begin{subfigure}{.3\textwidth}
            \centering
            \caption*{GT}
            \includegraphics[width=\textwidth]{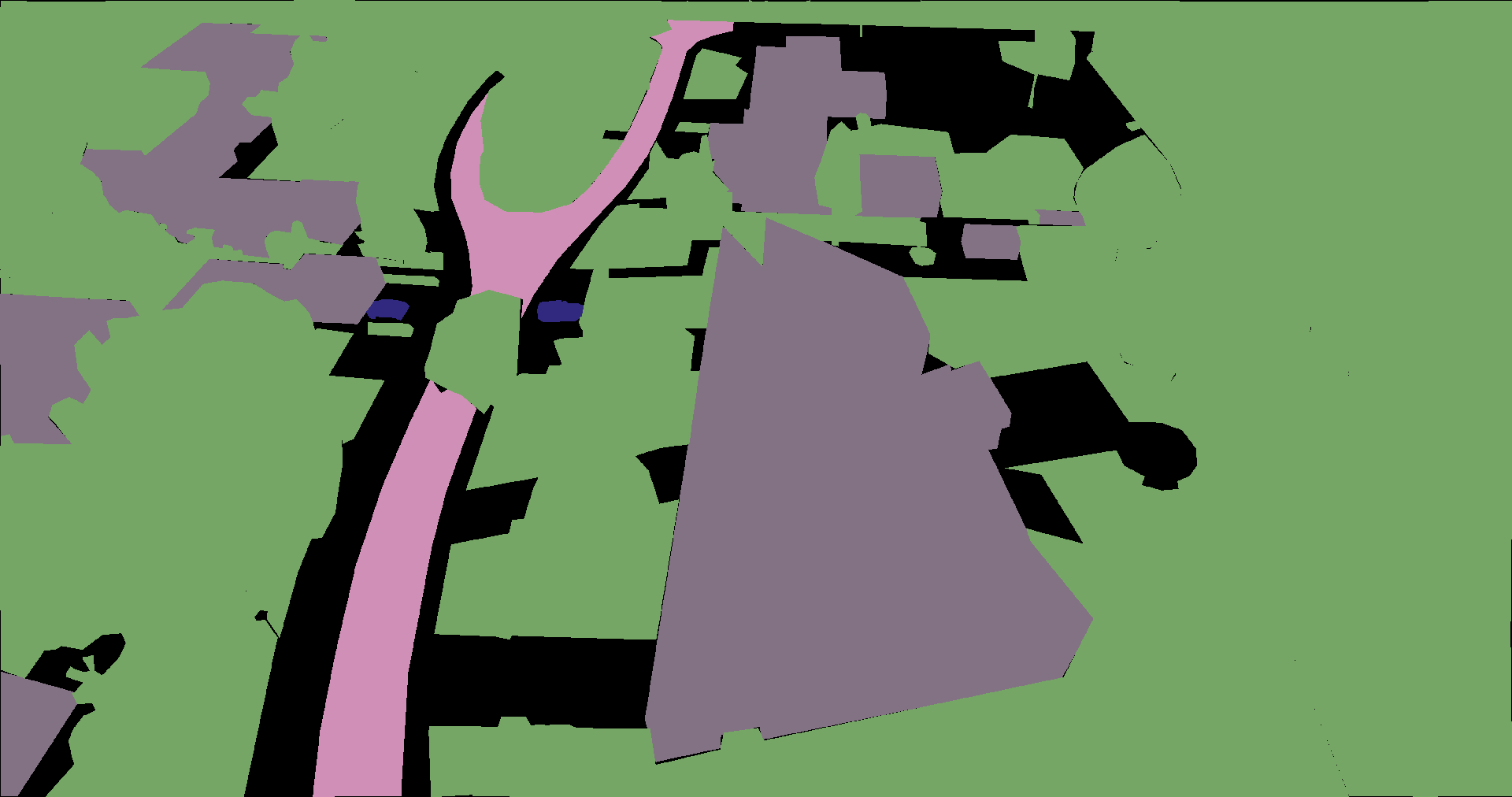}
        \end{subfigure}
    \end{subfigure}
        \begin{subfigure}{\textwidth}
        \centering
        \begin{subfigure}{.24\textwidth}
            \centering
            \includegraphics[width=\textwidth]{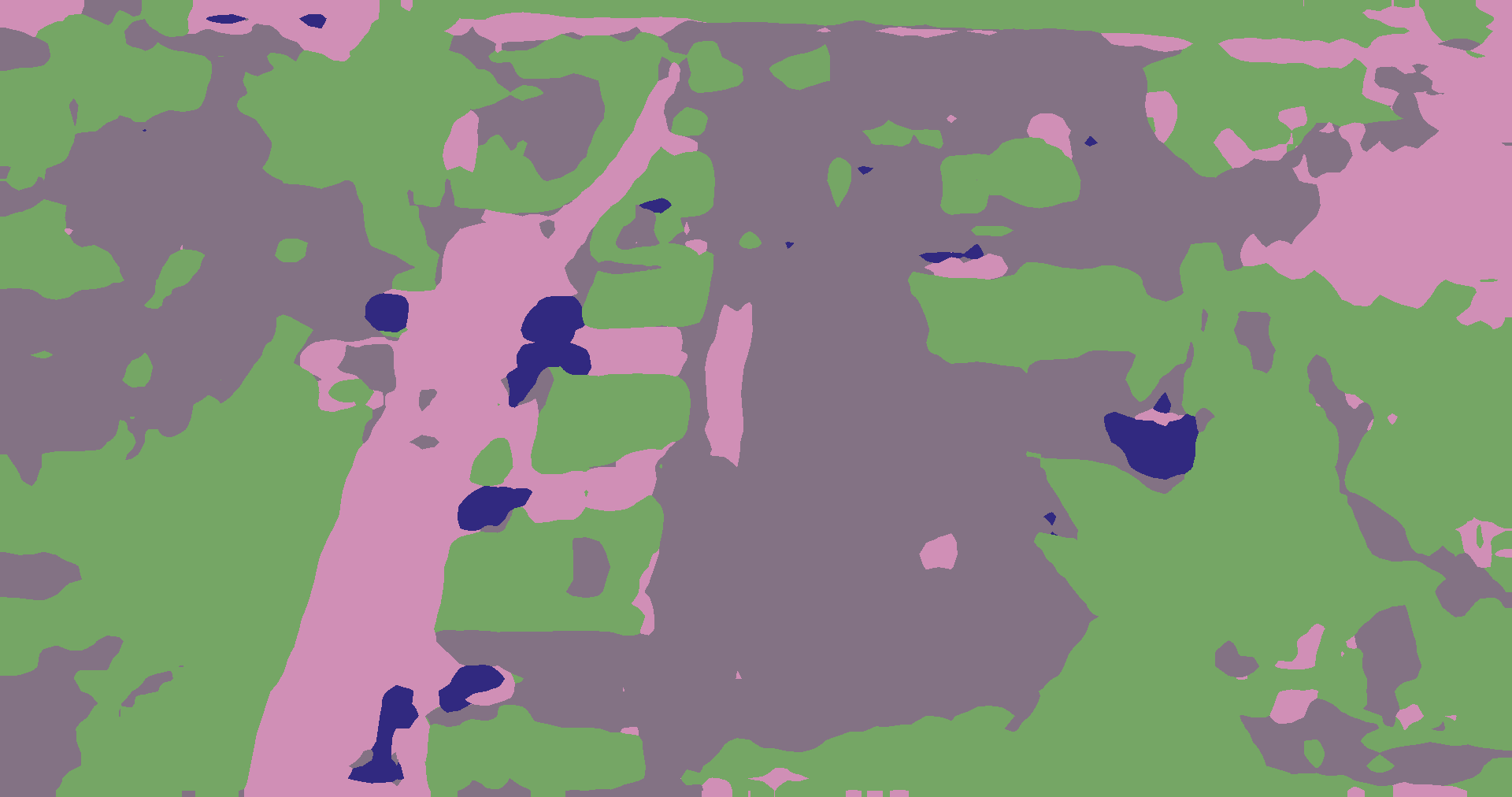}
            \caption*{RGB Only Prediction}
        \end{subfigure}
        \begin{subfigure}{.24\textwidth}
            \centering
            \includegraphics[width=\textwidth]{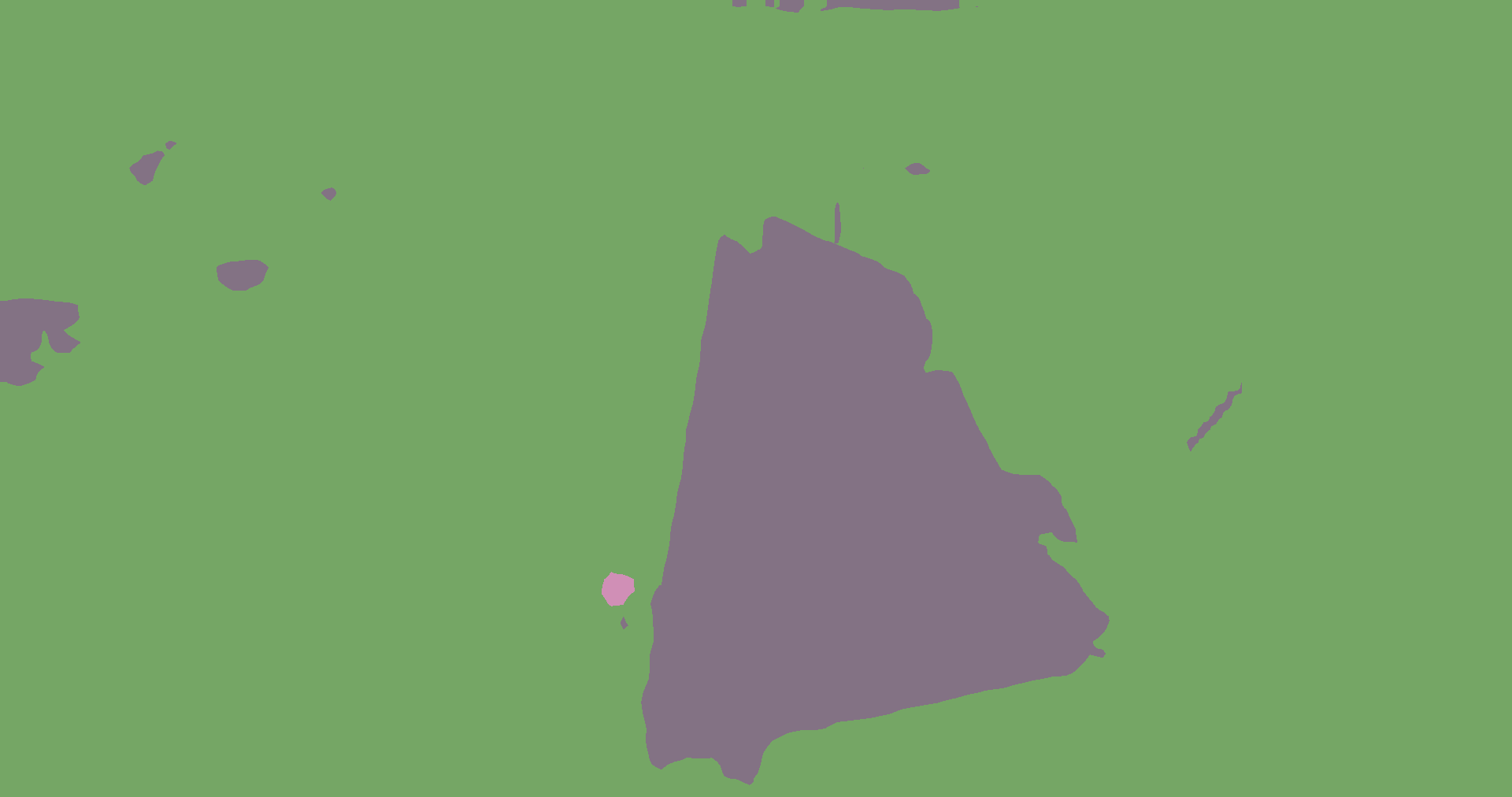}
            \caption*{Depth Only Prediction}
        \end{subfigure}
        \begin{subfigure}{.24\textwidth}
            \centering
            \includegraphics[width=\textwidth]{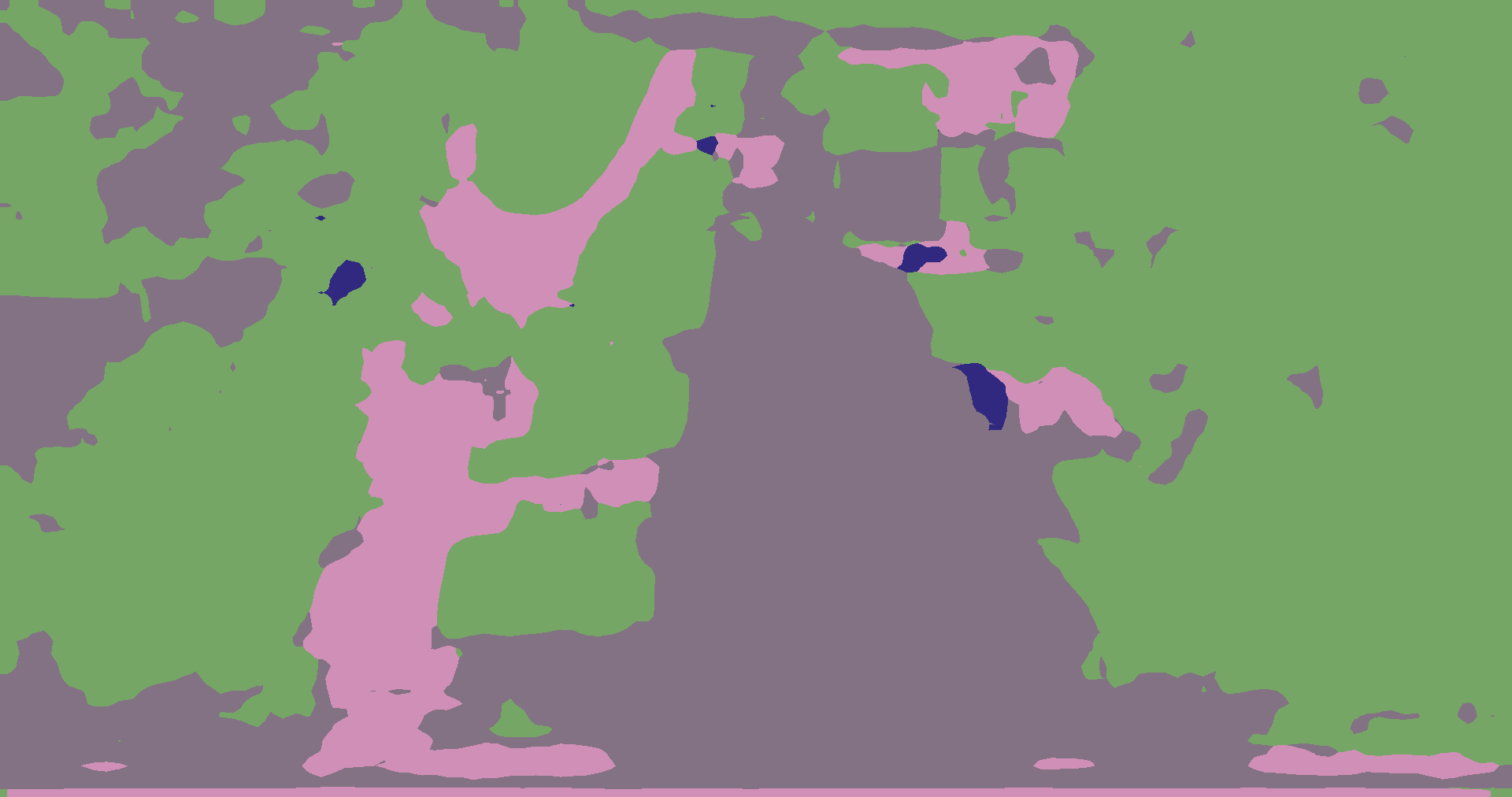}
            \caption*{RGBD Early Fusion Pred.}
        \end{subfigure}
        \begin{subfigure}{.24\textwidth}
            \centering
            \includegraphics[width=\textwidth]{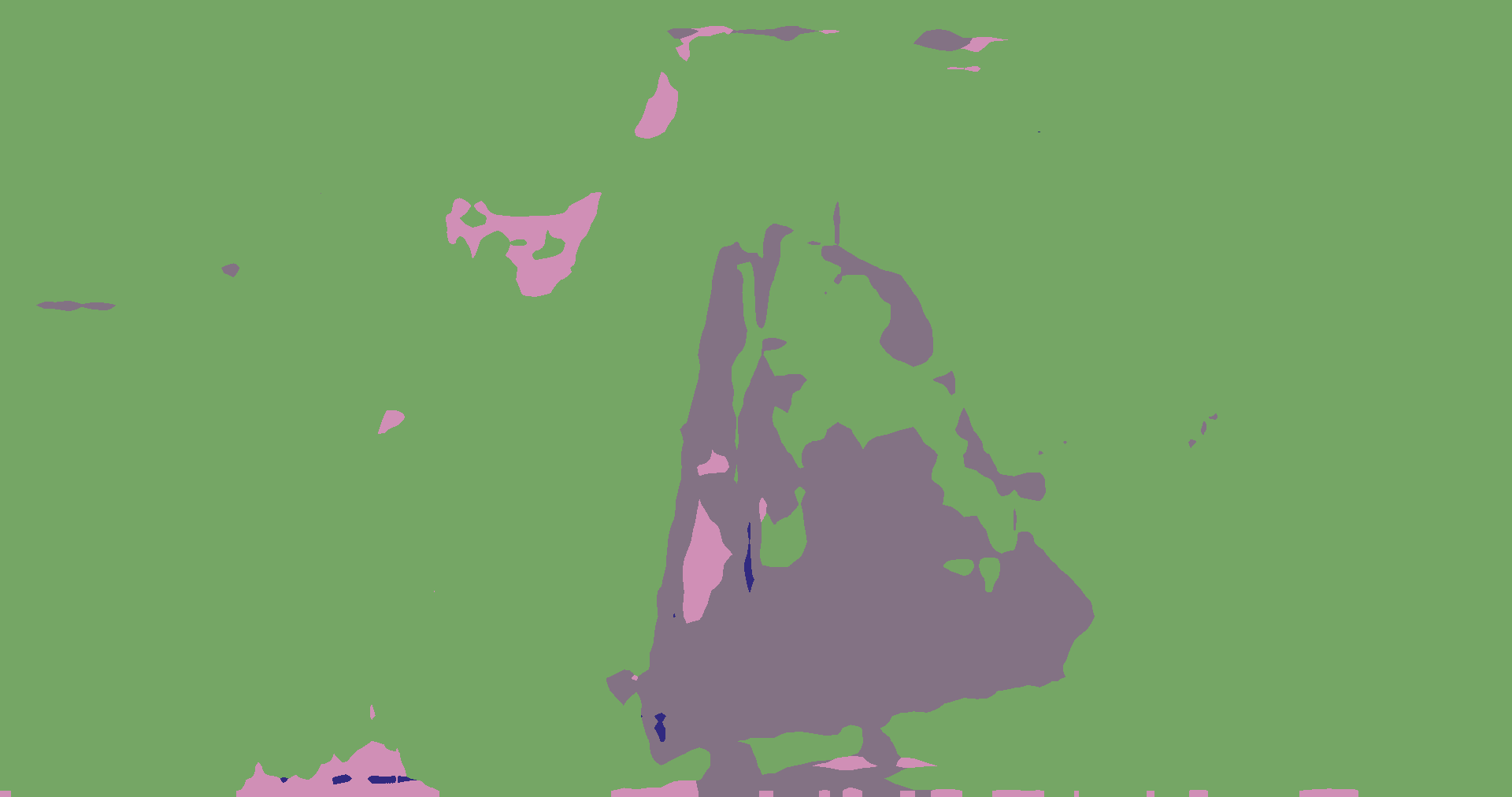}
            \caption*{RGBD Late Fusion Pred.}
        \end{subfigure}
    \end{subfigure}
    \caption{Qualitative experiments: the second row shows the predictions from  models trained with color alone, depth alone, and multimodal data and tested on real samples. }
    \label{fig:quali_multimodal}
\end{figure*}

The benefits of multimodal fusion are evident in Table~\ref{tab:all_real}, which compares the performance of models trained on RGB, depth (D), and their combination using early or late fusion. Although the depth-only model performs poorly (mIoU of $17.0\%$), incorporating depth information through early fusion with RGB significantly boosts performance to $47.8\%$ mIoU. 
However, late fusion of RGB and depth modalities yields suboptimal results ($25.6\%$ mIoU), highlighting the importance of early multimodal integration for effective feature learning.
These findings are confirmed by the qualitative results reported in Figure \ref{fig:quali_multimodal}, where one can appreciate how, compared to the second-best (RGB), the early fusion strategy leads to a much better segmentation of the Y bend in the road, as well as no confusion of the vegetation on the right side of the scene.

    \section{Conclusions and Future Work} \label{sec:conclusions}

In this paper we introduced FlyAwareV2, a novel large-scale dataset for UAV computer vision applications encompassing synthetic and real world multimodal information in varying weather conditions.
We provide experimental benchmarks showing how the large amount of provided synthetic data can be used to train segmentation models achieving effective performances on real world imagery. We also explored the domain transfer capabilities of segmentation models across different weathers, daytimes, flying heights and environments.
Finally we also showed how performances can be improved by exploiting multimodal data.
Further extensions of the dataset will consider other computer vision tasks such as object detection or panoptic segmentation and the extension of the amount of real world data.

\vspace{1cm}

\noindent \textbf{Acknowledgement:}
This work was partially supported by the European
Union under the Italian National Recovery and Resilience Plan (NRRP) of NextGenerationEU, partnership
on ”Telecommunications of the Future” (PE00000001- program ”RESTART”).

\newpage

    \bibliographystyle{ieeetr}
    \bibliography{refs}
\end{document}